\documentclass[acmsmall]{acmart}

\usepackage{amsmath,amsfonts}
\usepackage{algorithmic}
\usepackage{graphicx}
\usepackage{textcomp}
\usepackage{xcolor}
\usepackage{booktabs}
\usepackage{tikz}
\usepackage{float}
\usepackage{multirow}
\usepackage{booktabs}
\usepackage{comment}
\usepackage{subfigure}
\usepackage{amsmath}
\usepackage{xspace}
\usepackage{balance}
\usepackage{todonotes}
\usepackage{caption} 
\usepackage{newtxmath}
\usepackage{dblfloatfix}
\usepackage{gensymb}
\usepackage{lipsum} 
\usepackage[]{mdframed}
\usepackage{tcolorbox}

\newcommand{\attack}{$Lens Attack$\xspace}
\newcommand\cbox[1]{\begin{tcolorbox}#1\end{tcolorbox}}

\AtBeginDocument{%
  }

\setcopyright{acmlicensed}
\copyrightyear{2018}
\acmYear{2018}
\acmDOI{XXXXXXX.XXXXXXX}

\acmJournal{JACM}
\acmVolume{37}
\acmNumber{4}
\acmArticle{111}
\acmMonth{8}

\begin{document}

\title{Optical Lens Attack on Monocular Depth Estimation for Autonomous Driving}

\author{Ce Zhou}
\email{zhouce@msu.edu}
\affiliation{%
  \institution{Michigan State University}
  \city{East Lansing}
  \state{Michigan}
  \country{USA}
}

\author{Qiben Yan}
\authornote{Corresponding author.}
\email{qyan@msu.edu}
\affiliation{%
  \institution{Michigan State University}
  \city{East Lansing}
  \state{Michigan}
  \country{USA}
}

\author{Daniel Kent}
\affiliation{
\institution{Michigan State University}
  \city{East Lansing}
  \state{Michigan}
  \country{USA}}
\email{kentdan3@egr.msu.edu}

\author{Guangjing Wang}
\affiliation{
\institution{University of South Florida}
  \city{Tampa}
  \state{Florida}
  \country{USA}}
 \email{guangjingwang@usf.edu} 

 \author{Weikang Ding}
\affiliation{
\institution{Michigan State University}
  \city{East Lansing}
  \state{Michigan}
  \country{USA}}
 \email{dingweik@msu.edu} 

\author{Ziqi Zhang}
\affiliation{%
 \institution{Peking University}
  \city{Beijing}
  \country{China}}
\email{ziqi_zhang@pku.edu.cn}  

\author{Hayder Radha}
\affiliation{
\institution{Michigan State University}
  \city{East Lansing}
  \state{Michigan}
  \country{USA}}
 \email{radha@egr.msu.edu}

\renewcommand{\shortauthors}{Zhou et al.}

\begin{abstract}
  Monocular Depth Estimation (MDE) is a pivotal component of vision-based Autonomous Driving (AD) systems, enabling vehicles to estimate the depth of surrounding objects using a single camera image. This estimation guides essential driving decisions, such as braking before an obstacle or changing lanes to avoid collisions. In this paper, we explore vulnerabilities of MDE algorithms in AD systems, presenting \attack, a novel physical attack that strategically places optical lenses on the camera of an autonomous vehicle to manipulate the perceived object depths. \attack encompasses two attack formats: concave lens attack and convex lens attack, each utilizing different optical lenses to induce false depth perception. We first develop a mathematical model that outlines the parameters of the attack, followed by simulations and real-world evaluations to assess its efficacy on state-of-the-art MDE models. Additionally, we adopt an attack optimization method to further enhance the attack success rate by optimizing the attack focal length. To better evaluate the implications of \attack on AD, we conduct comprehensive end-to-end system simulations using the CARLA platform. The results reveal that \attack can significantly disrupt the depth estimation processes in AD systems, posing a serious threat to their reliability and safety. Finally, we discuss some potential defense methods to mitigate the effects of the proposed attack. 
\end{abstract}

\ccsdesc[500]{Security and privacy~Systems security}
\ccsdesc[500]{Computing methodologies~Object detection}

\keywords{Autonomous Driving; Camera; Monocular Depth Estimation; Autonomous Vehicle; Optical Lens}

\received{20 February 2007}
\received[revised]{12 March 2009}
\received[accepted]{5 June 2009}

\maketitle

\section{Introduction}


Autonomous Driving (AD) systems rely on their perception modules to track and regulate the proximity to surrounding obstacles, a vital capability for ensuring the system’s safety and dependable performance. Various methods exist to accomplish this task, such as direct measurement techniques using radar or Lidar~\cite{piotrowsky2019enabling,li2020lidar}, or employing stereoscopic 3D imaging to create a dense depth map of the environment~\cite{mayer2016large,chang2018pyramid,tay2019aanet}.

Cameras are among the most critical sensors in AD systems, as demonstrated by their use in vehicles from companies like Tesla~\cite{Autopilot}, Uber~\cite{Uber}, and Waymo~\cite{Waymo}. These cameras leverage computer vision technology for various AD tasks~\cite{zhou2023comprehensive}. Researchers have made significant advancements, allowing monocular cameras to estimate scene depth~\cite{monodepth2,wong2020targeted,casser2019depth}. Utilizing monocular cameras can reduce the need for additional sensors, offering benefits in terms of space, weight, and cost efficiency. Although estimating depth with a single camera poses challenges, recent deep-learning approaches have achieved performance levels close to those of stereo 3D depth estimation techniques.
 
Previous security studies have introduced various attack techniques targeting cameras to disrupt different AD tasks, including object detection and classification~\cite{eykholt2018robust,man2020ghostimage,nassi2020phantom}, lane detection~\cite{jing2021too,sato2021dirty}, traffic light detection~\cite{yan2022rolling}, and vision-based depth estimation~\cite{zhou2022doublestar,zhang2020adversarial,wong2020targeted}. Additionally, the security of vision-based depth estimation tasks has also been investigated. 
To compromise 3D stereo depth estimation, Zhou et al.~\cite{zhou2022doublestar} propose a long-range stereo depth estimation attack that injects fake obstacle depth by projecting pure light from two complementary light sources. 
For monocular depth estimation (MDE) algorithms, 
Zhang et al.~\cite{zhang2020adversarial} and Wong et al.~\cite{wong2020targeted} present white-box attacks that use imperceptible additive adversarial perturbations to alter the depth estimation results in the digital world, while a black-box attack is introduced by Daimo et al. in \cite{daimo2021black}. However, these invisible perturbations are ineffective in the physical world due to the impacts of environmental variables. Therefore, Yamanaka et al.~\cite{yamanaka2020adversarial} and Cheng et al.~\cite{cheng2022physical} create visible adversarial patches. These patches deceive depth estimation algorithms into estimating a false depth for the regions where the patterns are placed in the physical world. However, human drivers can easily detect the patches. Moreover, patches are scene-sensitive, and may not work well in dynamic environments. As opposed to existing physical attacks, we propose a universal black-box attack, \attack, that enables a new type of robust physical attack using optical lenses. 
 
\attack leverages a fundamental vulnerability of MDE systems, where even a slight change in the perceived object size in an image can lead to a corresponding shift in estimated depth. Our attack utilizes the optical lens to change the formed object size on the image sensor. Specifically, by attaching a tiny attack lens in the near front (e.g., $5cm$) of the car camera, the sensed object size will be altered, which affects the depth estimation results. 

There are four major challenges in realizing \attack. (i) ``How to design the attacks that can induce various false depth predictions?'' (ii) ``How to mathematically calculate the induced depth and gain control over depth estimation?'' (iii) ``How to configure the optimal attack parameters to accurately manipulate the depth?'' And (iv) ``How to verify the effectiveness of the \attack in both AI and system level?''

To address the first challenge, 
we design \attack in two attack formats: concave lens attack and convex lens attack, which can either 
increase or decrease the object depth. 
To solve the second challenge, we mathematically model our attack using different lenses in various attack scenarios. To improve the depth manipulation precision (the third challenge), we first devise our attack in the form of targeted and untargeted attacks. Then, we adopt an optimization method to formulate the attacking process and output the optimized parameters for enhancing the attack performance. To address the last challenge, we verify the efficacy of \attack through both simulation and real-world experiments with a prototype autonomous vehicle (AV) against three state-of-the-art MDE algorithms. We also simulate our attack at an end-to-end system level using the CARLA simulator to demonstrate its potential impact in the physical world. 

Our experimental results demonstrate that our attack remains effective across a wide range of optical lens parameter configurations. 
When the optical lens is within the field of view, the attack optimization further enhances the accuracy of manipulated depths.
We set up a demo website\footnote{\url{https://lensattack.github.io/.}} to show our attack results, attack simulations, and physical attack video demos.

The main contributions of this work are summarized as follows:
\begin{itemize}
\item We propose a novel universal physical attack, \attack, on MDE algorithms that utilizes optical lenses. By investigating the vulnerability of the MDE, we propose the concave and convex lens attacks and mathematically model them in different attack scenarios.
\item We evaluate the attacks on a smartphone camera and an AV in real-world experiments to demonstrate that our attack is effective with various optical lens parameter settings. The concave lens attack results in an average error rate of 11.48\% in estimated depths, whereas the convex lens attack leads to a 29.84\% average error rate. 

\item We present an attack optimization method that further improves the attack success rate by considering the factor of blurriness. The targeted average attack error rate is reduced by 6.26\%, while the untargeted average attack distortion rate is increased by 11.58\%.

\item We perform an end-to-end system-level simulation in the CARLA simulator and demonstrate the real-world ramifications of the proposed attack.

\end{itemize}

This work extends our preliminary work~\cite{zhou2024lensattack} with significant enhancements in Sections~\ref{sec:attack_design}, ~\ref{sec:evl} and~\ref{sec:discussion}. To further improve the effectiveness of our attack, we introduce an optimization method to optimize the focal length used in both targeted and untargeted attacks with the goal of improving the attack success rate (Section~\ref{sec:optimization}). We then evaluate the performance of the optimization method in simulation and demonstrate its improvements (Section~\ref{sec:evl_opti}). Additionally, to validate the feasibility of \attack, we conduct a comprehensive end-to-end system-level simulation using the CARLA simulator (Section~\ref{sec:end-to-end}). We also present more in-depth discussions and results on the potential defense methods in Section~\ref{sec:discussion}.  

\section{Background}

In this section, we briefly introduce the preliminary background knowledge of \attack, including the optical principles for optical lenses and monocular vision based depth estimation.

\subsection{Optical Principles for Lenses}\label{sec:background}
An optical lens, typically made of transparent materials such as glasses, is employed to produce an image through the concentration of light rays emanating from an object~\cite{Britannica}. This is accomplished by exploiting the phenomenon of refraction, which arises when light passes from one medium (such as air) to another (the lens). As a result, refraction occurs both upon entering the lens and upon exiting it back into the air. Optical lenses have wide-ranging applications, including but not limited to, eyeglasses, magnifiers, projection condensers, signal lights, viewfinders, cameras, and more.

An optical lens typically has a circular shape and possesses two polished surfaces, which can either be concave or convex. There are different types of lenses based on the curvature of the two opposite surfaces. Regarding the prevalence and availability, we mainly focus on the lens whose two surfaces are both concave or convex, which is called \emph{double concave} or \emph{double convex} lens. For simplicity, we will refer to them as ``concave'' and ``convex'' lenses. 

\begin{figure}
    \centering    
    \subfigure[]{\includegraphics[width=0.22\textwidth]{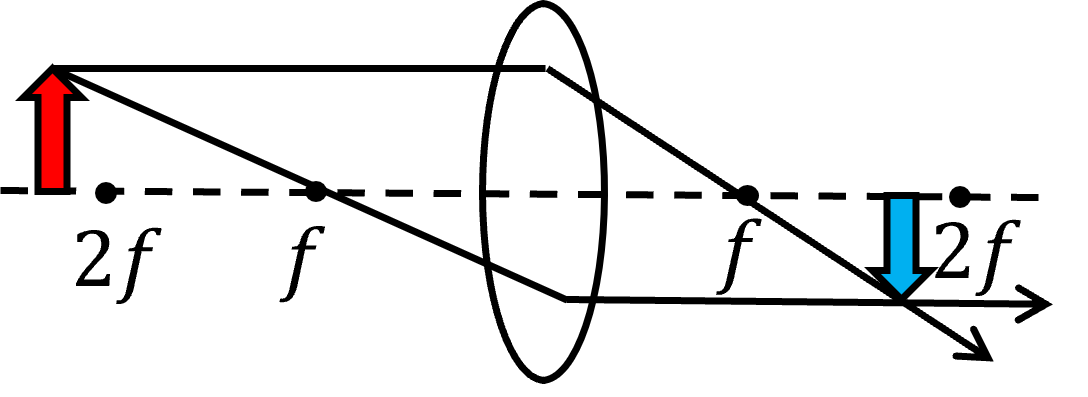}} 
    \subfigure[]{\includegraphics[width=0.23\textwidth]{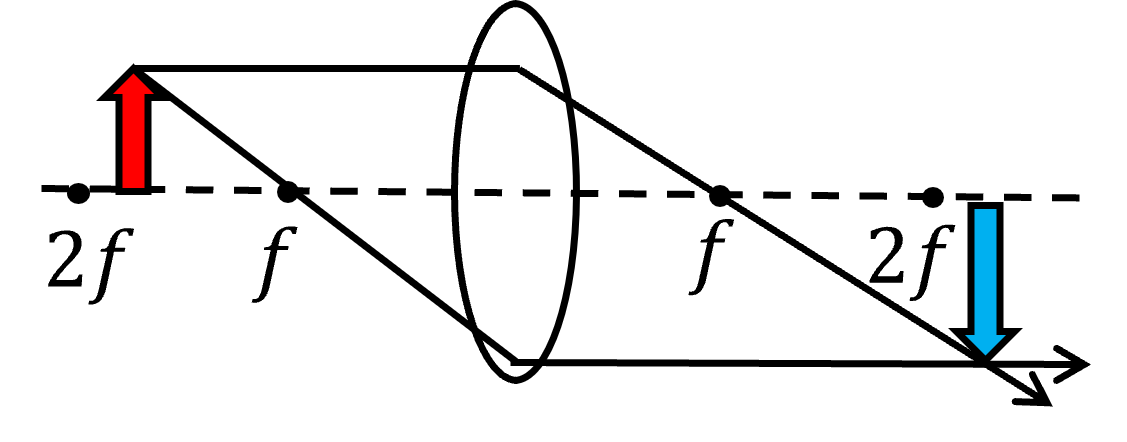}} 
    \subfigure[]{\includegraphics[width=0.19\textwidth]{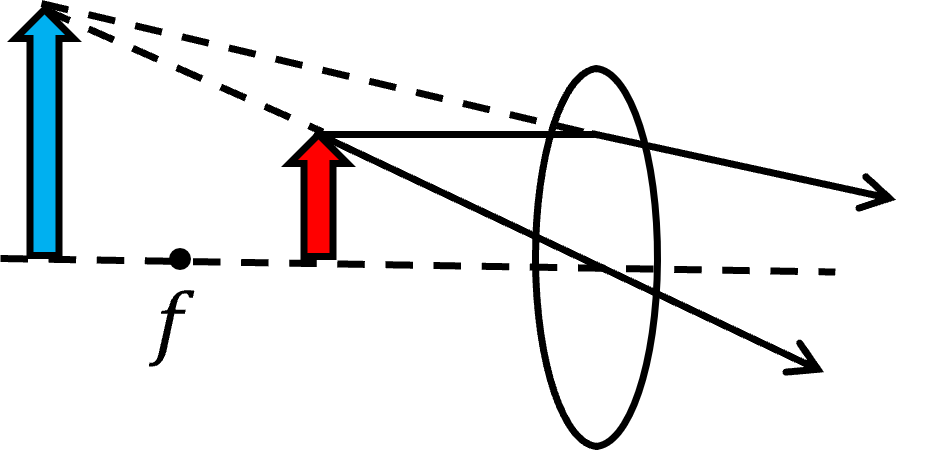}}   
    \subfigure[]{\includegraphics[width=0.19\textwidth]{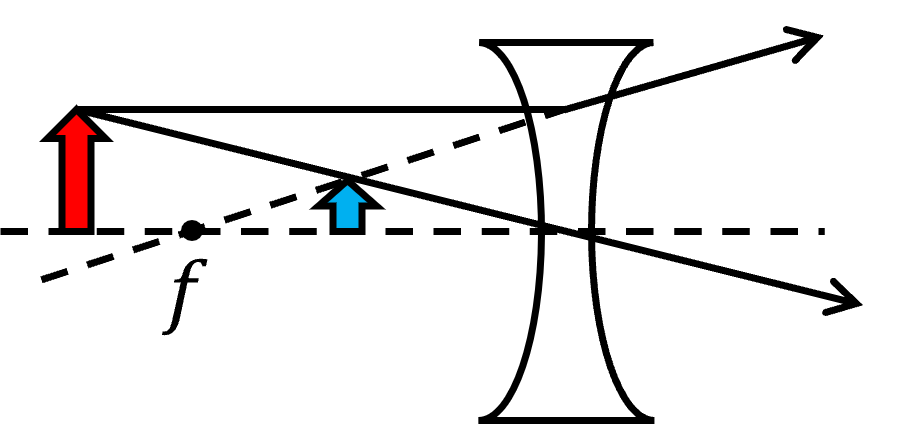}} 
        \vspace{-10pt}
    \caption{Ray diagrams of concave and convex lenses. (a)(b)(c) show the images formed by a convex lens, and (d) displays the images formed by a concave lens. The object is shown in red arrow and its corresponding image formed by the attack lens is shown in blue. $f$ represents the focal length.}
    \label{fig:background_oplens}
\end{figure}

A focal point, also known as a principal focus (denoted as $f$ in Fig.~\ref{fig:background_oplens}), is the point at which parallel rays can be made to converge or appear to diverge~\cite{Britannica,Focus}. A lens has two focal points, one on each side, so light can pass through it in either direction. The distance from the center of the lens to the focal point is the \emph{focal length}.

Fig.~\ref{fig:background_oplens} shows the visual image of the object formed by a convex lens and a concave lens due to ray refractions. For the convex lens, the size of the images formed can vary significantly compared to the object, depending on the focal length of the lens $f$ and the distance between the lens and the object $d_o$. There exist three possible cases: (Case 1) When $d_o \geq 2f$, it forms a real, inverted and smaller image where the distance between the lens and the image is $f\leq d_i\leq 2f$. (Case 2) When $f < d_o < 2f$, it creates a real, inverted and larger image at $d_i> 2f$. (Case 3) When $ 0< d_o < f$, it produces a virtual, upright, and larger image behind the object on the same side of the lens. On the other hand, the images formed by a concave lens are always virtual, upright, and smaller between the object and the lens regardless of the object's position.

The relationship between $f$, $d_o$ and $d_i$ can be written as: 
\begin{equation}\label{equ1}
    \frac{1}{f}=\frac{1}{d_o}+\frac{1}{d_i},
\end{equation}
where $f$ is a positive number if it is a convex lens, otherwise a negative number, $d_o$ is always a positive number, and $d_i$ is a positive number if the lens is on the opposite side of the object, otherwise a negative number.

The magnification of the formed image is:
\begin{equation}\label{equ2}
    m=-\frac{d_i}{d_o}.
\end{equation}
If $m$ is a positive number, the image is upright. 

Based on Eqs.~(\ref{equ1}) and (\ref{equ2}), in Case (3) with images formed by a convex lens, a larger absolute value of focal length results in a smaller image. In other cases, a larger absolute value of focal length leads to a larger image of an object, whereas a smaller absolute value of focal length forms a smaller image.



\subsection{Monocular Depth Estimation}\label{section:MonocularDepthEstimation}
The objective of MDE is to determine each pixel's depth value from a single 2D RGB image.  MDE has become popular in the study of robotic tasks~\cite{datar2023learning,datar2023toward}. Previous studies use supervised training methods to estimate the depth from a single image~\cite{liu2015deep,eigen2015predicting,dijk2019neural,yin2019enforcing}. The lack of high-quality depth maps, however, has spurred the adoption of unsupervised/self-supervised learning. Instead of using crowd-sourced data~\cite{chen2016single} for training, the recently proposed methods use stereo-pairs~\cite{garg2016unsupervised,godard2017unsupervised,pillai2019superdepth} or monocular video sequences~\cite{zhou2017unsupervised,wang2018learning,casser2019depth,yin2018geonet,guizilini20203d}. Video-based algorithms for depth estimation typically provide depth values in an unknown scaled format, whereas stereo-based methods can predict depth in metric units if the baseline between the cameras is known~\cite{wong2020targeted}.

Monodepth2~\cite{monodepth2} uses both stereo and video-based methods and leverages reprojection losses to eliminate potential occlusions. However, reprojection losses from stereo-based self-supervision typically have multiple local minima which restrict the network learning, and further lead to the limited quality of the predicted depth map. Depth Hints~\cite{watson2019self} enhances an existing loss function to better guide a network to learn weights. 
However, these algorithms are hard to be deployed on edge devices due to its heavy computation. Lite-mono~\cite{zhang2023lite} is a lightweight but effective model, which significantly reduces the number of trainable parameters, by employing a hybrid CNN and Transformer architecture. In this paper, we aim to attack these three cutting-edge methods to demonstrate the broad impact of our proposed attack.

A pinhole camera is a simple model for approximating the monocular camera imaging process. Although the actual imaging is much more complicated than the pinhole imaging model, the pinhole imaging model is very convenient to apply mathematically, and the approximation to imaging is often acceptable~\cite{Pinholecameramodel}. 

Based on the pinhole imaging model (more details can be found in Appendix~\ref{sec:Pinhole Camera Model}), we learn that the depth of the same object is inversely proportional to its object size in the image. 
A change in the size of an object in the image will result in a corresponding change in depth. Therefore, for monocular camera imaging, as the distance between the object and the camera increases, its size becomes smaller on an image, and vice versa. 

MDE algorithms are vulnerable due to the lack of sufficient physical indications for scene depth in a monocular image alone. Current MDE algorithms appear to rely on implicit knowledge learned from the training dataset (e.g., color, location, or shadows) rather than physical cues~\cite{dijk2019neural,yamanaka2020adversarial}). Because they rely heavily on non-depth elements in the provided image, they are vulnerable to attacks that tamper with the images.

\section{Threat Model and Attack Scenario}\label{sec:threat_model}

In this section, we introduce our threat model and present the attack scenarios.

\subsection{Threat Model}

The attacker's goal is to disrupt the regular operations of an AV by alternating the MDE results using the optical lens and triggering unintended system behaviors. 
We consider an AV relying on the monocular camera as the major source for depth estimation, such as Tesla vehicles with Full Self-Driving features and Active Safety Features~\cite{Autopilot_and_full_self}.
The attacker tries to change the depth information of the objects, e.g., vehicles or pedestrians, by applying optical lenses within a near distance (e.g., $5cm$) to the target camera. As a result, this can cause the victim AV to crash and potentially lead to a severe car accident, such as colliding with the vehicle in front.


Our attack is a \emph{black-box attack} against general monocular vision-based depth estimation algorithms in AD systems. The attacker is assumed to have no prior knowledge of the depth estimation algorithms used in AD systems. We assume the attacker does not have access to the camera images. We also assume that an attacker has very limited physical access to the hardware or firmware of the victim. 
For example, if the victim's vehicle is parked in a public area, the attacker can get close to the vehicle and install the attacking equipment, e.g., using fixed suction cups to absorb the attack device on the camera or the car body depending on the outlook of the camera.

The attacker may need to use a 3D-designed lens holder
to hold the attack lens to ensure the stability of the attack lens during driving. Typically, the attacker would design 
the 3D lens holder in a compact size and with a transparent color or a color similar to that of the car body. This design approach is crucial for achieving a relatively stealthy attack. The attacker could also design the attack lenses with a proper size to make the attacks more accurate and stealthy. The implementation details of the attack device on a real vehicle are illustrated in Appendix~\ref{app:manufactured}.

\begin{figure}[!ht]
\centering
    \subfigure[Concave lens attack]{\includegraphics[width=0.7\columnwidth]{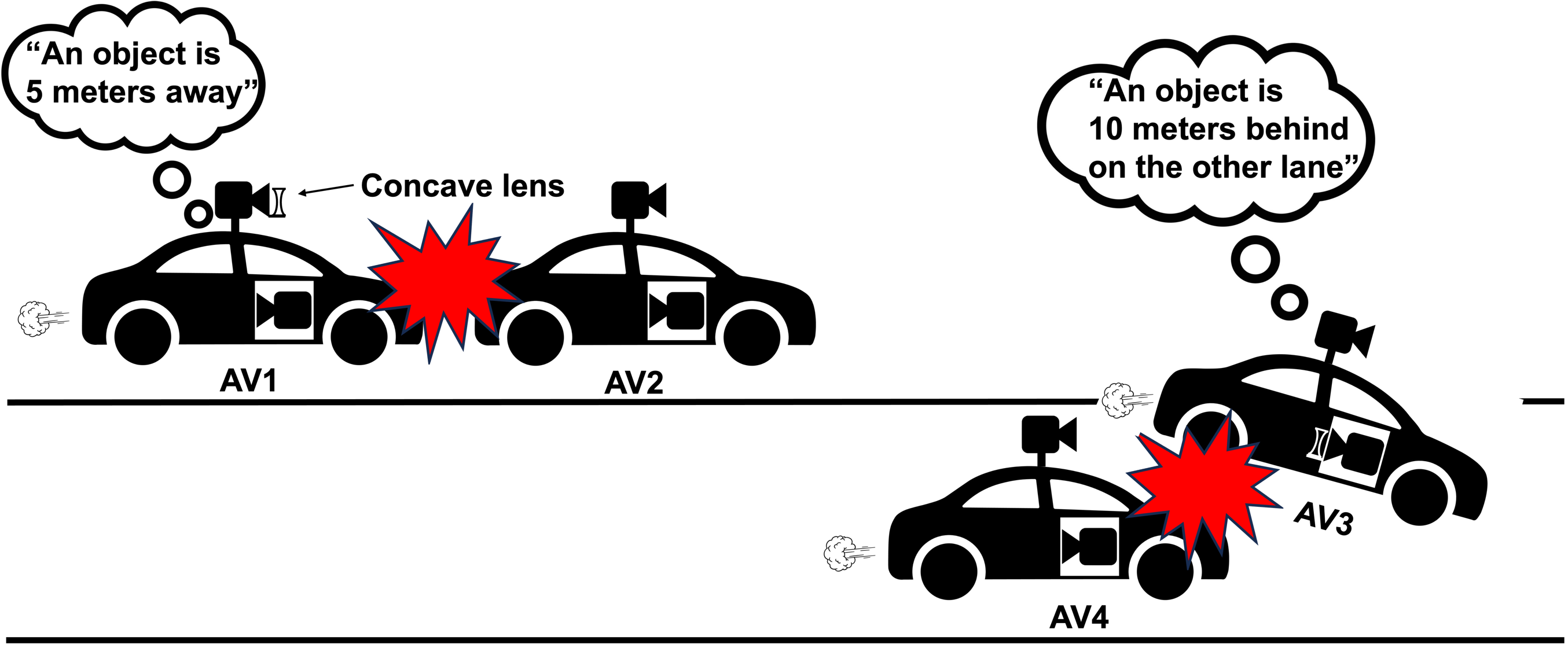}} 
    \label{fig:Op_lens_attack_cc}
    \subfigure[Convex lens attack]{\includegraphics[width=0.7\columnwidth]{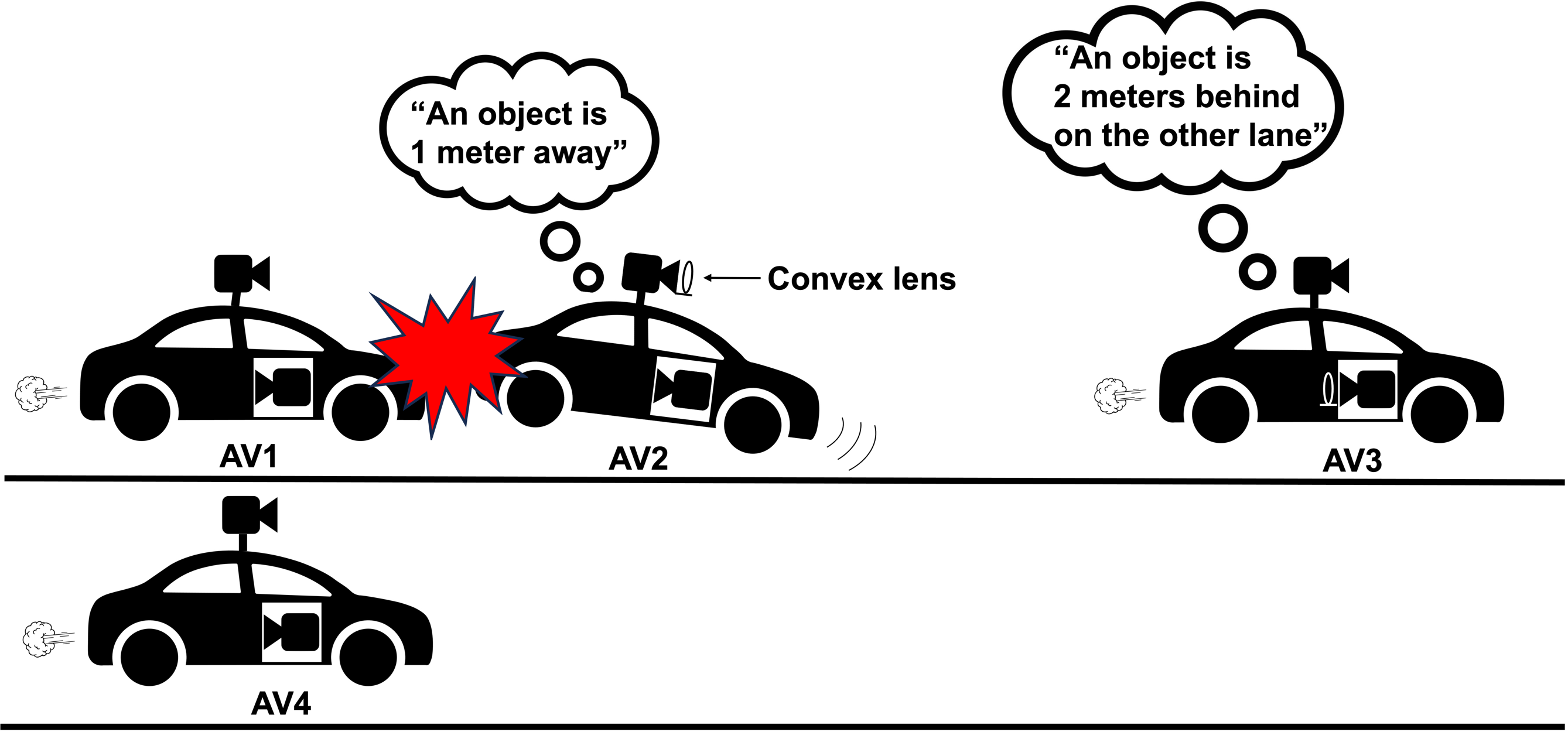}} 
    \label{fig:Op_lens_attack_cv}
    \caption{\attack attack scenarios. (a) illustrates the concave lens attack, and (b) showcases a convex lens attack.}
    \label{fig:Op_lens_attack}
\end{figure}

\subsection{Attack Scenario}
We present attack scenarios for normal driving and lane changing in Fig.~\ref{fig:Op_lens_attack}. 
In Fig.~\ref{fig:Op_lens_attack}(a), a concave lens is attached to the front camera of AV1, which extends the estimated depth for AV2 beyond its real distance. This false depth estimation could lead to a collision, posing a significant threat to human lives, or cause a hard brake that may injure passengers if other sensors are present and sensor fusion is employed. Even if the AV1 detects the proximity of the front car and attempts to brake, the distance between the AV1 and AV2 is smaller than the ``safety braking distance'', which is insufficient to prevent a collision. Besides, the AV2 may make a sudden lane change to avoid collision with the front vehicle, but this could potentially result in severe traffic jams or car accidents if the following vehicle on the other lane does not have sufficient time to react. Additionally, when a concave lens is applied to the side camera of AV3 during the lane changing, the depth estimation of AV4 by AV3 is farther away than its actual depth. As a result, AV3 shifts lanes accordingly, which could cause a potential collision with AV4 or hard braking of AV4.

Fig.~\ref{fig:Op_lens_attack}(b) demonstrates the convex lens attack. A convex lens on AV2's front camera modifies the depth of AV3 to be closer than its actual position. This misperception triggers a sudden brake, potentially causing a subsequent collision with AV1 if AV1 could not brake immediately. Similarly, a concave lens on the side camera of AV3 results in an estimated depth of AV4 being nearer than its actual depth. This may lead to delayed lane changing or, in certain situations, force AV3 to stop and wait to change lanes, potentially causing a traffic jam.

\section{Optical Lens Attack} \label{sec:attack_design}


\begin{figure}[t]
\centering
	\includegraphics[width=1\columnwidth]{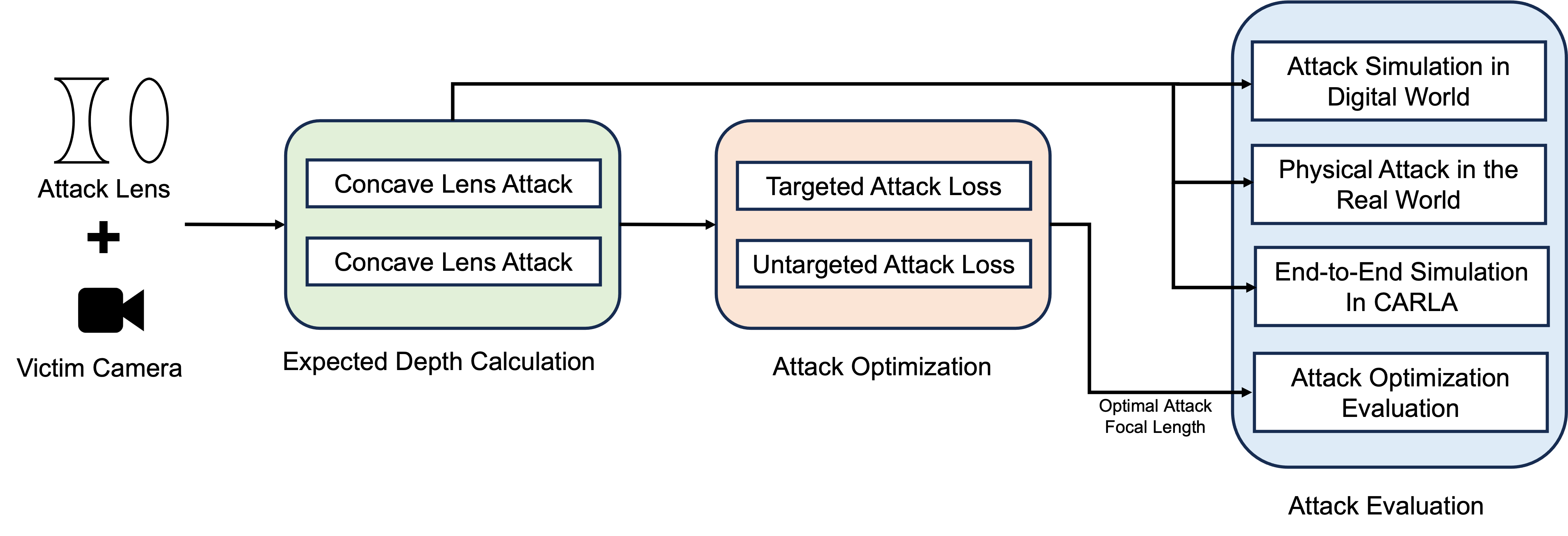}
	\caption{The pipeline of \attack. Based on the available attack lenses and the victim camera, the adversary first calculates the expected depth after the attack based on the principle of the optical lens. Then, they define and optimize the targeted and untargeted attacks to determine the optimal focal length.}
	\label{fig:attack_pipline}
\end{figure}

\begin{table}[t]
\centering
\caption{List of parameters and their meanings}
\label{tab:symbol}
\resizebox{0.7\columnwidth}{!}{%
\begin{tabular}{|l|l|l|}
\hline
\textbf{Symbol} &
  \textbf{Description} &
  \textbf{Sign} \\ \hline
$f$ &
  The focal length of the attack lens &
  \begin{tabular}[c]{@{}l@{}}Concave lens: negative;\\ Convex lens: positive\end{tabular} \\ \hline
$d_{o1}$ &
  \begin{tabular}[c]{@{}l@{}}The distance between the object\\ and the attack lens\end{tabular} &
  Positive \\ \hline
$d_{i1}$ &
  \begin{tabular}[c]{@{}l@{}}The distance between the attack lens\\ and the formed lens image\end{tabular} &
  \begin{tabular}[c]{@{}l@{}}Virtual image: positive;\\ real image: negative\end{tabular} \\ \hline
$m_1$ &
  \begin{tabular}[c]{@{}l@{}}Magnification of the formed attack lens\\ image compared to the real object size\end{tabular} &
  \begin{tabular}[c]{@{}l@{}}Virtual image: positive;\\ real image: negative\end{tabular} \\ \hline
$d_{i2}$ &
  \begin{tabular}[c]{@{}l@{}}The distance between the attack lens\\ and the formed camera image\end{tabular} &
  \begin{tabular}[c]{@{}l@{}}Virtual image: positive;\\ real image: negative\end{tabular} \\ \hline
$m_2$ &
  \begin{tabular}[c]{@{}l@{}}Magnification of the formed camera\\ image compared to the attack lens image\end{tabular} &
  \begin{tabular}[c]{@{}l@{}}Virtual image: positive;\\ real image: negative\end{tabular} \\ \hline
$f_c$ &
  The focal length of the camera lens &
  Positive \\ \hline
$d_b$ &
  \begin{tabular}[c]{@{}l@{}}The distance between the attack lens\\ and the camera lens\end{tabular} &
  Positive \\ \hline
$m_{total}$ &
  \begin{tabular}[c]{@{}l@{}}The total magnification of the formed camera\\ image compared to the real object size\end{tabular} &
  \begin{tabular}[c]{@{}l@{}}Virtual image: positive;\\ real image: negative\end{tabular} \\ \hline
$m_{ori}$ &
  \begin{tabular}[c]{@{}l@{}}Magnification of the formed camera image\\ compared to the real object size without \\ applying the attack lens\end{tabular} &
  \begin{tabular}[c]{@{}l@{}}Virtual image: positive;\\ real image: negative\end{tabular} \\ \hline
\end{tabular}%
\label{tab:para_list}
}
\end{table}

\attack exploits the vulnerabilities in the monocular depth perception. The basic idea is to apply the attack lenses in front of the camera lens to form a combination lens system, which will enlarge or reduce the size of the sensed images. 
We propose four different lens combinations based on convex and concave lens attacks. As a result, it causes the depth change of the objects. 

Fig.~\ref{fig:attack_pipline} shows the attack pipeline: (1) we calculate the attack depth based on the optical principles of concave lenses and convex lenses using the available attack lens and victim camera; (2) we define the targeted and untargeted attacks; (3) we optimize our attacks by considering blurriness to derive the optimal focal length; and (4) we evaluate our attack performance in four different aspects. The parameters used in this section along with their respective meanings are summarized in Table~\ref{tab:para_list}.


\subsection{Concave Lens Attack}
\begin{figure*}[t]
    \centering
    \subfigure[]{\includegraphics[width=0.21\textwidth]{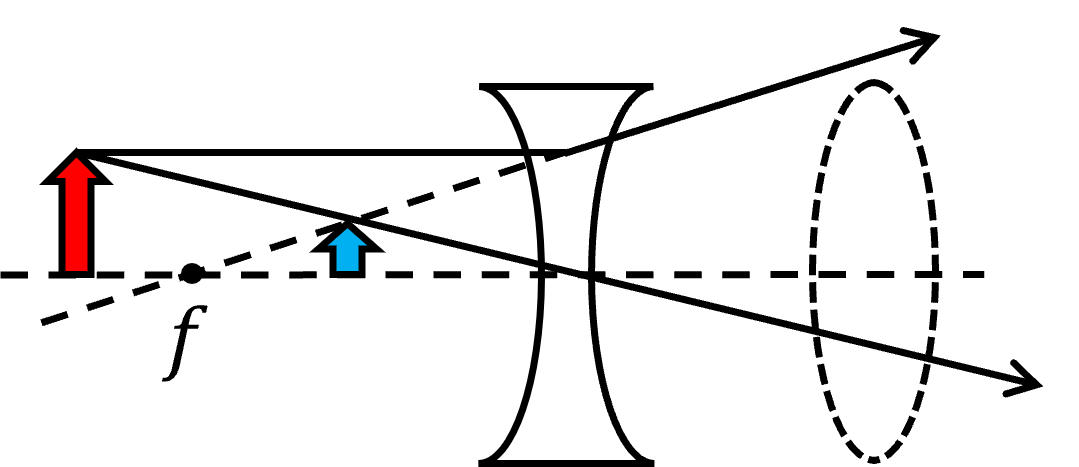}} 
    \subfigure[]{\includegraphics[width=0.23\textwidth]{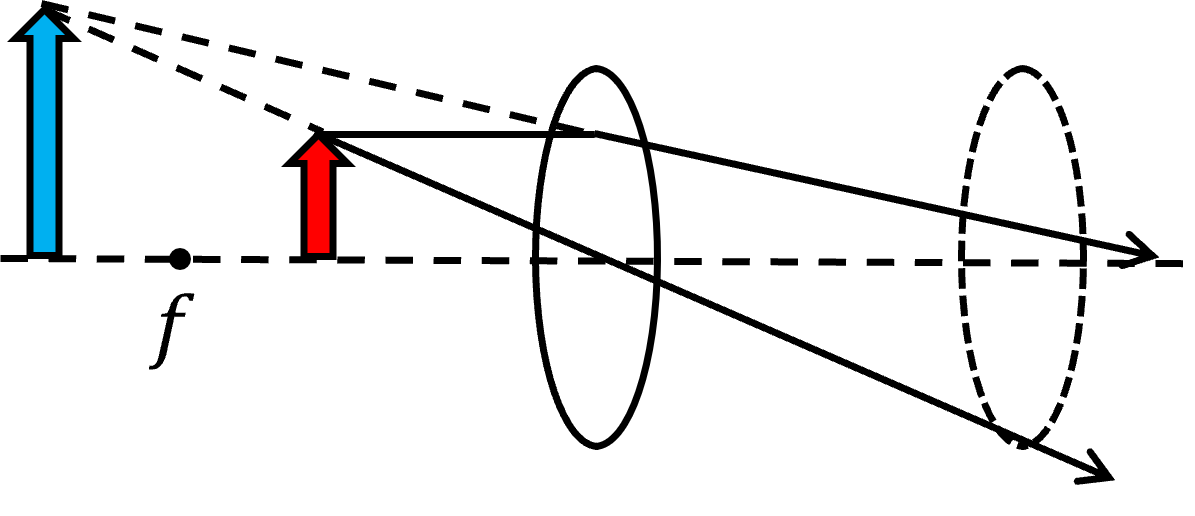}} 
    \subfigure[]{\includegraphics[width=0.25\textwidth]{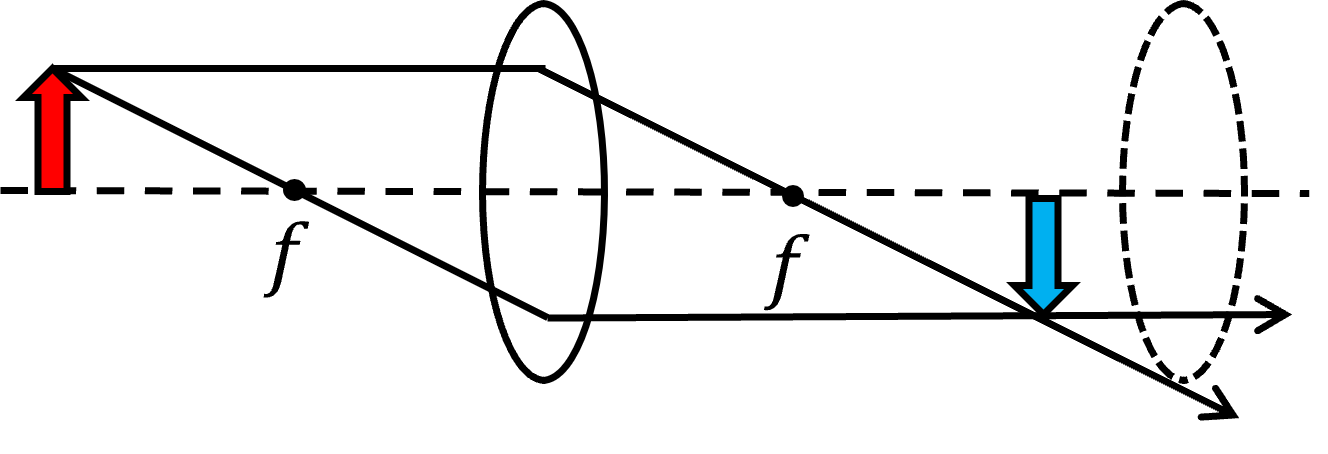}}
    \subfigure[]{\includegraphics[width=0.25\textwidth]{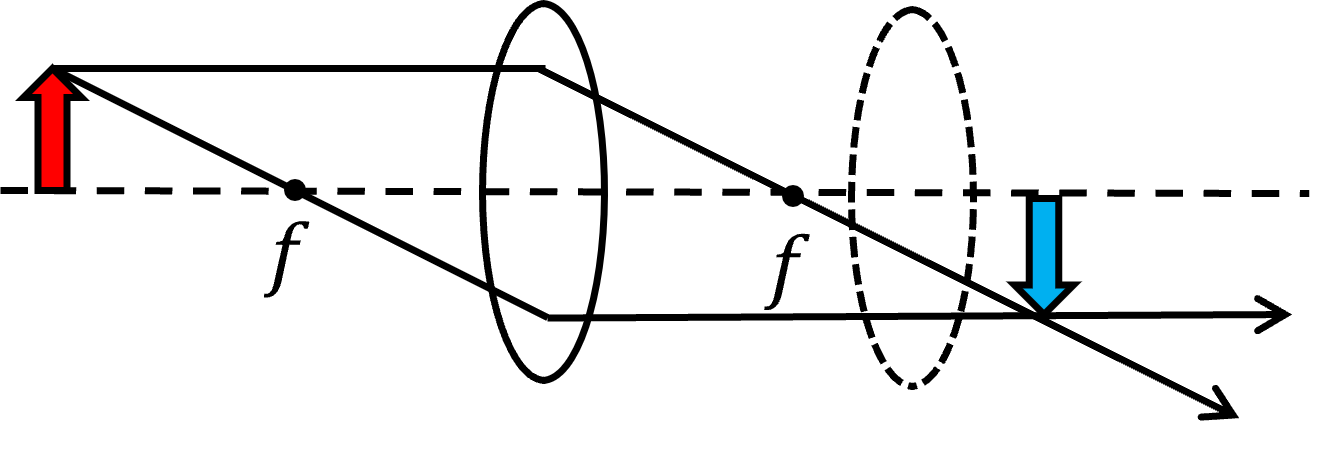}}
    \caption{Ray diagrams for (a) concave lens attack, and (b) first (c) second (d) third attack scenarios in convex lens attack. The left lens is the attack lens and the right dotted one stands for the monocular camera lens. The object is shown in red arrow and its corresponding image formed by the attack lens is shown in blue. $f$ represents the focal length of the attack lens.}
    \label{fig:att_design}
\end{figure*}

As discussed in Section~\ref{sec:background}, only one type of image is formed by the concave lens. The lens combination is shown in Fig.~\ref{fig:att_design}(a).
Based on Eq.~(\ref{equ1}), we have:
\begin{equation}\label{equ}
    \frac{1}{f}=\frac{1}{d_{o1}}+\frac{1}{d_{i1}},
\end{equation}
where $d_{o1}$ stands for the distance between the object and the attack lens, and $d_{i1}$ denotes the distance between the attack lens and the image.
Therefore, $d_{i1}$ can be written as:
\begin{equation}\label{equ}
    d_{i1}=-\frac{d_{o1} f}{d_{o1}-f}.
\end{equation}

Based on Eq.~(\ref{equ2}), we have the magnification $m_1$ for the attack lens as:
\begin{equation}\label{equ}
    m_1=-\frac{d_{i1}}{d_{o1}}=-\frac{f}{d_{o1}-f}.
\end{equation}

Similarly, for the camera lens, we have the distance between the lens and the image $d_{i2}$ and magnification $m_2$ as:
\begin{equation}\label{equ}
    d_{i2}=-\frac{(|d_{i1}|+d_b) f_c}{(|d_{i1}|+d_b)-f_c},
\end{equation}
and
\begin{equation}\label{equ}
    m_2=-\frac{d_{i2}}{d_{o2}}=-\frac{f_c}{(|d_{i1}|+d_b)-f_c},
\end{equation}
where $f_c$ is the focal length of the camera lens and $d_b$ is the distance between the attack lens and the camera lens. $d_b$ is always a positive number.

Thus, the total magnification $m_{total}$ can be expressed as:
\begin{equation}\label{equ:concave}
    m_{total}=m_1m_2=\frac{ff_c}{(d_{o1}-f)(|\frac{d_{o1}f}{d_{o1}-f}|+d_b-f_c)}.
\end{equation}

In the vehicle's camera, the focal length is rather small. The object's distance (i.e., the distance between the object and the camera) is usually much larger than twice the focal length, so the image is always formed at the focal point (i.e., the location of the image sensor) regardless of the object's distance. Therefore, the size of the image formed on the image sensor is inversely proportional to the object's distance. The larger the image is formed, the smaller the object's distance can be. Since the original magnification $m_{ori}$ is:
\begin{equation}\label{equ:magnification}
    m_{ori}=-\frac{d_{i}}{d_{o}}=-\frac{f_c}{d_{o1}+d_b-f_c},
\end{equation}
the manipulated depth becomes $|m_{ori}/m_{total}|$ of the original depth.

\subsection{Convex Lens Attack}
There are three attack cases based on the different distances between the object and the lens (Figs.~\ref{fig:att_design}(b)(c)(d)). If $0<d_{o1}<f$ (\textit{as in the first attack scenario}), we can place the camera anywhere we desire. It has the same mathematical analysis as in the concave lens attack. On the other hand, if $d_{o1} > f$, $d_b$ may be further away than the image formed by the attack lens, indicating $d_b\geq d_{i1}$ (\textit{as in the second attack scenario}). Or, it could be closer than the image's location formed by the attack lens, indicating $d_b\leq d_{i1}$ (\textit{as in the third attack scenario}). 

Following the same derivation procedure as the convex lens attack, we have the same expression of $d_{i1}$ and $m_1$ for the attack lens. For the second attack scenario ($d_b\geq d_{i1}$), $d_{i2}$ and $m_2$ for the camera lens can be expressed as:
\begin{equation}\label{equ}
    d_{i2}=-\frac{(d_b-|d_{i1}|) f_c}{(d_b-|d_{i1}|)-f_c},
\end{equation}
\begin{equation}\label{equ}
    m_2=-\frac{d_{i2}}{d_{o2}}=-\frac{f_c}{(d_b-|d_{i1}|)-f_c}.
\end{equation}
Thus, the total magnification $m_{total}$ is:
\begin{equation}\label{equ}
    m_{total}=m_1m_2=\frac{ff_c}{(d_{o1}-f)(d_b-|\frac{d_{o1}f}{d_{o1}-f}|-f_c)}.
\end{equation}
For the third attack scenario ($d_b\leq d_{i1}$), $d_{i2}$ and $m_2$ for the camera lens can be expressed as:
\begin{equation}\label{equ}
    d_{i2}=-\frac{(|d_{i1}|-d_b) f_c}{(|d_{i1}|-d_b)-f_c},
\end{equation}
\begin{equation}\label{equ}
    m_2=-\frac{d_{i2}}{d_{o2}}=-\frac{f_c}{(|d_{i1}|-d_b)-f_c}.
\end{equation}
Thus, the total magnification $m_{total}$ is: 
\begin{equation}\label{equ:convex}
    m_{total}=m_1m_2=\frac{ff_c}{(d_{o1}-f)(|\frac{d_{o1}f}{d_{o1}-f}|-d_b-f_c)}.
\end{equation}

In theory, all the attack scenarios should be feasible. However, due to the constraints in physical attacks, only the concave lens attack and the third scenario of the convex lens attack would lead to a successful attack. More details are presented in Sections~\ref{sec:evl} and ~\ref{sec:discussion}.

\subsection{Attack Optimization}\label{sec:optimization}

\subsubsection{Factor of Blurriness}

A photographic lens that does not possess adjustable focus capability is referred to as a fixed-focus lens~\cite{fixed_focus_wiki}.
Typically, advanced driver-assistance system (ADAS), drone, and AD cameras are fixed-focus~\cite{wittpahl2018realistic}, because it is best for handling high-vibration environments~\cite{fixed_focus}. When we add an optical lens in front of the camera, it will separate the whole image into the in-lens area and out-of-lens area if the lens is in the camera view. The focal length of the in-lens area will be the combination of the focal length from the optical focal length and the victim camera lens. The focal lens of the out-of-lens area is still the same as the benign one. 

Since the in-lens and out-of-lens have different focal lengths, blur usually will be added to the in-lens area due to the effect of depth of field (DOF). Note that DOF is defined as the distance between the closest and farthest objects in an image captured with a camera that is in acceptable sharp focus~\cite{Depthoffield}. On the other hand, the camera with autofocus (AF), a function of a camera to automatically focus on a subject (e.g., smartphone camera), can also be affected by DOF in our attack. Due to the focal length difference caused by the addition of the attack lens, the camera will either focus on the in-lens area or out-of-lens area, leading to blurriness in the part that is not in focus.

Blur in the image can affect the depth estimation result. Therefore, reducing the blurriness while maintaining accurate attack depth is critical for a successful attack. To improve the attack accuracy when partially attacking the object depth, we formulate the attacking process as an optimization problem to obtain the optimized attack parameters. We first simulate the model of the imaging process, and then we design the loss functions for targeted and untargeted attacks. Our goal is to maintain the depth unchanged in the out-of-lens area while modifying the target object depth as close as possible to the target depth (targeted) or as far away as possible from its original depth (untargeted). 

\subsubsection{Model of Imaging Process}
When we apply the attack lens inside the view of the camera, the image will be transformed into the in-lens part and the out-of-lens part. The in-lens area usually contains the attack target object, e.g., vehicles. For the concave lens attack, the blur is usually at the out-of-lens area,  whereas the blur often happens in the in-lens area when the convex lens attack is applied. 

Due to the different values of $d_b$, $d_o$ and $f$, the benign image $\boldsymbol{I}$ can be transformed into various attack images $\boldsymbol{I^{att}}$. The transformation not only shrinks or enlarges the in-lens scene but also adds a different portion of the blur to the in-lens or out-of-lens area in the image according to the type of attack lens. This can be formalized as:
\begin{equation}
    \boldsymbol{I^{att}}=\mathbb{T}(\boldsymbol{I},(d_b,d_o,f)),
\end{equation}
where $\mathbb{T}$ is the transformation function. To simplify the imaging process, we set $d_b$ and $d_o$ to constant values. To blur the image, we set each pixel to the average value of the pixels in a square box with radius pixels proportional to $f$ extending in each direction. For example, radius 1 takes 1 pixel in each direction resulting in 9 pixels in total.  


\subsubsection{Loss Function}

We propose a loss function for both the targeted and untargeted attacks, which consists of two components:
\begin{equation}
    L_{total} = (1-\alpha)L_{veh}+\alpha L_{out},
    \label{equ:loss}
\end{equation}
where $\alpha$ is weighting coefficients that are determined experimentally.  
The first component of the loss function is the target vehicle loss $L_{veh}$, and the second component is the out-of-lens loss $L_{out}$. 

For the targeted attack, $L_{veh}$ is used to minimize the target vehicle depth from the target depth, which is defined as: 
\begin{equation}
    L_{veh}= \min_{f}L(\boldsymbol{F}(\boldsymbol{I^{att}},\theta)\cdot M_{veh},y^{tar}),
\end{equation}
where $\boldsymbol{F}$ and $\theta$ are the target MDE neural network model and the corresponding network parameters, respectively. $M_{veh}$ serves as a mask to selectively retain the depth information solely for the target vehicle while excluding other portions of the image. The mask boundary is the bounding box output of the target object from the YOLO v8~\cite{Jocher_YOLO_by_Ultralytics_2023}. $y^{tar}$ is the target depth value of the target vehicle, and $L$ represents the $L_{1}$ loss function.

On the other hand, the untargeted attack requires maximizing the depth of the target vehicle from its true depth value. $L_{veh}$ is given as:
\begin{equation}
    L_{veh}= -\max_{f}L(\boldsymbol{F}(\boldsymbol{I^{att}},\theta)\cdot M_{veh},\boldsymbol{F}(\boldsymbol{I},\theta)\cdot M_{veh}),
\end{equation}
where the minus sign indicates the minimization of the loss function.

The out-of-lens loss, $L_{out}$, minimizes the difference between the attacked out-of-lens area with the benign one, which is defined as: 
\begin{equation}
   L_{out}=\min_{f}L(\boldsymbol{F}(\boldsymbol{I^{att}},\theta)\cdot M_{out},\boldsymbol{F}(\boldsymbol{I},\theta)\cdot M_{out}),
\end{equation}
where $M_{out}$ functions as a mask designed to exclude the in-lens area, while preserving the out-of-lens area.

The objective is to minimize the combined total of these two loss functions, $L_{total}$, to determine the optimal $f$ to enhance the attack accuracy. 
The detailed implementation is in Section~\ref{Attack Optimization Implementation Details}.

\section{Evaluation} \label{sec:evl}

To demonstrate the impact of \attack, we first simulate it in the digital world, and then we evaluate its performance in the physical world. 
We further demonstrate that the optimization of the attack enhances the attack performance. We finally conduct an end-to-end system-level simulation to showcase the potential impact of our attack in the real world. 

\subsection{Attack Setup}\label{sec:attack_setup}

\subsubsection{Target Models}
To evaluate the accuracy and effectiveness of our attack, we consider three state-of-the-art MDE algorithms,  Monodepth2~\cite{monodepth2},  Depth Hints~\cite{watson2019self}, and Lite-mono~\cite{zhang2023lite} as the attack targets. They are stereo and video-based, stereo-based, and video-based algorithms, respectively. The selection is based on their representativeness, timeliness, and popularity. The output of the Monodepth2 can be a depth map or disparity map, whereas the output of Depth Hints and Lite-mono is only a disparity map. The disparity map can be converted into the depth map if it is trained on the stereo pairs and the baseline and focal length are known. The relation between the disparity and depth is as follows: 
\begin{equation}
    disparity = \frac{(baseline*focal\:length)}{depth}.
\end{equation}
Note that we only examine the attack performance on Monodepth2 when the evaluation requires the real depth value.

\subsubsection{Evaluation Metrics}
We adopt Attack Distortion Rate (ADR) and Attack Error Rate (AER) as the evaluation metrics to show the performance of the attack in both the digital world and the physical world. The ADR and AER are defined as follows:
\begin{equation}
    ADR = \frac{|attacked\:depth-benign\:depth|}{benign\:depth},
\end{equation}
and 
\begin{equation}
    AER = \frac{|attacked\:depth-target\:depth|}{target\:depth}.
\end{equation}
A higher ADR implies a greater depth difference caused by the attack. Conversely, a lower AER indicates a smaller deviation from the target or expected depth value, indicating a higher attack success rate. 



\subsubsection{Dataset in Simulation}
All three target models are trained and evaluated on the KITTI dataset~\cite{geiger2013vision}. Therefore, in the attack simulation, we use some images in KITTI semantic split~\cite{abu2018augmented}. We assume that the attack area can either be part of the image or the full image. In the former case, the attack area will be circular which is the same shape as the physical attack lens. We apply different ratios to enlarge or reduce the size of the attack area, and then we compare the attack results in the disparity map.

 \begin{figure*}
    \centering    
    \subfigure[Full view of the attack]{\includegraphics[width=0.245\textwidth]{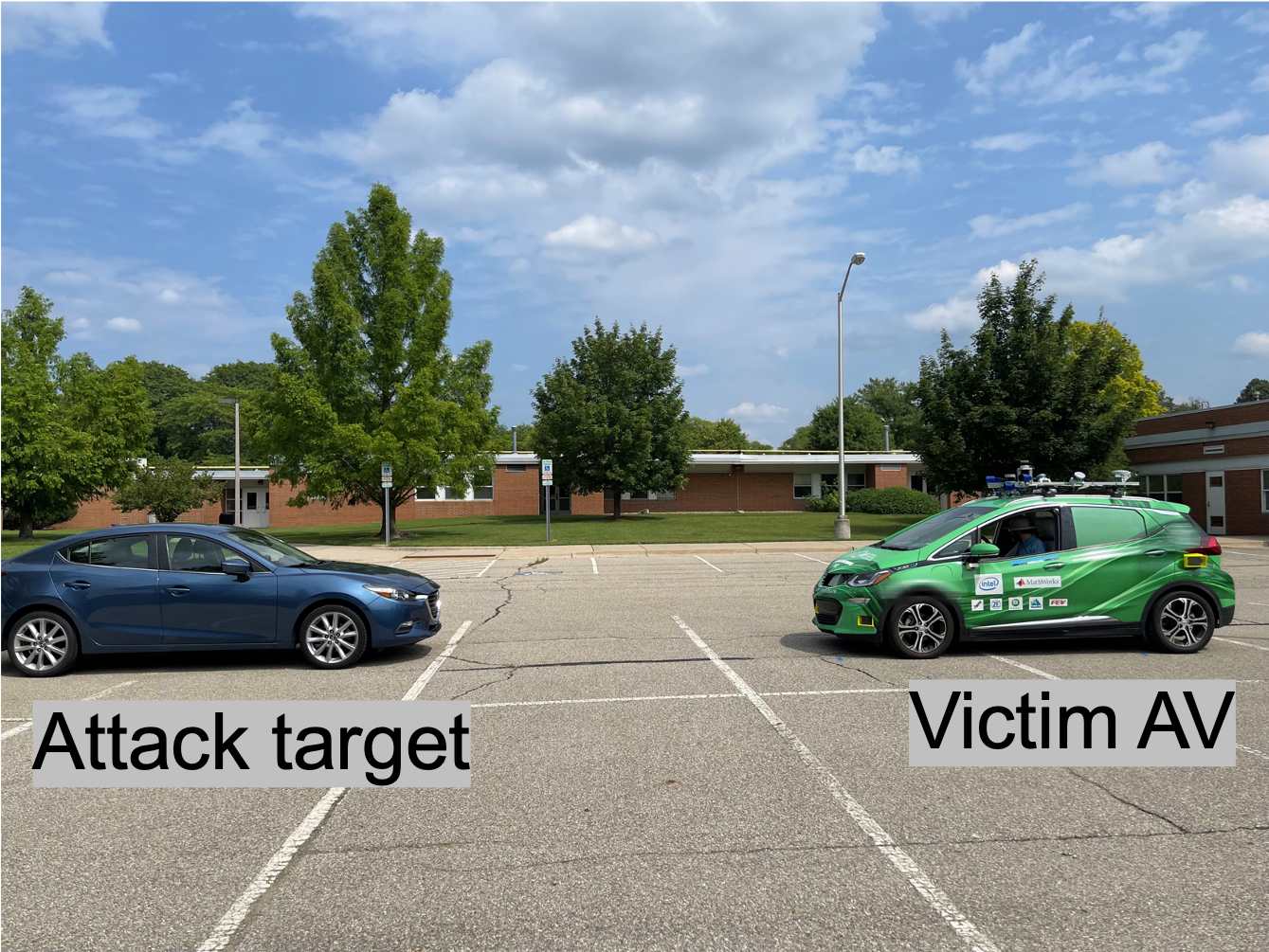}}    
    \subfigure[Data collection on AV]{\includegraphics[width=0.255\textwidth]{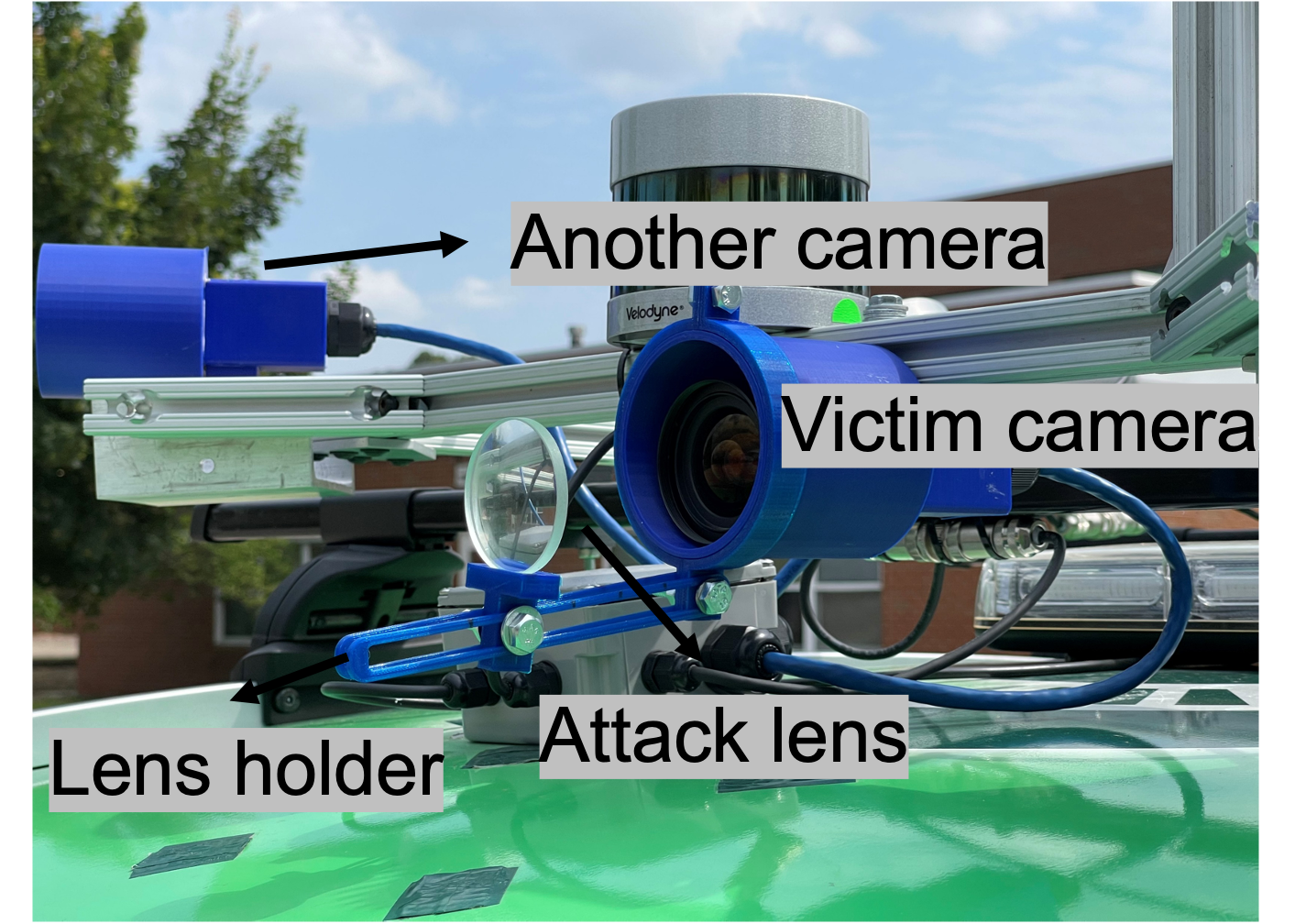}}
    \subfigure[3D-printed lens holder]{\includegraphics[width=0.255\textwidth]{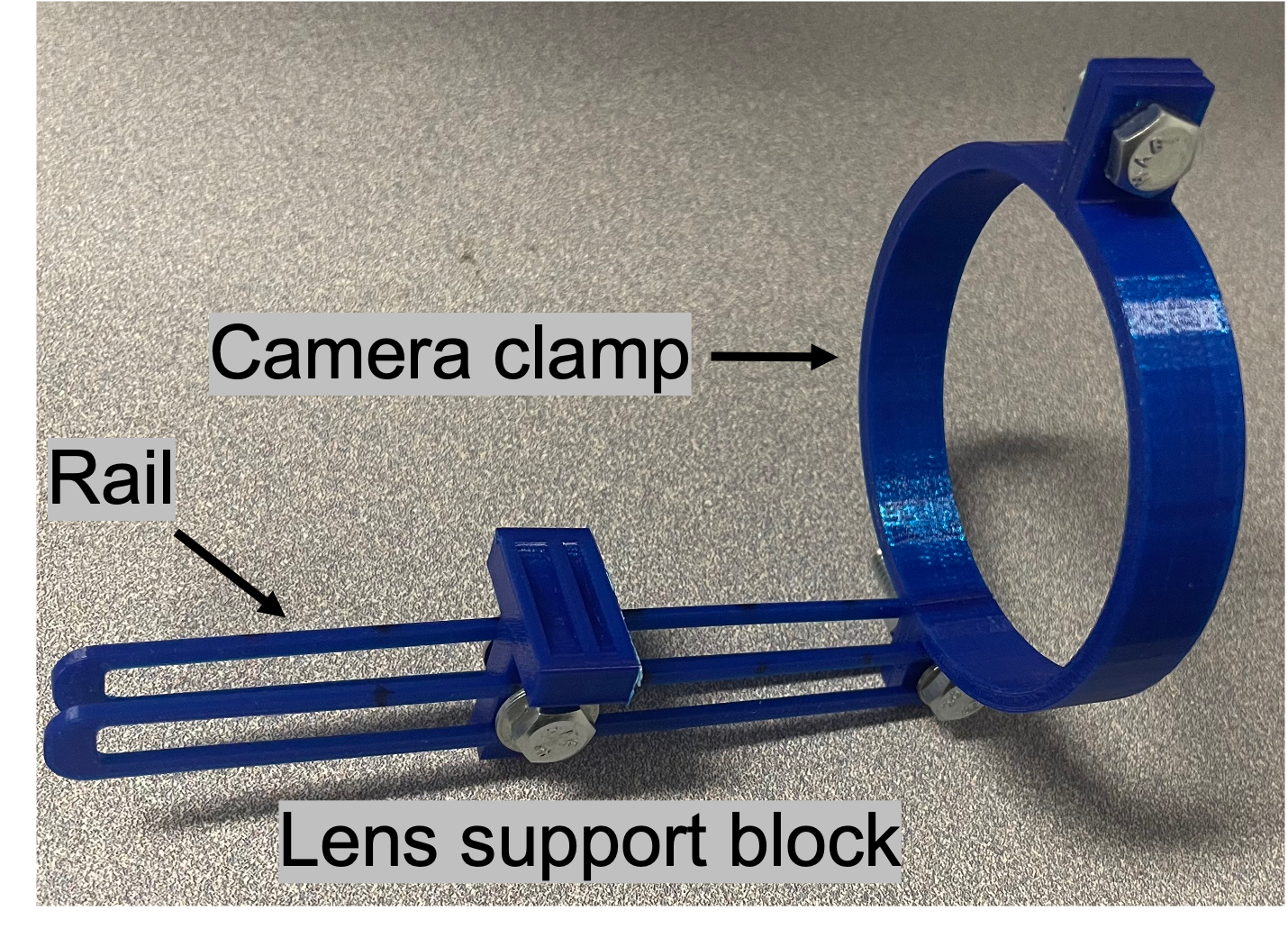}\label{fig:5c}} 
    \subfigure[Data collection in driving scenario]{\includegraphics[width=0.217\textwidth]{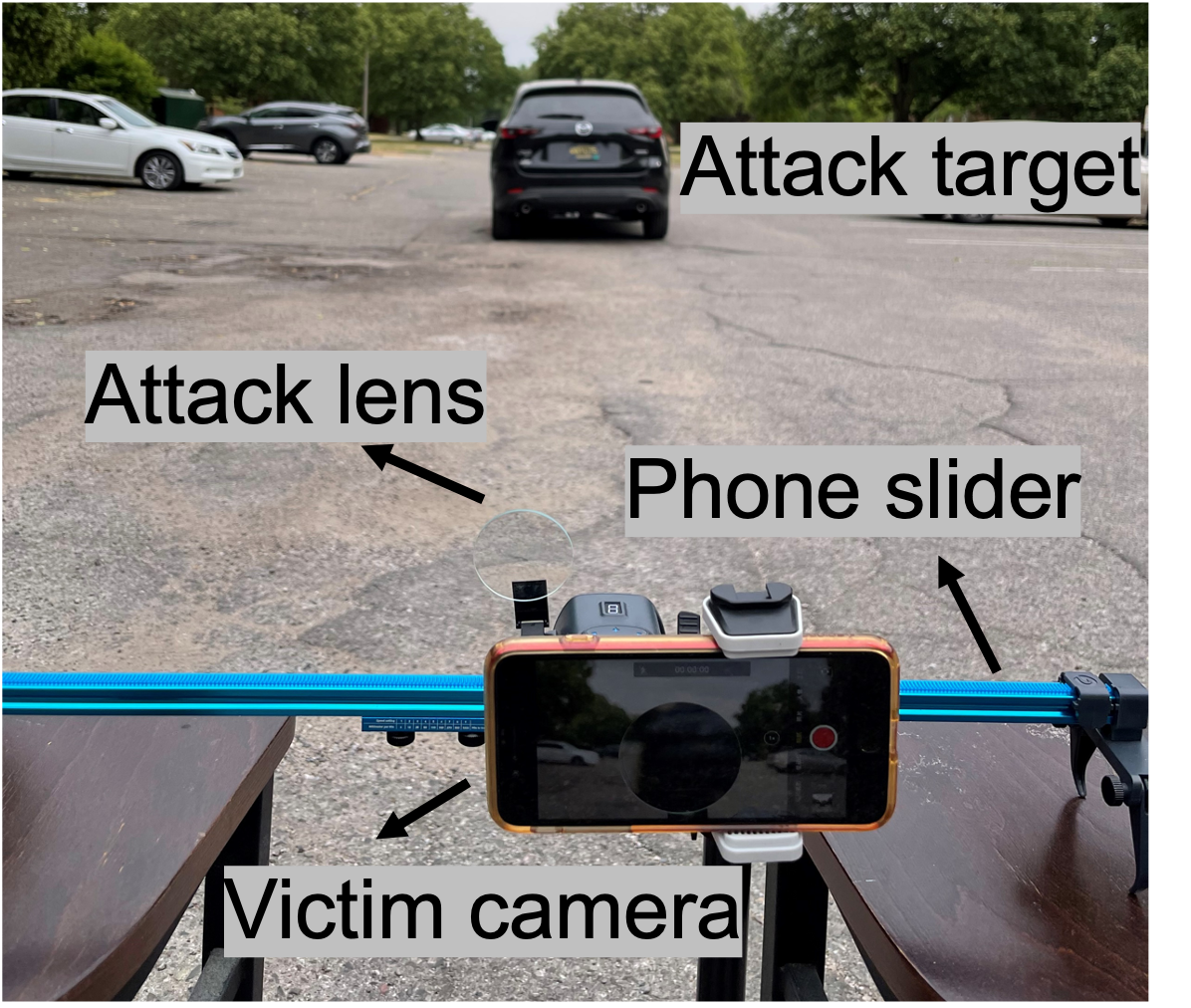}}
    \caption{Experimental setup for the AV and real-world driving.}
    \label{fig:experimental_setup}
\end{figure*}

\subsubsection{Physical Experimental Setup}
The human eye has a focal length of approximately $22mm$~\cite{cambridgeincolour}. Car cameras usually have a similar focal length as the human eye's. A longer focal length leads to a narrower field of view. As for AD vehicles, a shorter focal length provides a broader field of view. Besides, we find that the commonly available optical lenses in the market without particular order or design are usually within the focal length of $50mm$ to $500mm$.

In the physical experiments, we test our attack on an AV camera and a smartphone camera. The experimental setup is shown in Fig.~\ref{fig:experimental_setup}. The victim AV is a 2017 Chevy Bolt Electric Vehicle with three FLIR BFS-PGE-31S4C cameras mounted on the car's roof, which will be called ``AV'' for simplicity. The focal length of the FLIR camera with lens is approximately $23mm$. To collect the attack images in real life, we design a lens holder that is 3D printable as shown in Fig.~\ref{fig:5c}. It contains a camera clamp, a rail, and a lens support block. The lens support block can hold the lens and slide on the rail so that we can adjust the distance between the attack lens and the camera lens ($d_b$). The lens holder can be positioned externally to the FLIR camera such that the victim AV can collect images that contain the target vehicle. 

In real driving scenarios (see Fig.~\ref{fig:experimental_setup}(d)), we collect images of driving scenes using iPhone 12 Pro Max, whose focal length is $26mm$ \cite{dxomark}. We will call it ``iPhone'' for the rest of the content. Since the focal lengths of the FLIR camera and the iPhone camera are almost the same, we will take $26mm$ for the expected attack depth calculation. Two sets of concave and convex lenses~\cite{AmlongCrystal} are used in the physical attacks with focal lengths of $20cm$, $30cm$, and $50cm$ in each set. To compare the attack results, we use various values of $f$, $d_b$, and $d_o$. For the real-world physical attack, we have the attack setup inside the victim's vehicle. We use a phone slider to hold the attack lens so that we can remotely control the position of the attack lens using a wireless controller, allowing for a more flexible attack. We have also integrated YOLO v8 with the MDE algorithm in the real-world driving scenario. Physical attack video demos are available on our website.

\subsubsection{Attack Optimization Implementation Details}\label{Attack Optimization Implementation Details}

Since the blurriness in the simulation corresponding to the focal length is usually hard to match as it is in the physical world, we evaluate our attack optimization in the digital world. Therefore, rather than using the real focal length, we discretize the focal length and generate the blurriness using a self-defined proportion.

The lens induces proportional changes in the in-lens area and the out-of-lens area, affecting all pixels, based on the value of focal length. Besides, our attack is a black-box attack, which means that we only know the attack parameter inputs and the corresponding depth estimation outputs. Therefore, we use a brute force method to identify the value of $f$ to minimize the loss function.

Regarding attack practicality in real attack scenarios, we restrict the number of available focal lengths. We assume that the number of possible focal lengths is 9, and assign them the discrete value from 1 to 9. When $f=1$, the least blurriness is added, whereas the maximized blurriness is added when $f=9$. Moreover, we perform our attack on the disparity map on all three target models. We set the $y_{tar}=0.43$ for the concave attack and $y_{tar}=0.60$ for the convex attack. Note that the disparity of Lite-mono is not the same scale as the other two models, therefore, we convert it to the same scale disparity as others by dividing it with a constant value of 5.4.

\subsection{Attack Simulation in Digital World} 

The goal of \attack is to modify the target object size on an image such that the estimated depth can be manipulated. We consider the lens can be applied to either the entire image or a portion of it. We simulate our attack in three attack scenarios: full image cropping (or full image enlarging), partial image enlarging, and partial image shrinking. Note that full image shrinking is just the reverse case of full image enlarging, so we do not consider it separately. 

\begin{figure}[H]
\centering
	\includegraphics[width=0.6\columnwidth]{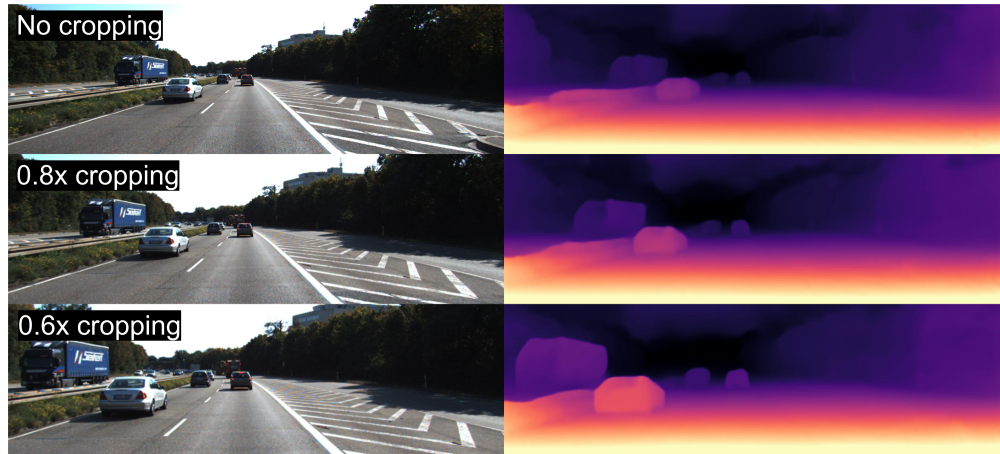}
	\caption{Full image cropping with 0.8x and 0.6x cropping ratios.}
	\label{fig:cropping}
\end{figure}

\begin{figure}[H]
\centering
	\includegraphics[width=0.6\columnwidth]{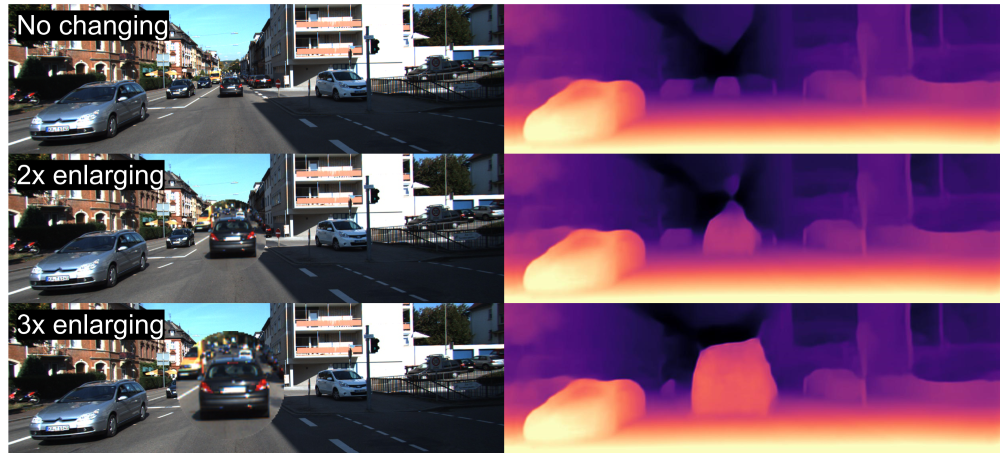}
	\caption{Partial image enlarging with 2x and 3x enlarging ratios.}
	\label{fig:enlarging}
\end{figure}

\begin{figure}[H]
\centering
	\includegraphics[width=0.6\columnwidth]{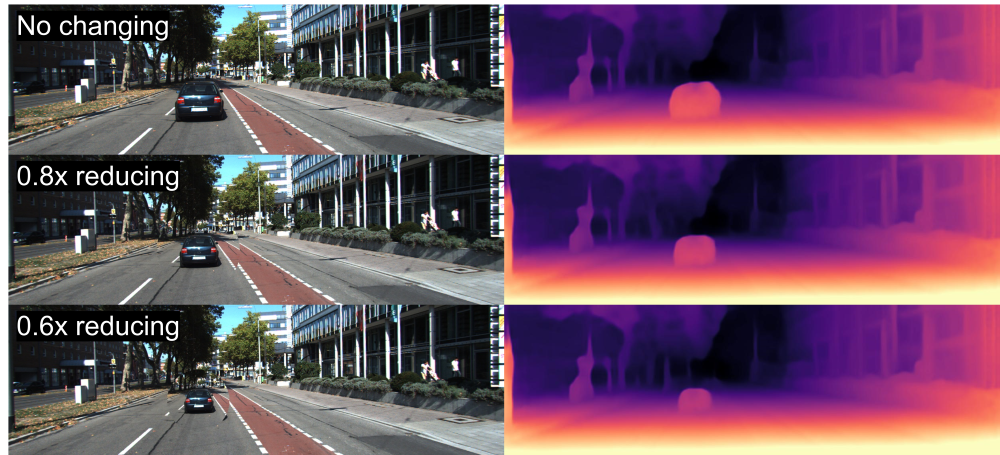}
	\caption{Partial image reducing with 0.8x and 0.6x reducing ratios.}
	\label{fig:reducing}
\end{figure}

To show the concept and feasibility of our attack,
the attack simulation is conducted on all three MDE algorithms without the consideration of blurriness. The attack results of full image cropping on Monodepth2 are shown in Fig.~\ref{fig:cropping}. The simulated images are shown in the left column, and the corresponding disparity maps are displayed in the right column. When the attack lens is applied to the whole image, we can see that the more we crop the image, the more the depth can be altered. Similarly, as shown in Figs.~\ref{fig:enlarging} and ~\ref{fig:reducing}, when the attack lens is applied to the partial image, the more we enlarge or shrink the attack area, the more the depth can be changed. 

\begin{table*}[]
\centering
\caption{Masked mean object depth and the corresponding ADR in the attack simulation with different image modifications.}
\label{tab:simulation_results}
\resizebox{0.9\textwidth}{!}{%
\begin{tabular}{|cc|c|ccc|ccc|}
\hline
\multicolumn{2}{|c|}{\multirow{2}{*}{Image Modification}} &
  \multirow{2}{*}{\begin{tabular}[c]{@{}c@{}}Masked Object Mean Depth Value \\ in Monodepth2 (meters)\end{tabular}} &
  \multicolumn{3}{c|}{Masked Object Mean Disparity Value} &
  \multicolumn{3}{c|}{ADR} \\ \cline{4-9} 
\multicolumn{2}{|c|}{} &
   &
  \multicolumn{1}{c|}{Monodepth2} &
  \multicolumn{1}{c|}{Depth Hints} &
  Lite-Mono &
  \multicolumn{1}{c|}{Monodepth2} &
  \multicolumn{1}{c|}{Depth Hints} &
  Lite-Mono \\ \hline
\multicolumn{1}{|c|}{\multirow{3}{*}{Cropping}} &
  Benign &
  21.08 &
  \multicolumn{1}{c|}{0.28} &
  \multicolumn{1}{c|}{0.31} &
  1.89 &
  \multicolumn{1}{c|}{-} &
  \multicolumn{1}{c|}{-} &
  - \\ \cline{2-9} 
\multicolumn{1}{|c|}{} &
  0.8x &
  16.44 &
  \multicolumn{1}{c|}{0.36} &
  \multicolumn{1}{c|}{0.37} &
  2.19 &
  \multicolumn{1}{c|}{28.6\%} &
  \multicolumn{1}{c|}{19.3\%} &
  15.9\% \\ \cline{2-9} 
\multicolumn{1}{|c|}{} &
  0.6x &
  14.00 &
  \multicolumn{1}{c|}{0.45} &
  \multicolumn{1}{c|}{0.46} &
  2.66 &
  \multicolumn{1}{c|}{60.7\%} &
  \multicolumn{1}{c|}{48.4\%} &
  40.7\% \\ \hline
\multicolumn{1}{|c|}{\multirow{3}{*}{Enlarging}} &
  Benign &
  24.66 &
  \multicolumn{1}{c|}{0.23} &
  \multicolumn{1}{c|}{0.23} &
  1.47 &
  \multicolumn{1}{c|}{-} &
  \multicolumn{1}{c|}{-} &
  - \\ \cline{2-9} 
\multicolumn{1}{|c|}{} &
  2x &
  17.29 &
  \multicolumn{1}{c|}{0.36} &
  \multicolumn{1}{c|}{0.36} &
  2.18 &
  \multicolumn{1}{c|}{56.5\%} &
  \multicolumn{1}{c|}{56.5\%} &
  48.3\% \\ \cline{2-9} 
\multicolumn{1}{|c|}{} &
  3x &
  11.37 &
  \multicolumn{1}{c|}{0.50} &
  \multicolumn{1}{c|}{0.52} &
  3.03 &
  \multicolumn{1}{c|}{117.4\%} &
  \multicolumn{1}{c|}{126.1\%} &
  106.1\% \\ \hline
\multicolumn{1}{|c|}{\multirow{3}{*}{Shrinking}} &
  Benign &
  13.97 &
  \multicolumn{1}{c|}{0.44} &
  \multicolumn{1}{c|}{0.46} &
  2.68 &
  \multicolumn{1}{c|}{-} &
  \multicolumn{1}{c|}{-} &
  - \\ \cline{2-9} 
\multicolumn{1}{|c|}{} &
  0.8x &
  14.30 &
  \multicolumn{1}{c|}{0.40} &
  \multicolumn{1}{c|}{0.41} &
  2.42 &
  \multicolumn{1}{c|}{9\%} &
  \multicolumn{1}{c|}{10.9\%} &
  9.7\% \\ \cline{2-9} 
\multicolumn{1}{|c|}{} &
  0.6x &
  14.78 &
  \multicolumn{1}{c|}{0.38} &
  \multicolumn{1}{c|}{0.38} &
  2.24 &
  \multicolumn{1}{c|}{13.6\%} &
  \multicolumn{1}{c|}{17.4\%} &
  16.4\% \\ \hline
\end{tabular}%
}
\end{table*}

Meanwhile, we also mask the target object using the object detection results from YOLO v8, and compute its mean depth value in the masked area which represents the average distance of the target object to the camera. In Table~\ref{tab:simulation_results}, we show the mean depth value of masked objects in Monodepth2 in meters, the mean disparity value of masked objects in all target models, and their corresponding ADRs. It can be observed that partially enlarging the target object changes the depth the most. With $3x$ enlarging, the depth of the target vehicle can be reduced by around $13m$ with the corresponding ADR over 100\% for all three depth estimation algorithms. Then, the full image cropping comes second regarding the extent of manipulating the depth of the target vehicle. 
Shrinking the size of the target vehicle can only change the depth up to $1m$ with  $0.6x$ image shrinking. Therefore, the results show that different image modifications can cause the depth of the target object to vary. 


\cbox{To summarize, \attack has caused similar distortion effect 
across all three target models, demonstrating the generalization potential of the attack.}



\subsection{Real-World Physical Attack}\label{sec:Real-World Physical Attack}

To further investigate the practicality and generality, we launch \attack in the physical world. We perform both concave lens attacks and convex lens attacks with various object distances, i.e., $6m$, $9m$, and $12m$. 

\subsubsection{Performance of Concave Lens Attacks}

For the concave lens attack, we investigate the effects of $f$ and $d_b$ in altering the target object depth with different $d_{o1}$.
The results in Table~\ref{tab:attack_error_rate_concave} indicate a smaller $f$ leads to a larger change in the object depth. It also shows that smaller $d_b$ results in less impact on the depth. 

\begin{table*}[]
\centering
\caption{The AER of the concave lens attack with the various object distance $d_{o1}$, $f$, and $d_b$.}
\label{tab:attack_error_rate_concave}
\resizebox{\textwidth}{!}{%
\begin{tabular}{|cc|ccccc|ccccc|ccccc|}
\hline
\multicolumn{2}{|c|}{\multirow{2}{*}{Attack Parameters}} &
  \multicolumn{5}{c|}{$d_{o1}$=6m} &
  \multicolumn{5}{c|}{$d_{o1}$=9m} &
  \multicolumn{5}{c|}{$d_{o1}$=12m} \\ \cline{3-17} 
\multicolumn{2}{|c|}{} &
  \multicolumn{1}{c|}{\multirow{2}{*}{\begin{tabular}[c]{@{}c@{}}Expected\\ Depth (meters)\end{tabular}}} &
  \multicolumn{2}{c|}{\begin{tabular}[c]{@{}c@{}}Experimental Depth\\ (meters)\end{tabular}} &
  \multicolumn{2}{c|}{\begin{tabular}[c]{@{}c@{}}AER\\ (\%)\end{tabular}} &
  \multicolumn{1}{c|}{\multirow{2}{*}{\begin{tabular}[c]{@{}c@{}}Expected\\ Depth (meters)\end{tabular}}} &
  \multicolumn{2}{c|}{\begin{tabular}[c]{@{}c@{}}Experimental Depth\\ (meters)\end{tabular}} &
  \multicolumn{2}{c|}{\begin{tabular}[c]{@{}c@{}}AER\\ (\%)\end{tabular}} &
  \multicolumn{1}{c|}{\multirow{2}{*}{\begin{tabular}[c]{@{}c@{}}Expected\\ Depth (meters)\end{tabular}}} &
  \multicolumn{2}{c|}{\begin{tabular}[c]{@{}c@{}}Experimental Depth\\ (meters)\end{tabular}} &
  \multicolumn{2}{c|}{\begin{tabular}[c]{@{}c@{}}AER\\ (\%)\end{tabular}} \\ \cline{1-2} \cline{4-7} \cline{9-12} \cline{14-17} 
\multicolumn{1}{|c|}{$f$} &
  $d_b$ &
  \multicolumn{1}{c|}{} &
  \multicolumn{1}{c|}{AV} &
  \multicolumn{1}{c|}{iPhone} &
  \multicolumn{1}{c|}{AV} &
  iPhone &
  \multicolumn{1}{c|}{} &
  \multicolumn{1}{c|}{AV} &
  \multicolumn{1}{c|}{iPhone} &
  \multicolumn{1}{c|}{AV} &
  iPhone &
  \multicolumn{1}{c|}{} &
  \multicolumn{1}{c|}{AV} &
  \multicolumn{1}{c|}{iPhone} &
  \multicolumn{1}{c|}{AV} &
  iPhone \\ \hline
\multicolumn{1}{|c|}{20cm} &
  2cm &
  \multicolumn{1}{c|}{5.82} &
  \multicolumn{1}{c|}{-} &
  \multicolumn{1}{c|}{7.09} &
  \multicolumn{1}{c|}{-} &
  21.9 &
  \multicolumn{1}{c|}{8.73} &
  \multicolumn{1}{c|}{-} &
  \multicolumn{1}{c|}{7.67} &
  \multicolumn{1}{c|}{-} &
  12.12 &
  \multicolumn{1}{c|}{11.64} &
  \multicolumn{1}{c|}{-} &
  \multicolumn{1}{c|}{10.62} &
  \multicolumn{1}{c|}{-} &
  8.76 \\ \hline
\multicolumn{1}{|c|}{20cm} &
  4cm &
  \multicolumn{1}{c|}{6.42} &
  \multicolumn{1}{c|}{5.85} &
  \multicolumn{1}{c|}{6.85} &
  \multicolumn{1}{c|}{9.10} &
  6.77 &
  \multicolumn{1}{c|}{9.63} &
  \multicolumn{1}{c|}{9.82} &
  \multicolumn{1}{c|}{10.54} &
  \multicolumn{1}{c|}{1.97} &
  9.44 &
  \multicolumn{1}{c|}{12.84} &
  \multicolumn{1}{c|}{11.82} &
  \multicolumn{1}{c|}{12.46} &
  \multicolumn{1}{c|}{7.91} &
  2.92 \\ \hline
\multicolumn{1}{|c|}{20cm} &
  8cm &
  \multicolumn{1}{c|}{7.61} &
  \multicolumn{1}{c|}{7.73} &
  \multicolumn{1}{c|}{6.00} &
  \multicolumn{1}{c|}{1.59} &
  21.13 &
  \multicolumn{1}{c|}{11.42} &
  \multicolumn{1}{c|}{10.89} &
  \multicolumn{1}{c|}{12.65} &
  \multicolumn{1}{c|}{4.62} &
  10.80 &
  \multicolumn{1}{c|}{15.23} &
  \multicolumn{1}{c|}{12.86} &
  \multicolumn{1}{c|}{13.58} &
  \multicolumn{1}{c|}{15.55} &
  10.81 \\ \hline
\multicolumn{1}{|c|}{20cm} &
  12cm &
  \multicolumn{1}{c|}{8.78} &
  \multicolumn{1}{c|}{5.13} &
  \multicolumn{1}{c|}{6.89} &
  \multicolumn{1}{c|}{41.56} &
  21.46 &
  \multicolumn{1}{c|}{13.19} &
  \multicolumn{1}{c|}{10.56} &
  \multicolumn{1}{c|}{13.62} &
  \multicolumn{1}{c|}{19.93} &
  3.26 &
  \multicolumn{1}{c|}{17.60} &
  \multicolumn{1}{c|}{12.26} &
  \multicolumn{1}{c|}{16.73} &
  \multicolumn{1}{c|}{30.32} &
  4.90 \\ \hline
\multicolumn{1}{|c|}{30cm} &
  2cm &
  \multicolumn{1}{c|}{5.88} &
  \multicolumn{1}{c|}{-} &
  \multicolumn{1}{c|}{7.05} &
  \multicolumn{1}{c|}{-} &
  19.82 &
  \multicolumn{1}{c|}{8.82} &
  \multicolumn{1}{c|}{-} &
  \multicolumn{1}{c|}{8.37} &
  \multicolumn{1}{c|}{-} &
  5.09 &
  \multicolumn{1}{c|}{11.76} &
  \multicolumn{1}{c|}{-} &
  \multicolumn{1}{c|}{12.28} &
  \multicolumn{1}{c|}{-} &
  4.40 \\ \hline
\multicolumn{1}{|c|}{30cm} &
  4cm &
  \multicolumn{1}{c|}{6.28} &
  \multicolumn{1}{c|}{6.37} &
  \multicolumn{1}{c|}{7.08} &
  \multicolumn{1}{c|}{1.42} &
  12.78 &
  \multicolumn{1}{c|}{9.42} &
  \multicolumn{1}{c|}{8.87} &
  \multicolumn{1}{c|}{10.54} &
  \multicolumn{1}{c|}{5.79} &
  11.93 &
  \multicolumn{1}{c|}{12.56} &
  \multicolumn{1}{c|}{12.02} &
  \multicolumn{1}{c|}{12.45} &
  \multicolumn{1}{c|}{4.29} &
  0.90 \\ \hline
\multicolumn{1}{|c|}{30cm} &
  8cm &
  \multicolumn{1}{c|}{7.07} &
  \multicolumn{1}{c|}{6.39} &
  \multicolumn{1}{c|}{7.51} &
  \multicolumn{1}{c|}{9.64} &
  6.22 &
  \multicolumn{1}{c|}{10.61} &
  \multicolumn{1}{c|}{8.78} &
  \multicolumn{1}{c|}{11.91} &
  \multicolumn{1}{c|}{17.23} &
  12.23 &
  \multicolumn{1}{c|}{14.15} &
  \multicolumn{1}{c|}{12.02} &
  \multicolumn{1}{c|}{14.65} &
  \multicolumn{1}{c|}{15.06} &
  3.52 \\ \hline
\multicolumn{1}{|c|}{30cm} &
  12cm &
  \multicolumn{1}{c|}{7.85} &
  \multicolumn{1}{c|}{5.10} &
  \multicolumn{1}{c|}{7.78} &
  \multicolumn{1}{c|}{35.04} &
  0.97 &
  \multicolumn{1}{c|}{11.79} &
  \multicolumn{1}{c|}{9.64} &
  \multicolumn{1}{c|}{11.57} &
  \multicolumn{1}{c|}{18.25} &
  1.87 &
  \multicolumn{1}{c|}{15.73} &
  \multicolumn{1}{c|}{13.07} &
  \multicolumn{1}{c|}{15.90} &
  \multicolumn{1}{c|}{16.92} &
  1.04 \\ \hline
\multicolumn{1}{|c|}{50cm} &
  2cm &
  \multicolumn{1}{c|}{5.93} &
  \multicolumn{1}{c|}{-} &
  \multicolumn{1}{c|}{7.15} &
  \multicolumn{1}{c|}{-} &
  20.68 &
  \multicolumn{1}{c|}{8.89} &
  \multicolumn{1}{c|}{-} &
  \multicolumn{1}{c|}{7.21} &
  \multicolumn{1}{c|}{-} &
  18.94 &
  \multicolumn{1}{c|}{11.86} &
  \multicolumn{1}{c|}{-} &
  \multicolumn{1}{c|}{11.44} &
  \multicolumn{1}{c|}{-} &
  3.53 \\ \hline
\multicolumn{1}{|c|}{50cm} &
  4cm &
  \multicolumn{1}{c|}{6.17} &
  \multicolumn{1}{c|}{6.38} &
  \multicolumn{1}{c|}{7.02} &
  \multicolumn{1}{c|}{3.45} &
  13.87 &
  \multicolumn{1}{c|}{9.25} &
  \multicolumn{1}{c|}{8.68} &
  \multicolumn{1}{c|}{10.35} &
  \multicolumn{1}{c|}{6.16} &
  11.82 &
  \multicolumn{1}{c|}{12.34} &
  \multicolumn{1}{c|}{11.39} &
  \multicolumn{1}{c|}{12.56} &
  \multicolumn{1}{c|}{7.67} &
  1.82 \\ \hline
\multicolumn{1}{|c|}{50cm} &
  8cm &
  \multicolumn{1}{c|}{6.64} &
  \multicolumn{1}{c|}{5.82} &
  \multicolumn{1}{c|}{7.71} &
  \multicolumn{1}{c|}{12.42} &
  16.09 &
  \multicolumn{1}{c|}{9.97} &
  \multicolumn{1}{c|}{8.75} &
  \multicolumn{1}{c|}{11.11} &
  \multicolumn{1}{c|}{12.17} &
  11.50 &
  \multicolumn{1}{c|}{13.29} &
  \multicolumn{1}{c|}{11.54} &
  \multicolumn{1}{c|}{13.91} &
  \multicolumn{1}{c|}{13.15} &
  4.66 \\ \hline
\multicolumn{1}{|c|}{50cm} &
  12cm &
  \multicolumn{1}{c|}{7.11} &
  \multicolumn{1}{c|}{5.46} &
  \multicolumn{1}{c|}{6.9} &
  \multicolumn{1}{c|}{23.24} &
  2.97 &
  \multicolumn{1}{c|}{10.67} &
  \multicolumn{1}{c|}{8.97} &
  \multicolumn{1}{c|}{10.32} &
  \multicolumn{1}{c|}{15.99} &
  3.28 &
  \multicolumn{1}{c|}{14.24} &
  \multicolumn{1}{c|}{10.64} &
  \multicolumn{1}{c|}{14.49} &
  \multicolumn{1}{c|}{25.25} &
  1.76 \\ \hline
\multicolumn{1}{|c|}{None} &
  None &
  \multicolumn{1}{c|}{6} &
  \multicolumn{1}{c|}{6.5} &
  \multicolumn{1}{c|}{6.89} &
  \multicolumn{1}{c|}{8.41} &
  14.89 &
  \multicolumn{1}{c|}{9} &
  \multicolumn{1}{c|}{8.60} &
  \multicolumn{1}{c|}{9.94} &
  \multicolumn{1}{c|}{4.49} &
  10.40 &
  \multicolumn{1}{c|}{12} &
  \multicolumn{1}{c|}{11.67} &
  \multicolumn{1}{c|}{12.67} &
  \multicolumn{1}{c|}{2.78} &
  5.59 \\ \hline
\multicolumn{2}{|c|}{\textbf{Average AER}} &
  \multicolumn{3}{c|}{} &
  \multicolumn{1}{c|}{\textbf{15.27}} &
  \textbf{13.72} &
  \multicolumn{3}{c|}{\textbf{}} &
  \multicolumn{1}{c|}{\textbf{11.35}} &
  \textbf{9.36} &
  \multicolumn{3}{c|}{\textbf{}} &
  \multicolumn{1}{c|}{\textbf{15.12}} &
  \textbf{4.09} \\ \hline
\end{tabular}%
}
\end{table*}

To further investigate the AER in the physical world, we calculate the expected depths in meters based on Eqs.~(\ref{equ:concave}) and (\ref{equ:magnification}) and compare them with the experimental values in Table~\ref{tab:attack_error_rate_concave}. The last row shows the prediction accuracy of the Monodepth2 on the benign image. Note that since the depth estimation algorithm is not perfect and the physical experiments introduce some measuring errors, we consider the AER of the depth prediction lower than 15\% as accurate. 

When $d_b=2cm$, the attack lens can fully cover the image on the iPhone camera. However, the attack lens size is always smaller than the FLIR camera lens, so the attack lens we use cannot fully cover the image on AV. We measure the full image attack on iPhone in our experiments. Note that, with the larger size of the attack lens, we are still able to launch the full image attack on AV. The average AER is around 11\% regardless of the focal lengths, which means that the experimental depth value closely matches the expected depth value. 

\begin{figure*}[]
\centering
	\includegraphics[width=1\textwidth]{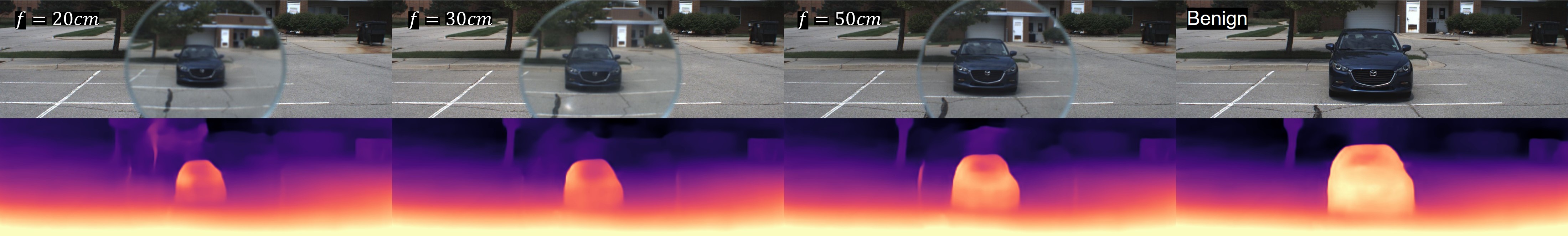}
	\caption{The concave lens attack on an AV with $d_{o1}=9m$, $d_b=12cm$ and different $f$ values.}
	\label{fig:cc_overview_9m_f_12cm}
\end{figure*}

When $d_b$ becomes larger than $4cm$, the attack lens is always in the camera view, which forms the scenario of the partial image attack. We can observe that the AER increases up to around 40\% on AV and 30\% on iPhone. Based on the average AER listed in the last row, we can conclude that our attack works similarly for both AV and iPhone. The overall average AER on both AV and iPhone is 11.48\%. It is also noticeable that the smaller $f$ introduces a higher AER, which is mainly caused by the blur in the in-lens or out-of-lens area due to the effect of DOF. For example, we show the collected images and their corresponding disparity map of concave lens attacks in Fig.~\ref{fig:cc_overview_9m_f_12cm} have the same $d_{o1}=9m$ and $d_b=12cm$ with different $f$ values. 

\subsubsection{Performance of Convex Lens Attacks}

Regarding the convex lens attack, we first investigate the first attack scenario. To realize the attack, we need to ensure $d_{o1} < f$. However, the focal length of commonly available convex lenses in the market is usually less than $1m$. Therefore, the first attack scenario is unsuitable for real-world driving because the expected object distance is too short in the attack. 

For the second attack scenario in the convex lens attacks, the formed image is inverted from the attack lens. Besides, it also requires the $d_b$ to be as large as $f$. Due to the inverted image and large $d_b$, it is also impractical to physically apply this attack. First, the inverted image may be detected by the AD error detection system easily. Second, because the $d_b$ is so large, e.g., $25cm$, it is difficult to conceal the attack lens without being noticed by a human driver. 

\begin{figure*}[]
\centering
	\includegraphics[width=1\textwidth]{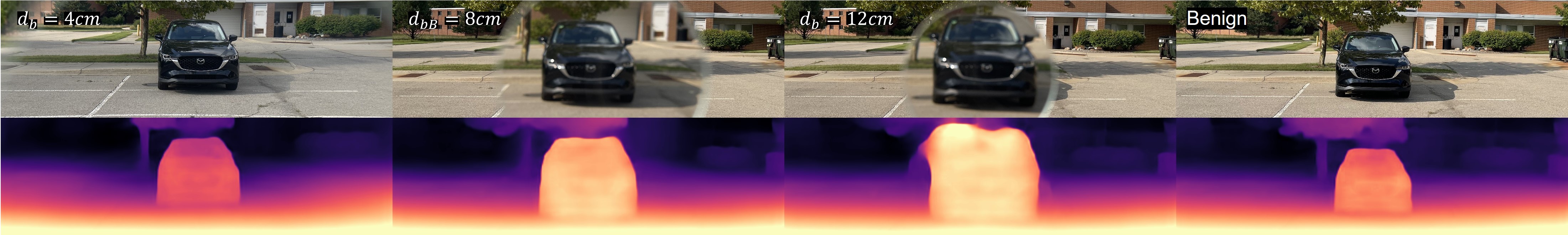}
	\caption{The convex lens attack on an iPhone with $d_{o1}=9m$, $f=50cm$, and various $d_b$.}
	\label{fig:cv_overview_9m_50cm_db}
\end{figure*}

In terms of the third attack scenario, it works with various $f$ and $d_b$. More detailed discussions regarding the feasible attack types in the physical world are discussed in Appendix~\ref{sec:Attack Types in Real AD Scenarios}. Some attack examples on iPhone with $d_{o1}=9m$, $f=50cm$ and various $d_b$ are shown in Fig.~\ref{fig:cv_overview_9m_50cm_db}. The attacked region is quite blurry with the larger $d_b$ because the formed image is focused in front of the image sensor, which is similar to a near-eyesight case. However, when $d_b$ is smaller, the blurriness is released. For example, when $d_b=4cm$, the blurriness is reduced significantly. In terms of the disparity map, as the blurriness is reduced, the disparity map becomes clearer and more accurate. Meanwhile, the result indicates that a larger $d_b$ can introduce a more significant impact on the depth. Note that this attack can also work for different attack distances as long as $d_{o1}>f$. We can also notice that a larger $f$ has more effect on the depth from Table~\ref{tab:attack_error_rate_convex}.

\begin{table*}[]
\centering
\caption{The AER of the convex lens attack with the various object distance $d_{o1}$, $f$, and $d_b$.}
\label{tab:attack_error_rate_convex}
\resizebox{\textwidth}{!}{%
\begin{tabular}{|cc|ccccc|ccccc|ccccc|}
\hline
\multicolumn{2}{|c|}{} &
  \multicolumn{5}{c|}{$d_{o1}$=6m} &
  \multicolumn{5}{c|}{$d_{o1}$=9m} &
  \multicolumn{5}{c|}{$d_{o1}$=12m} \\ \cline{3-17} 
\multicolumn{2}{|c|}{\multirow{-2}{*}{Attack Parameters}} &
  \multicolumn{1}{c|}{} &
  \multicolumn{2}{c|}{\begin{tabular}[c]{@{}c@{}}Experimental Depth\\ (meters)\end{tabular}} &
  \multicolumn{2}{c|}{\begin{tabular}[c]{@{}c@{}}AER\\ (\%)\end{tabular}} &
  \multicolumn{1}{c|}{} &
  \multicolumn{2}{c|}{\begin{tabular}[c]{@{}c@{}}Experimental Depth\\ (meters)\end{tabular}} &
  \multicolumn{2}{c|}{\begin{tabular}[c]{@{}c@{}}AER\\ (\%)\end{tabular}} &
  \multicolumn{1}{c|}{} &
  \multicolumn{2}{c|}{\begin{tabular}[c]{@{}c@{}}Experimental Depth\\ (meters)\end{tabular}} &
  \multicolumn{2}{c|}{\begin{tabular}[c]{@{}c@{}}AER\\ (\%)\end{tabular}} \\ \cline{1-2} \cline{4-7} \cline{9-12} \cline{14-17} 
\multicolumn{1}{|c|}{$f$} &
  $d_b$ &
  \multicolumn{1}{c|}{\multirow{-2}{*}{\begin{tabular}[c]{@{}c@{}}Expected\\ Depth (meters)\end{tabular}}} &
  \multicolumn{1}{c|}{AV} &
  \multicolumn{1}{c|}{iPhone} &
  \multicolumn{1}{c|}{AV} &
  iPhone &
  \multicolumn{1}{c|}{\multirow{-2}{*}{\begin{tabular}[c]{@{}c@{}}Expected\\ Depth (meters)\end{tabular}}} &
  \multicolumn{1}{c|}{AV} &
  \multicolumn{1}{c|}{iPhone} &
  \multicolumn{1}{c|}{AV} &
  iPhone &
  \multicolumn{1}{c|}{\multirow{-2}{*}{\begin{tabular}[c]{@{}c@{}}Expected\\ Depth (meters)\end{tabular}}} &
  \multicolumn{1}{c|}{AV} &
  \multicolumn{1}{c|}{iPhone} &
  \multicolumn{1}{c|}{AV} &
  iPhone \\ \hline
\multicolumn{1}{|c|}{20cm} &
  2cm &
  \multicolumn{1}{c|}{4.67} &
  \multicolumn{1}{c|}{-} &
  \multicolumn{1}{c|}{4.84} &
  \multicolumn{1}{c|}{-} &
  3.62 &
  \multicolumn{1}{c|}{6.98} &
  \multicolumn{1}{c|}{-} &
  \multicolumn{1}{c|}{8.90} &
  \multicolumn{1}{c|}{-} &
  27.49 &
  \multicolumn{1}{c|}{9.29} &
  \multicolumn{1}{c|}{-} &
  \multicolumn{1}{c|}{8.39} &
  \multicolumn{1}{c|}{-} &
  9.74 \\ \hline
\multicolumn{1}{|c|}{20cm} &
  4cm &
  \multicolumn{1}{c|}{4.08} &
  \multicolumn{1}{c|}{6.08} &
  \multicolumn{1}{c|}{5.80} &
  \multicolumn{1}{c|}{49.17} &
  42.26 &
  \multicolumn{1}{c|}{6.09} &
  \multicolumn{1}{c|}{7.92} &
  \multicolumn{1}{c|}{9.66} &
  \multicolumn{1}{c|}{30.07} &
  58.66 &
  \multicolumn{1}{c|}{8.10} &
  \multicolumn{1}{c|}{12.07} &
  \multicolumn{1}{c|}{12.16} &
  \multicolumn{1}{c|}{49.10} &
  50.16 \\ \hline
\multicolumn{1}{|c|}{20cm} &
  8cm &
  \multicolumn{1}{c|}{2.90} &
  \multicolumn{1}{c|}{10.73} &
  \multicolumn{1}{c|}{6.88} &
  \multicolumn{1}{c|}{{\color[HTML]{FE0000} 269.86}} &
  {{\color[HTML]{FE0000} 137.27}} &
  \multicolumn{1}{c|}{4.31} &
  \multicolumn{1}{c|}{5.27} &
  \multicolumn{1}{c|}{7.54} &
  \multicolumn{1}{c|}{22.31} &
  74.93 &
  \multicolumn{1}{c|}{5.72} &
  \multicolumn{1}{c|}{7.00} &
  \multicolumn{1}{c|}{12.31} &
  \multicolumn{1}{c|}{22.45} &
  {\color[HTML]{FE0000} 115.18} \\ \hline
\multicolumn{1}{|c|}{20cm} &
  12cm &
  \multicolumn{1}{c|}{1.74} &
  \multicolumn{1}{c|}{15.59} &
  \multicolumn{1}{c|}{4.57} &
  \multicolumn{1}{c|}{{\color[HTML]{FE0000} 796.73}} &
  {\color[HTML]{FE0000} 162.69} &
  \multicolumn{1}{c|}{2.55} &
  \multicolumn{1}{c|}{5.30} &
  \multicolumn{1}{c|}{14.12} &
  \multicolumn{1}{c|}{{\color[HTML]{FE0000} 108.04}} &
  {\color[HTML]{FE0000} 453.99} &
  \multicolumn{1}{c|}{3.36} &
  \multicolumn{1}{c|}{7.62} &
  \multicolumn{1}{c|}{5.93} &
  \multicolumn{1}{c|}{{\color[HTML]{FE0000} 126.80}} &
  {76.40} \\ \hline
\multicolumn{1}{|c|}{30cm} &
  2cm &
  \multicolumn{1}{c|}{5.13} &
  \multicolumn{1}{c|}{-} &
  \multicolumn{1}{c|}{7.07} &
  \multicolumn{1}{c|}{-} &
  37.80 &
  \multicolumn{1}{c|}{7.67} &
  \multicolumn{1}{c|}{-} &
  \multicolumn{1}{c|}{8.96} &
  \multicolumn{1}{c|}{-} &
  16.81 &
  \multicolumn{1}{c|}{10.21} &
  \multicolumn{1}{c|}{-} &
  \multicolumn{1}{c|}{10.90} &
  \multicolumn{1}{c|}{-} &
  6.75 \\ \hline
\multicolumn{1}{|c|}{30cm} &
  4cm &
  \multicolumn{1}{c|}{4.73} &
  \multicolumn{1}{c|}{4.91} &
  \multicolumn{1}{c|}{7.35} &
  \multicolumn{1}{c|}{3.79} &
  55.17 &
  \multicolumn{1}{c|}{7.07} &
  \multicolumn{1}{c|}{8.19} &
  \multicolumn{1}{c|}{10.17} &
  \multicolumn{1}{c|}{15.79} &
  43.75 &
  \multicolumn{1}{c|}{9.42} &
  \multicolumn{1}{c|}{10.36} &
  \multicolumn{1}{c|}{14.52} &
  \multicolumn{1}{c|}{10.01} &
  54.19 \\ \hline
\multicolumn{1}{|c|}{30cm} &
  8cm &
  \multicolumn{1}{c|}{3.95} &
  \multicolumn{1}{c|}{4.97} &
  \multicolumn{1}{c|}{6.22} &
  \multicolumn{1}{c|}{{\color[HTML]{333333} 25.92}} &
  57.40 &
  \multicolumn{1}{c|}{5.89} &
  \multicolumn{1}{c|}{7.95} &
  \multicolumn{1}{c|}{7.50} &
  \multicolumn{1}{c|}{34.99} &
  27.36 &
  \multicolumn{1}{c|}{7.83} &
  \multicolumn{1}{c|}{6.25} &
  \multicolumn{1}{c|}{11.42} &
  \multicolumn{1}{c|}{20.17} &
  45.89 \\ \hline
\multicolumn{1}{|c|}{30cm} &
  12cm &
  \multicolumn{1}{c|}{3.18} &
  \multicolumn{1}{c|}{15.42} &
  \multicolumn{1}{c|}{5.58} &
  \multicolumn{1}{c|}{{\color[HTML]{FE0000} 385.54}} &
  75.54 &
  \multicolumn{1}{c|}{4.72} &
  \multicolumn{1}{c|}{6.19} &
  \multicolumn{1}{c|}{7.99} &
  \multicolumn{1}{c|}{31.18} &
  69.35 &
  \multicolumn{1}{c|}{6.26} &
  \multicolumn{1}{c|}{6.17} &
  \multicolumn{1}{c|}{9.39} &
  \multicolumn{1}{c|}{1.45} &
  50.07 \\ \hline
\multicolumn{1}{|c|}{50cm} &
  2cm &
  \multicolumn{1}{c|}{5.50} &
  \multicolumn{1}{c|}{-} &
  \multicolumn{1}{c|}{6.86} &
  \multicolumn{1}{c|}{-} &
  24.70 &
  \multicolumn{1}{c|}{8.22} &
  \multicolumn{1}{c|}{-} &
  \multicolumn{1}{c|}{9.03} &
  \multicolumn{1}{c|}{-} &
  9.82 &
  \multicolumn{1}{c|}{10.95} &
  \multicolumn{1}{c|}{-} &
  \multicolumn{1}{c|}{11.51} &
  \multicolumn{1}{c|}{-} &
  5.11 \\ \hline
\multicolumn{1}{|c|}{50cm} &
  4cm &
  \multicolumn{1}{c|}{5.26} &
  \multicolumn{1}{c|}{5.73} &
  \multicolumn{1}{c|}{7.18} &
  \multicolumn{1}{c|}{{\color[HTML]{333333} 8.90}} &
  36.47 &
  \multicolumn{1}{c|}{7.87} &
  \multicolumn{1}{c|}{8.68} &
  \multicolumn{1}{c|}{11.63} &
  \multicolumn{1}{c|}{10.38} &
  47.88 &
  \multicolumn{1}{c|}{10.47} &
  \multicolumn{1}{c|}{11.75} &
  \multicolumn{1}{c|}{14.92} &
  \multicolumn{1}{c|}{12.26} &
  42.54 \\ \hline
\multicolumn{1}{|c|}{50cm} &
  8cm &
  \multicolumn{1}{c|}{4.79} &
  \multicolumn{1}{c|}{4.99} &
  \multicolumn{1}{c|}{5.80} &
  \multicolumn{1}{c|}{{\color[HTML]{333333} 4.19}} &
  21.11 &
  \multicolumn{1}{c|}{7.16} &
  \multicolumn{1}{c|}{8.18} &
  \multicolumn{1}{c|}{8.16} &
  \multicolumn{1}{c|}{14.35} &
  14.08 &
  \multicolumn{1}{c|}{9.52} &
  \multicolumn{1}{c|}{10.72} &
  \multicolumn{1}{c|}{10.75} &
  \multicolumn{1}{c|}{12.60} &
  12.94 \\ \hline
\multicolumn{1}{|c|}{50cm} &
  12cm &
  \multicolumn{1}{c|}{4.33} &
  \multicolumn{1}{c|}{7.57} &
  \multicolumn{1}{c|}{6.79} &
  \multicolumn{1}{c|}{75.04} &
  56.99 &
  \multicolumn{1}{c|}{6.45} &
  \multicolumn{1}{c|}{7.90} &
  \multicolumn{1}{c|}{7.43} &
  \multicolumn{1}{c|}{22.43} &
  15.13 &
  \multicolumn{1}{c|}{8.57} &
  \multicolumn{1}{c|}{8.03} &
  \multicolumn{1}{c|}{9.87} &
  \multicolumn{1}{c|}{6.34} &
  15.06 \\ \hline
\multicolumn{1}{|c|}{None} &
  None &
  \multicolumn{1}{c|}{6} &
  \multicolumn{1}{c|}{6.50} &
  \multicolumn{1}{c|}{6.89} &
  \multicolumn{1}{c|}{8.41} &
  14.78 &
  \multicolumn{1}{c|}{9} &
  \multicolumn{1}{c|}{8.60} &
  \multicolumn{1}{c|}{9.94} &
  \multicolumn{1}{c|}{4.49} &
  10.40 &
  \multicolumn{1}{c|}{12} &
  \multicolumn{1}{c|}{11.67} &
  \multicolumn{1}{c|}{12.67} &
  \multicolumn{1}{c|}{2.78} &
  5.59 \\ \hline
\multicolumn{2}{|c|}{\textbf{Average AER}} &
  \multicolumn{3}{c|}{\textbf{}} &
  \multicolumn{1}{c|}{\textbf{18.39}} &
  \textbf{47.28} &
  \multicolumn{3}{c|}{\textbf{}} &
  \multicolumn{1}{c|}{\textbf{22.69}} &
  \textbf{36.84} &
  \multicolumn{3}{c|}{\textbf{}} &
  \multicolumn{1}{c|}{\textbf{16.80}} &
  \textbf{37.06} \\ \hline
\end{tabular}%
}
\end{table*}

Furthermore, to investigate the AER in the physical world, we calculate the expected depth in meters using Eqs.~(\ref{equ:magnification}) and (\ref{equ:convex}). Table~\ref{tab:attack_error_rate_convex} shows the AER of the convex lens attack with varied $f$ and $d_b$ values for different object distances. Similar to the results of the concave lens attacks in Table~\ref{tab:attack_error_rate_concave}, when $d_b$ is $2cm$, the AER is around 14\%. 

\begin{figure}[b]
\centering
	\includegraphics[width=0.65\columnwidth]{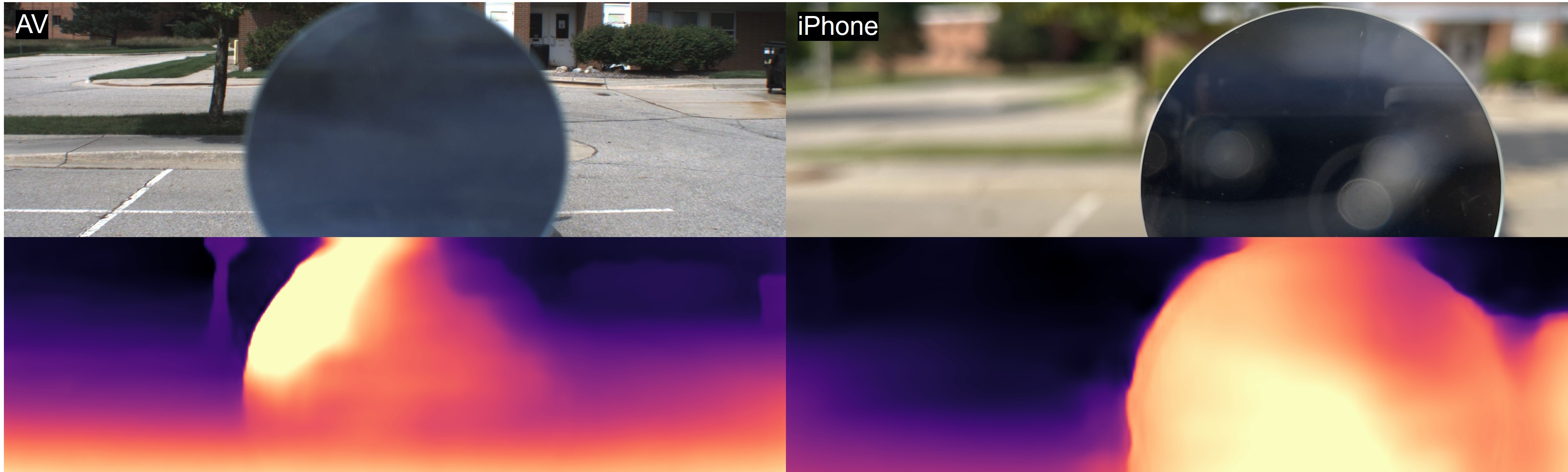}
	\caption{The convex lens attack on AV and iPhone cameras with $d_{o1}=6m$, $f=20cm$, and $d_b=12cm$.}
	\label{fig:non_success_convex}
\end{figure} 

However, for partial image attacks, with the smaller $f$ and larger $d_b$, the AER increases. Sometimes, it could increase to more than 100\% (highlighted in red in Table~\ref{tab:attack_error_rate_convex}). 
We show an example in Fig.~\ref{fig:non_success_convex}.
When we look at the convex lens attack on AV and iPhone cameras with $d_{o1}=6m$, $f=20cm$, and $d_b=12cm$, we can see that it is much blurrier in the in-lens or out-of-lens area compared to the case with the smaller $d_{b}$ and larger $f$.
This occurs due to the introduction of a stronger depth of field, which is achieved by employing larger $d_{b}$ values and smaller $f$ values. Due to the strong blurriness, convex lens attacks also obscure the target object, leading to detection failures in YOLO v8. Therefore, we mark the object detection failure case using red color in Table~\ref{tab:attack_error_rate_convex}. We notice that car detection failure often matches with a high AER, which is not preferred when the attacker launches the attack. Therefore, we calculate the average AER in the last row by omitting these red numbers. The overall AER on both AV and iPhone is much higher than the AER in concave lens attacks. Meanwhile, the average AER on the iPhone is usually higher than that on AV. The average AER across both the AV and iPhone is 29.84\%.

\cbox{To summarize, \attack works for both AV and iPhone in the physical world in partial and full image attacks. The average AER is lower on iPhone in the concave lens attack and lower on AV in the convex lens attack.
The overall average AER for both the AV and iPhone is approximately 11.48\% in the concave lens attack and 29.84\% in the convex lens attack, which demonstrates the effectiveness of \attack in real AD scenarios. 
}

\subsection{Evaluation of Attack Optimization}\label{sec:evl_opti}


We evaluate \attack by adjusting $\alpha$, which controls the relative weights of $L_{veh}$ and $L_{out}$ in the total loss. Since $\alpha$ is predefined by the attacker in real-world scenarios, they can estimate the AER or ADR beforehand based on the optimization results. We explore values of $\alpha$ from 0.1 to 0.4 with a step size of 0.1, as values between 0.5 and 0.9 would produce the opposite result, as indicated by Equation~\ref{equ:loss}.


\subsubsection{Optimization of Concave Lens Attack}

\begin{table}[b]
\centering
\caption{Optimization results of concave lens attack.}
\label{tab:concave_optimization_results}
\resizebox{0.7\columnwidth}{!}{%
\begin{tabular}{|c|cccc|cccc|cccc|}
\hline
\multirow{3}{*}{$\alpha$} &
  \multicolumn{4}{c|}{Monodepth2} &
  \multicolumn{4}{c|}{Depth Hints} &
  \multicolumn{4}{c|}{Lite-Mono} \\ \cline{2-13} 
 &
  \multicolumn{2}{c|}{Targeted} &
  \multicolumn{2}{c|}{Untargeted} &
  \multicolumn{2}{c|}{Targeted} &
  \multicolumn{2}{c|}{Untargeted} &
  \multicolumn{2}{c|}{Targeted} &
  \multicolumn{2}{c|}{Untargeted} \\ \cline{2-13} 
 &
  \multicolumn{1}{c|}{$f$} &
  \multicolumn{1}{c|}{AER} &
  \multicolumn{1}{c|}{$f$} &
  ADR &
  \multicolumn{1}{c|}{$f$} &
  \multicolumn{1}{c|}{AER} &
  \multicolumn{1}{c|}{$f$} &
  ADR &
  \multicolumn{1}{c|}{$f$} &
  \multicolumn{1}{c|}{AER} &
  \multicolumn{1}{c|}{$f$} &
  ADR \\ \hline
0.1 &
  \multicolumn{1}{c|}{4} &
  \multicolumn{1}{c|}{0.77\%} &
  \multicolumn{1}{c|}{2} &
  3.32\% &
  \multicolumn{1}{c|}{7} &
  \multicolumn{1}{c|}{1.12\%} &
  \multicolumn{1}{c|}{9} &
  8.71\% &
  \multicolumn{1}{c|}{7} &
  \multicolumn{1}{c|}{0.67\%} &
  \multicolumn{1}{c|}{8} &
  15.57\% \\ \hline
0.2 &
  \multicolumn{1}{c|}{2} &
  \multicolumn{1}{c|}{1.12\%} &
  \multicolumn{1}{c|}{2} &
  3.32\% &
  \multicolumn{1}{c|}{7} &
  \multicolumn{1}{c|}{1.12\%} &
  \multicolumn{1}{c|}{7} &
  7.67\% &
  \multicolumn{1}{c|}{7} &
  \multicolumn{1}{c|}{0.67\%} &
  \multicolumn{1}{c|}{8} &
  15.57\% \\ \hline
0.3 &
  \multicolumn{1}{c|}{2} &
  \multicolumn{1}{c|}{1.12\%} &
  \multicolumn{1}{c|}{2} &
  3.32\% &
  \multicolumn{1}{c|}{6} &
  \multicolumn{1}{c|}{1.77\%} &
  \multicolumn{1}{c|}{1} &
  1.50\% &
  \multicolumn{1}{c|}{7} &
  \multicolumn{1}{c|}{0.67\%} &
  \multicolumn{1}{c|}{8} &
  15.57\% \\ \hline
0.4 &
  \multicolumn{1}{c|}{2} &
  \multicolumn{1}{c|}{1.12\%} &
  \multicolumn{1}{c|}{2} &
  3.32\% &
  \multicolumn{1}{c|}{4} &
  \multicolumn{1}{c|}{4.05\%} &
  \multicolumn{1}{c|}{1} &
  1.50\% &
  \multicolumn{1}{c|}{6} &
  \multicolumn{1}{c|}{1.21\%} &
  \multicolumn{1}{c|}{8} &
  15.57\% \\ \hline
\begin{tabular}[c]{@{}c@{}}Average\\ with Optimization\end{tabular} &
  \multicolumn{2}{c|}{1.03\%} &
  \multicolumn{2}{c|}{3.32\%} &
  \multicolumn{2}{c|}{2.01\%} &
  \multicolumn{2}{c|}{4.85\%} &
  \multicolumn{2}{c|}{0.81\%} &
  \multicolumn{2}{c|}{15.57\%} \\ \hline
\begin{tabular}[c]{@{}c@{}}Without\\ Optimization\end{tabular} &
  \multicolumn{2}{c|}{2.89\%} &
  \multicolumn{2}{c|}{2.22\%} &
  \multicolumn{2}{c|}{4.28\%} &
  \multicolumn{2}{c|}{3.65\%} &
  \multicolumn{2}{c|}{5.51\%} &
  \multicolumn{2}{c|}{9.09\%} \\ \hline
\end{tabular}%
}

\scriptsize{Note: Smaller AER is better and larger ADR is better.}
\end{table}

Table~\ref{tab:concave_optimization_results} shows the optimization results of the concave lens attacks, including targeted and untargeted attacks. For targeted attacks, with $\alpha$ increasing, $f$ decreases and the AER increases in all three model attacks. The reason is that with $\alpha$ increasing, the in-lens area becomes less dominant in $L_{total}$ and out-of-lens area loss $L_{out}$ gain more weight. In other words, $L_{total}$ cares more about the attack accuracy of the out-of-lens area. Smaller $f$ causes less blurriness at the out-of-lens area. As a result, the AER will increase as smaller $f$ is preferred, which leads to less accuracy of in-lens depth compared with the targeted depth. 

\begin{figure*}[t]
\centering
	\includegraphics[width=1\textwidth]{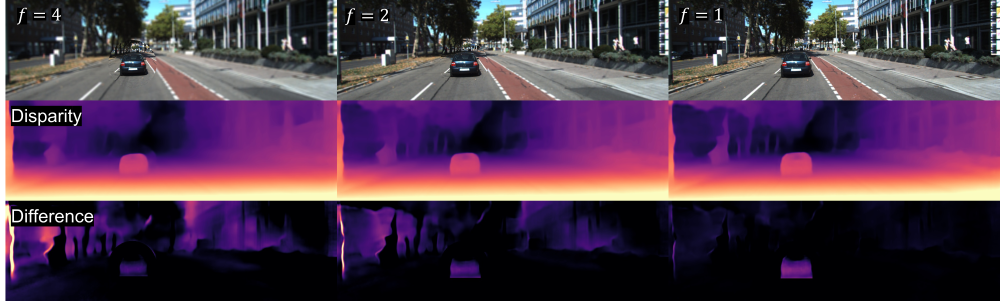}
	\caption{The targeted concave lens attacks with different $f$. The first row shows the blurred RGB image after attacks. The second row shows the disparity map. The third row shows the disparity difference compared to the targeted depth value.}
	\label{fig:targeted_op_concave}
\end{figure*}

In accordance with Table~\ref{tab:concave_optimization_results}, we visualize the targeted concave lens attack results in Fig.~\ref{fig:targeted_op_concave}. The first row shows the blurred RGB image after attacks. The second row shows the disparity map. The third row shows the disparity difference compared to the targeted depth value. When $f=4$, the disparity difference on the vehicle is less than the disparity difference on the vehicle when $f=1$, whereas the out-of-lens difference becomes larger.

Table~\ref{tab:concave_optimization_results} also presents the untargeted concave lens attack. With $\alpha$ increasing, $f$ decreases and the ADR decreases in all three model attacks. When $\alpha$ is small, the optimization loss tries to find a larger $f$ which can distort the benign depth of the in-lens area the most. Therefore, the ADR is large. However, with $\alpha$ increasing, more weight in the $L_{total}$ goes to $L_{out}$. The smaller $f$ has less depth difference in the out-of-lens area compared with the benign case. Thus, when $\alpha$ becomes large enough, the optimization process usually finds a smaller $f$ to optimize the attack.

We calculate the overall average values of AER and ADR for all $\alpha$ in the last second row of Table~\ref{tab:concave_optimization_results}. The last row indicates the AER or ADR of randomly choosing an attack lens in targeted or untargeted attacks. By comparing them, it can be seen that the AER/ADR of the targeted/untargeted attack using optimization is always better than the attack with the randomly chosen lens. 
Thus, attack optimization can indeed contribute to higher depth attack accuracy, and further improve the attack success rate.

\begin{table}[b]
\centering
\caption{Optimization results of convex lens attack.}
\label{tab:convex_optimization_results}
\resizebox{0.7\columnwidth}{!}{%
\begin{tabular}{|c|cccc|cccc|cccc|}
\hline
\multirow{3}{*}{$\alpha$} &
  \multicolumn{4}{c|}{Monodepth2} &
  \multicolumn{4}{c|}{Depth Hints} &
  \multicolumn{4}{c|}{Lite-Mono} \\ \cline{2-13} 
 &
  \multicolumn{2}{c|}{Targeted} &
  \multicolumn{2}{c|}{Untargeted} &
  \multicolumn{2}{c|}{Targeted} &
  \multicolumn{2}{c|}{Untargeted} &
  \multicolumn{2}{c|}{Targeted} &
  \multicolumn{2}{c|}{Untargeted} \\ \cline{2-13} 
 &
  \multicolumn{1}{c|}{$f$} &
  \multicolumn{1}{c|}{AER} &
  \multicolumn{1}{c|}{$f$} &
  ADR &
  \multicolumn{1}{c|}{$f$} &
  \multicolumn{1}{c|}{AER} &
  \multicolumn{1}{c|}{$f$} &
  ADR &
  \multicolumn{1}{c|}{$f$} &
  \multicolumn{1}{c|}{AER} &
  \multicolumn{1}{c|}{$f$} &
  ADR \\ \hline
0.1 &
  \multicolumn{1}{c|}{8} &
  \multicolumn{1}{c|}{2.7\%} &
  \multicolumn{1}{c|}{9} &
  38.32\% &
  \multicolumn{1}{c|}{6} &
  \multicolumn{1}{c|}{0.62\%} &
  \multicolumn{1}{c|}{9} &
  44.2\% &
  \multicolumn{1}{c|}{7} &
  \multicolumn{1}{c|}{1.03\%} &
  \multicolumn{1}{c|}{9} &
  29.77\% \\ \hline
0.2 &
  \multicolumn{1}{c|}{8} &
  \multicolumn{1}{c|}{2.7\%} &
  \multicolumn{1}{c|}{9} &
  38.32\% &
  \multicolumn{1}{c|}{6} &
  \multicolumn{1}{c|}{0.62\%} &
  \multicolumn{1}{c|}{9} &
  44.2\% &
  \multicolumn{1}{c|}{7} &
  \multicolumn{1}{c|}{1.03\%} &
  \multicolumn{1}{c|}{9} &
  29.77\% \\ \hline
0.3 &
  \multicolumn{1}{c|}{8} &
  \multicolumn{1}{c|}{2.7\%} &
  \multicolumn{1}{c|}{9} &
  38.32\% &
  \multicolumn{1}{c|}{6} &
  \multicolumn{1}{c|}{0.62\%} &
  \multicolumn{1}{c|}{9} &
  44.2\% &
  \multicolumn{1}{c|}{7} &
  \multicolumn{1}{c|}{1.03\%} &
  \multicolumn{1}{c|}{9} &
  29.77\% \\ \hline
0.4 &
  \multicolumn{1}{c|}{8} &
  \multicolumn{1}{c|}{2.7\%} &
  \multicolumn{1}{c|}{9} &
  38.32\% &
  \multicolumn{1}{c|}{6} &
  \multicolumn{1}{c|}{0.62\%} &
  \multicolumn{1}{c|}{9} &
  44.2\% &
  \multicolumn{1}{c|}{7} &
  \multicolumn{1}{c|}{1.03\%} &
  \multicolumn{1}{c|}{9} &
  29.77\% \\ \hline
\begin{tabular}[c]{@{}c@{}}Average\\ with Optimization\end{tabular} &
  \multicolumn{2}{c|}{2.7\%} &
  \multicolumn{2}{c|}{38.32\%} &
  \multicolumn{2}{c|}{0.62\%} &
  \multicolumn{2}{c|}{44.2\%} &
  \multicolumn{2}{c|}{1.03\%} &
  \multicolumn{2}{c|}{29.77\%} \\ \hline
\begin{tabular}[c]{@{}c@{}}Without\\ Optimization\end{tabular} &
  \multicolumn{2}{c|}{13.96\%} &
  \multicolumn{2}{c|}{16.55\%} &
  \multicolumn{2}{c|}{10.12\%} &
  \multicolumn{2}{c|}{22.73\%} &
  \multicolumn{2}{c|}{9.04\%} &
  \multicolumn{2}{c|}{12.31\%} \\ \hline
\end{tabular}%
}

\scriptsize{Note: Smaller AER is better and larger ADR is better.}
\end{table}

\subsubsection{Optimization of Convex Lens Attack}
Table~\ref{tab:convex_optimization_results} showcases the optimization results of the targeted and untargeted convex lens attacks. It is very different from what we have for concave lens attacks. Regardless of targeted or untargeted attacks, with varying values of $\alpha$, the optimized $f$ is always constant, which leads to an unchanged AER and ADR. The underlying cause is that the depth change of the out-of-lens area with varying $f$ is very minor. The blurriness contributed by the convex lens all goes to the in-lens area, whereas the out-of-lens area remains the same or very similar to the benign situation. Again, less RGB value change of the out-of-lens area leads to less depth change of the corresponding area. On the other hand, when blurriness is added to the in-lens area, not only does the object size of the in-lens area change but it also becomes more blurry, which explains why the AER in Table~\ref{tab:attack_error_rate_convex} is much larger than expected. 

Compared to the non-optimized scenarios (the last row), the AER of the optimized attack is always smaller than the non-optimized case in the targeted attack. Similarly, the ADR of the optimized attack is always larger. 


\begin{figure}[t]
\centering
	\includegraphics[width=0.7\columnwidth]{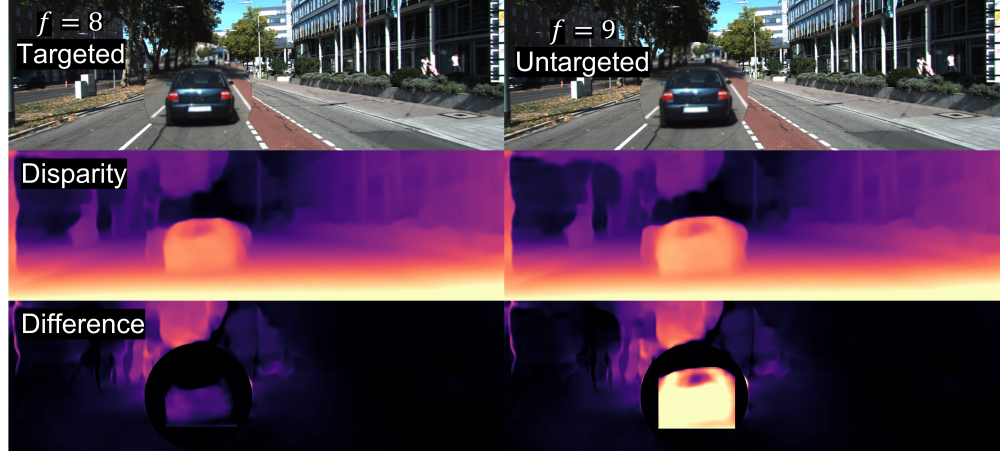}
	\caption{The optimized targeted and untargeted convex lens attacks. The first row shows the blurred RGB image after attacks. The second row shows the disparity map. The third row shows the disparity difference compared to the targeted depth value (targeted) or benign depth value (untargeted).}
	\label{fig:targeted_untargeted_op_convex}
\end{figure}

    In accordance with Table~\ref{tab:convex_optimization_results}, we also visualize the targeted and untargeted convex lens attacks with the optimized focal length for Monodepth2 in Fig.~\ref{fig:targeted_untargeted_op_convex}. The first column of Table~\ref{tab:convex_optimization_results} shows the targeted attack with the optimized focal length $f=8$. The disparity differences in both the target vehicle area (2.7\%) and the out-of-lens area are tiny. Moreover, the second column demonstrates the untargeted attack with the optimized focal length $f=9$. The vehicle area exhibits significant depth distortion (38.32\%), while the out-of-lens area experiences minimal depth change. 

We produce these results by assuming that $d_b$ and $d_o$ are constant. However, constant $f$ with varying $\alpha$ does not imply that our attack optimization is ineffective. Rather, this indicates that the attack is easier to control in a physical context and is more robust across various weight coefficients. 
When different $d_b$ and $d_o$ are taken into account, the optimized $f$ will vary.


\cbox{To summarize, with attack optimization, the overall targeted AER can be reduced by 6.26\%, and the untargeted attack ADR can be increased by around 11.58\%. Especially, with a small $\alpha$, the AER is small enough to launch a successful targeted attack, and the ADR is also large enough to make an effective untargeted attack. The results show that our attack optimization can indeed improve the attack performance compared to the attack without optimization.
}


\subsection{End-to-End System Simulation}\label{sec:end-to-end}

So far, we have investigated the impact of our attack at the AI component level. A broader AV system level~\cite{shen2022sok} evaluation would provide a more comprehensive understanding of the realistic impacts of our attack. Therefore, we conduct closed-loop simulations involving other integral components of AD, such as prediction, planning, and control using CARLA simulator~\cite{Dosovitskiy17}. CARLA is an open-source simulator designed to facilitate the advancement, training, and validation of AD systems. The simulator selection is based on its representativeness and versatility.

\begin{figure}[t]
\centering
	\includegraphics[width=0.65\columnwidth]{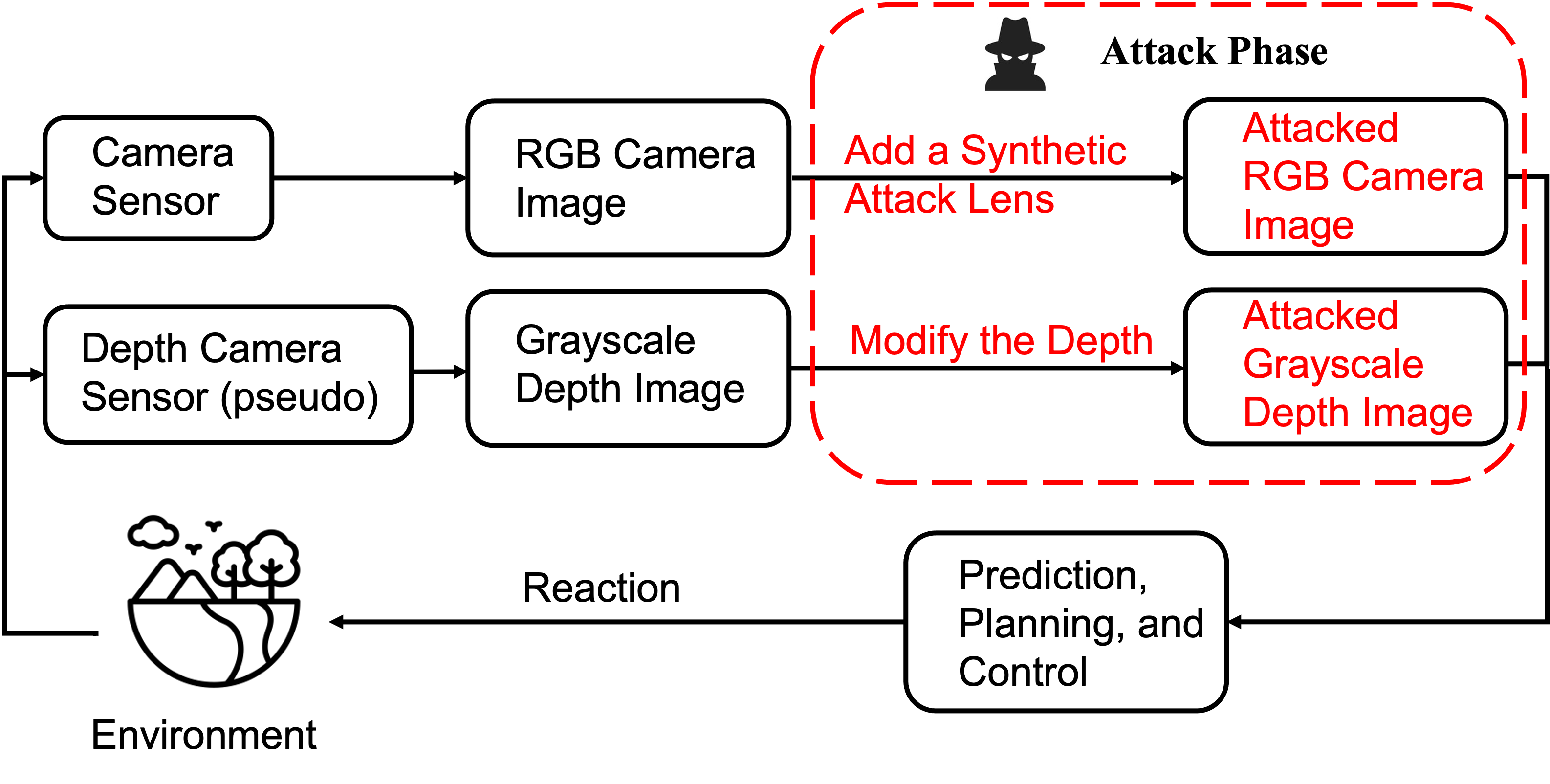}
	\caption{\attack simulation workflow using CARLA.}
	\label{fig:Carla_Workflow}
\end{figure}

Fig.~\ref{fig:Carla_Workflow} shows the \attack simulation workflow using CARLA. We choose ``Tesla Model 3'' as the AV model as it only uses camera sensors to sense the environment~\cite{Autopilot}. 
The depth map in CARLA is the ground truth value from the pseudo-sensor~\cite{carlasensor}. We enable both the RGB camera sensor and depth camera sensor to simulate our attack. CARLA provides the PythonAPI, which allows the attacker to obtain and modify the sensor data. Therefore, we add a synthetic attack lens to the RGB camera image and modify the depth value corresponding to the location of the attack lens in the depth map. Next, the manipulated RGB image and depth map will be sent to the prediction, planning, and control modules. Finally, the perception module will output a driving decision that reacts to the sensed environment. 

\begin{figure*}[t]
    \centering
    \subfigure[Starting braking (benign)]{\includegraphics[width=0.45\textwidth]{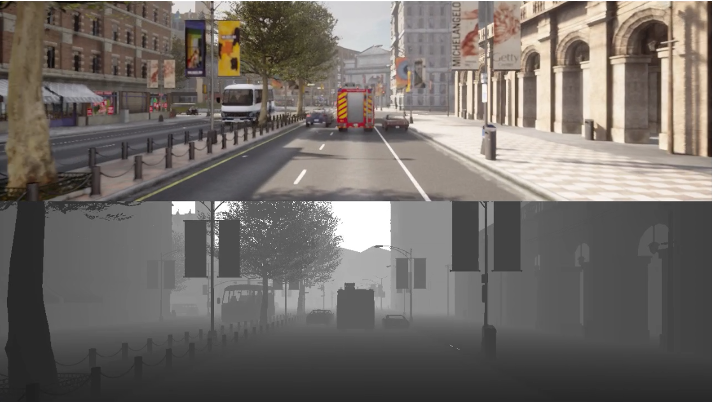}} 
    \subfigure[Fully stop (benign)]{\includegraphics[width=0.45\textwidth]{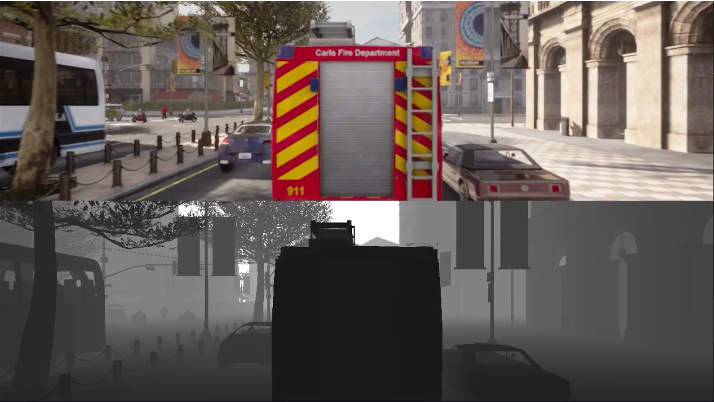}} 
    \subfigure[Starting braking (under attack)]{\includegraphics[width=0.45\textwidth]{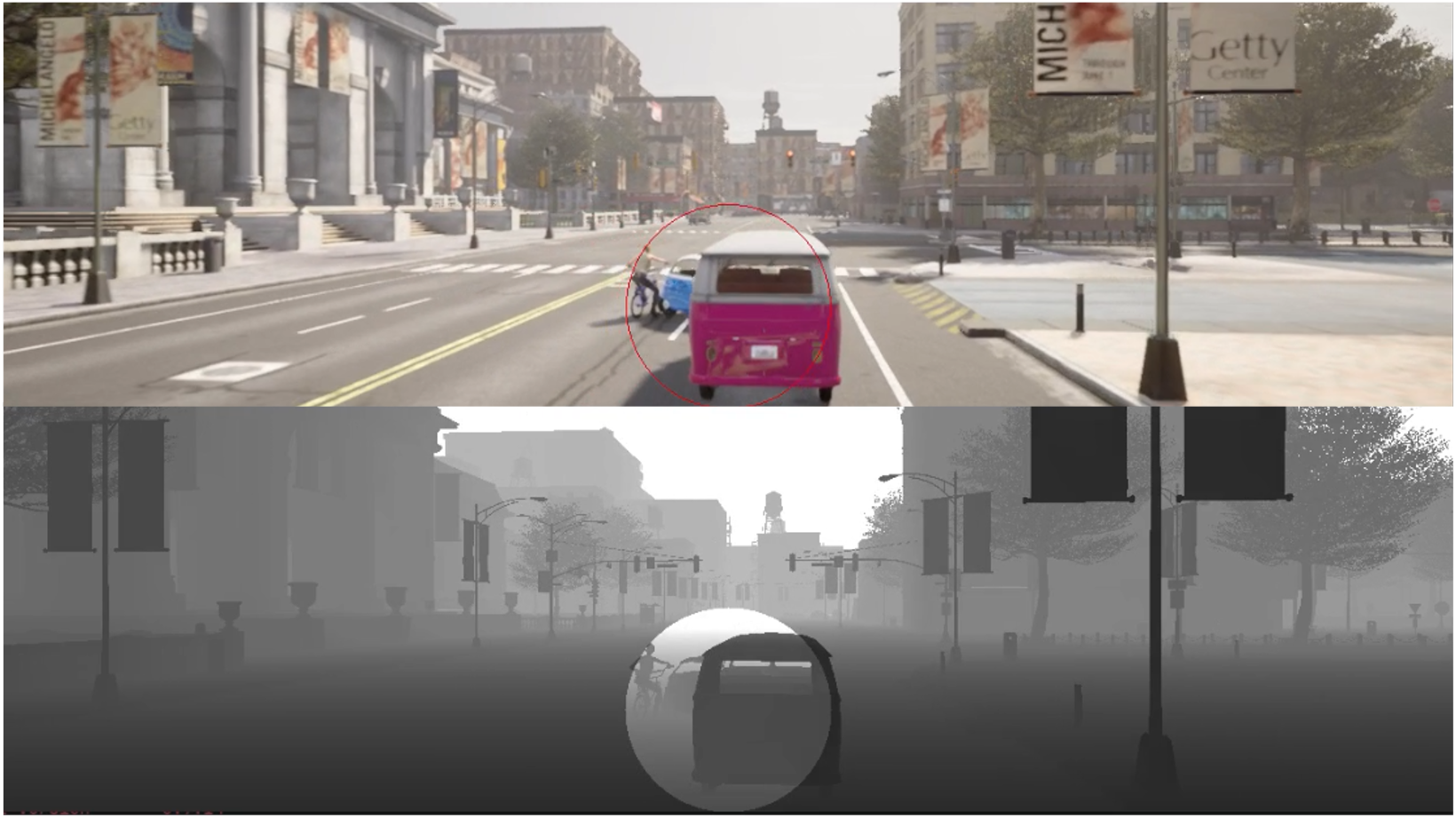}}
    \subfigure[Car crash (under attack)]{\includegraphics[width=0.45\textwidth]{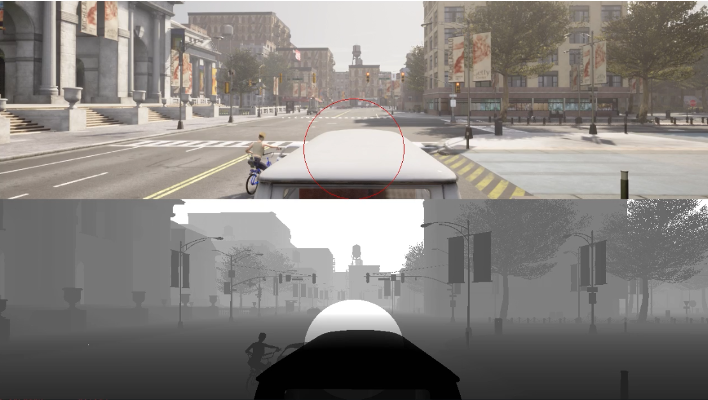}}
    \caption{CARLA simulation of benign scenario and the scenario under \attack. RGB camera sensor view and its corresponding depth are shown for comparison. (a) shows the benign driving case that the AV senses the depth of the fire truck is near and starts to brake to avoid collision, and (b) shows the AV fully stops behind the fire truck; (c) shows the AV starts to brake when depth estimation is compromised, and (d) shows the AV crashes with the front car due to the short braking distance. Note: the red circle is the synthetic attack lens.}
    \label{fig:carla_results}
\end{figure*}

We compare simulation results in benign scenarios and the scenario under \attack. We showcase simulation examples in Fig.~\ref{fig:carla_results}. When the AV is driving in the benign case (Figs.~\ref{fig:carla_results}(a)(b)), it senses the fire truck with the correct depth. It starts braking at a proper distance from the fire truck, which allows it to stop right behind it. However, when the attack lens is applied (Figs.~\ref{fig:carla_results}(c)(d)), the depth of the front car is estimated wrongly. The depth of the leading car is considered far away, whereas, it is close by in reality. As a result, the available braking distance between the AV and the front car is much shorter than it is in the benign case, resulting in a car crash.

\cbox{To summarize, we simulate \attack in the end-to-end system with the CARLA simulator. Our simulation results highlight the potential threats of our attack towards a real-world autonomous driving pipeline.}

\section{Discussion and Limitation}\label{sec:discussion}

We discuss some common concerns of our attack regarding attack types in real AD scenarios, attack on AV with multi-camera, the generality, and the limitation. We also present potential defense methods against \attack. 

\textbf{Attack Types in Real AD Scenarios.}
In the physical attack, regarding the practicality and stealthiness of the attack,  $d_b$ is usually not large, e.g., $5cm$. Besides, the object is often far from the camera, e.g., $10m$. Because of the constraints, only the concave lens attack and the third attack scenario of the convex lens attack are feasible. Therefore, as discussed in Section~\ref{sec:Real-World Physical Attack}, we  demonstrate these two attacks in physical AD scenarios. More detailed mathematical analysis can be found in Appendix~\ref{sec:Attack Types in Real AD Scenarios}.

\textbf{Attack on AV with Multi-camera.}
Commercial AVs, like Tesla, usually are equipped with more than one camera to sense the environment~\cite{tesla_camera}. Popular Tesla Vision~\cite{tesla_vision} based models, such as Model 3 and Model Y, are equipped with eight cameras and powerful vision processing which can stitch the image together and provide 360 degrees of visibility at up to 250 meters of range~\cite{tesla_camera,tesla_Stitching}. \attack can also work on AV when it is equipped with multiple cameras. To verify our attack on the AV, we apply the attack on the front middle camera, and then collect the 3 different front camera images. 
We stitch the images to show that the attack is not affected by the multiple cameras from the stitched image and the disparity map. 

\begin{figure}[t]
\centering
	\includegraphics[width=0.7\columnwidth]{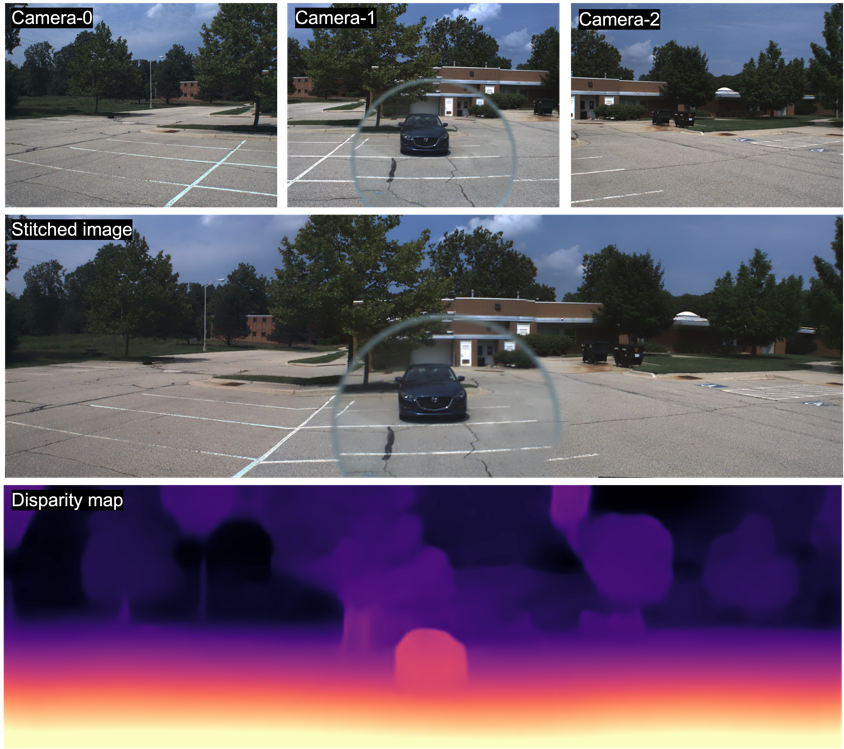}
	\caption{The images in the first row are captured from the 3 cameras on the AV, which are used to form the stitched image. The last image shows the disparity map of the stitched image.}
	\label{fig:stitched_image_depth}
\end{figure}

An example is shown in Fig.~\ref{fig:stitched_image_depth}, we collect images from the 3 FILR cameras on AV, i.e., camera-0, camera-1, and camera-2. Each of the cameras is with a $60\degree$ field of view horizontally. To emulate the vision processing on Tesla, we perform image stitching using these camera images. Then, we obtain its disparity map from Monodepth2. It can be seen that the attack is not affected by the multiple cameras from the stitched image and the disparity map. Therefore, \attack would still work for multi-camera AVs.

\textbf{Generality.}
The majority of car camera hardware could be vulnerable to \attack. 
By adjusting $f$ and $d_b$, we can manipulate the object depth. 
As the focal lengths of most car cameras are similar, by adjusting $d_b$, they will all fall victim to the proposed attack. Meanwhile, since the camera image serves as the input for the MDE models, the output depth map will be affected by our attack regardless of the model variations. 

Regarding vehicle software, existing AD systems have advanced object detection models. Despite being able to identify the presence of objects ahead in our experiments, these models lack distance information, making them ineffective in preventing our attack.

\textbf{Limitation.}
The primary limitation lies in the stealthiness of our attack. Captured images/video streams from the camera sensors are usually not visible to the human driver. However, our current attack prototype is noticeable. For commercial AVs, autopilot cameras are usually in the cabin or the car body. 
The car manufacturer usually installs the camera in an unobtrusive place by shielding it with the vehicle corners or the vehicle decoration~\cite{tesla_exterior}. 
The autopilot camera sensors are usually tiny, which allows us to launch our attack by simply sticking, taping, or absorbing the attack device on the car body near the car camera. We simulate our attack device implementation on Tesla Model 3 in Appendix~\ref{app:manufactured}.
To make the attack more stealthy, we could design and print a 
3D lens holder using some transparent materials, such as glass or plastic. However, stealthiness can still become an issue if the designed attack device is too large or placed in a relatively conspicuous place. 

Second, in the presence of a highly alert AD system, our attack could potentially be thwarted by blur detection. This could trigger an alert and prompt the system to cease operation until the attack lens is removed from the camera sensor. Therefore, blur detection can serve as a defense mechanism. We elaborate on various blur detection methods in the defense section below.

The final limitation is related to changing $f$ and controlling $d_b$. We know that $f$ and $d_b$ are the key parameters in controlling the attack depth. However, it is not easy to change the focal length of the attack lens $f$ without accessing the victim's vehicle. Therefore, we control $d_b$ using a design lens slider or a phone slider in the experiments to continuously manipulate the attack depth. With an increasing cost, a more sophisticated version of our attack could employ a purpose-built apparatus similar in principle to a consumer digital camera lens, perhaps even employing electrically tunable lenses to precisely calibrate focus, focal length, and lens distance.




\textbf{Defense.}
One of the most effective countermeasures is by using sensor fusion~\cite{wang2017sonic}, which combines the outputs of multiple sensors to produce an accurate result. Most AVs employ sensors other than cameras on their vehicles, such as Lidar, radar, and ultrasonic sensors. Unlike monocular cameras, these sensors convey accurate depth information, which obviates the need for monocular depth detection methods, and thus can defeat our attack. However, depending on the architecture of the fusion method in question, if monocular camera depth information is still used, our method could still reduce the accuracy. Manufacturers should test their systems against \attack to ensure that the fusion method they use is sufficiently robust.

Another defense method is to add a detection module that detects image blur, either in-lens blur or out-of-lens blur, caused by the proposed attack. 
Once a blur in the image is detected, the AV should trigger an alert, prompt the system to cease operation, and warn the human driver to inspect the physical condition of the car camera. Here, we use three methods to detect the image blur: variation of the Laplacian (VarLap)~\cite{Adrian_Rosebrock,WillBrennan}, High-frequency multiscale Fusion and Sort Transform (HiFST)~\cite{alireza2017spatially}, and local-based defocus blur segmentation (LDB)~\cite{yi2016lbp}. 

\begin{figure}[t]
\centering
	\includegraphics[width=1\columnwidth]{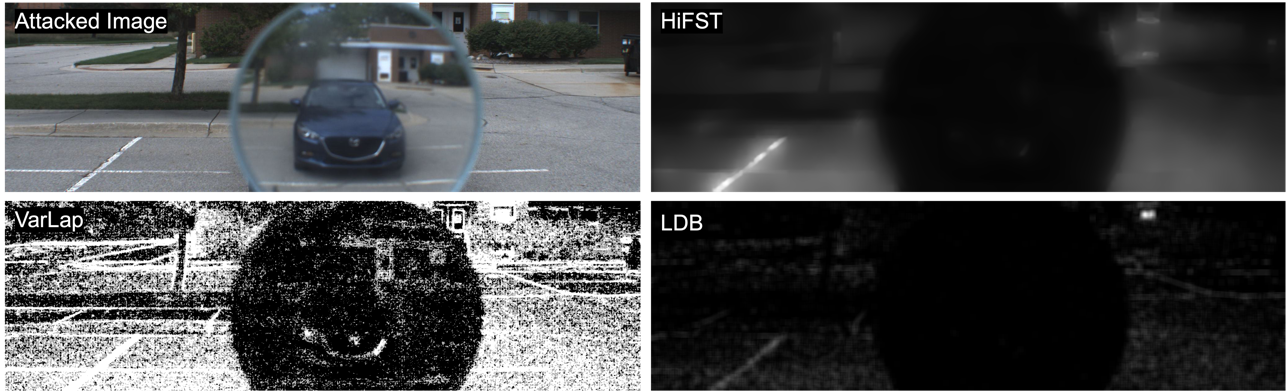}
	\caption{Detection results from three detection models.}
	\label{fig:detection}
\end{figure}

We use an image from the convex lens attack on the AV as input, and the output detection images are shown in Fig.~\ref{fig:detection}. 
VarLap utilizes the Laplacian operator, which serves the purpose of calculating the second derivative of an image. 
The basic idea is that with high variance in an image, there is a broad range of responses
typically seen in a clear and in-focus image. Conversely, in instances of very low variance, the responses are confined to a narrow range.
It is a well-known fact that an image has low variance when it becomes increasingly blurred. In Fig.~\ref{fig:detection}, the input image is blurred in the in-lens area, therefore, this region has narrow-range responses in the VarLap detection image. 

Blur can cause a reduction of the high-frequency components within an image. HiFST is an efficient method for detecting blur, which relies on a multiscale fusion and sorting transform approach focusing on high-frequency information. This method utilizes the Discrete Cosine Transform (DCT) coefficients of the gradient magnitudes from various resolutions to achieve its purpose. Fig.~\ref{fig:detection} shows that HiFST can well distinguish between the blurred and unblurred regions.

Another approach, LBP, uses a sharpness assessment metric grounded in local binary patterns, coupled with a resilient segmentation algorithm for distinguishing between in-focus and out-of-focus areas in images. This sharpness metric capitalizes on the insight that, in blurry regions, the majority of local image patches exhibit notably fewer specific local binary patterns when contrasted with patches in sharp areas. By integrating this metric with image matting and employing multi-scale inference, high-quality sharpness maps can be generated. Fig.~\ref{fig:detection} shows the sharpness map for our attack image, the blurry and non-blurry area is clearly separated. 

In addition to employing the aforementioned methods for detecting the attack, there is potential to devise an attack-aware depth estimation algorithm. This algorithm could correct the compromised depth by deblurring the image and rescaling the attacked area to the correct size and depth. We identify this as an area for future investigation and development.

\section{Related Work}

Zhang et al. are the first to investigate white-box adversarial attacks on MDE~\cite{zhang2020adversarial}. They employed imperceptible perturbations to execute three distinct types of attacks: non-targeted attacks on a specific image, targeted attacks on a particular object within an image, and universal attacks that can be applied to any image. Similarly, Wong et al. present a method for using imperceptible additive adversarial perturbations to selectively alter the perceived geometry of a scene for MDE~\cite{wong2020targeted}. To generalize the attacks, Daimo et al. propose black-box adversarial attacks on MDE using evolutionary multi-objective optimization~\cite{daimo2021black}. All of these attacks, however, are confined to the digital world and lack real-world applicability.

Recently, Yamanaka et al. devise artificial adversarial patches capable of deceiving the target methods into providing inaccurate depth estimations for the regions where these patterns were applied~\cite{yamanaka2020adversarial}. Different from the noticeable attack patches, Cheng et al. employ an optimization-based method to generate inconspicuous adversarial patches that are tailored towards physical objects to attack depth estimation~\cite{cheng2022physical}. Compared with existing white-box physical attacks, \attack enables a new genre of physical attack using optical lenses in a black-box setting, which is even more general and more robust.

\section{Conclusion}
In this paper, we present \attack, a new genre of physical attack towards MDE based AD systems using optical lenses.
By exploring the vulnerability of MDE, we formulate concave lens attack and convex lens attack mathematically. 
We conduct the attack simulation with different MDE algorithms to showcase the feasibility of our attack at both AI and system levels. Through extensive real-world experiments, we find \attack is effective across diverse attack parameter settings and at various object distances. The proposed optimization algorithm further improves the depth attack performance. 
The successful demonstration of \attack on monocular vision-based depth estimation suggests potential security implications
 on real-world AD systems.

\bibliographystyle{ACM-Reference-Format}
\bibliography{Reference}


\begin{thebibliography}{66}


\ifx \showCODEN    \undefined \def \showCODEN     #1{\unskip}     \fi
\ifx \showDOI      \undefined \def \showDOI       #1{#1}\fi
\ifx \showISBNx    \undefined \def \showISBNx     #1{\unskip}     \fi
\ifx \showISBNxiii \undefined \def \showISBNxiii  #1{\unskip}     \fi
\ifx \showISSN     \undefined \def \showISSN      #1{\unskip}     \fi
\ifx \showLCCN     \undefined \def \showLCCN      #1{\unskip}     \fi
\ifx \shownote     \undefined \def \shownote      #1{#1}          \fi
\ifx \showarticletitle \undefined \def \showarticletitle #1{#1}   \fi
\ifx \showURL      \undefined \def \showURL       {\relax}        \fi
\providecommand\bibfield[2]{#2}
\providecommand\bibinfo[2]{#2}
\providecommand\natexlab[1]{#1}
\providecommand\showeprint[2][]{arXiv:#2}

\bibitem[Abu~Alhaija et~al\mbox{.}(2018)]%
        {abu2018augmented}
\bibfield{author}{\bibinfo{person}{Hassan Abu~Alhaija}, \bibinfo{person}{Siva~Karthik Mustikovela}, \bibinfo{person}{Lars Mescheder}, \bibinfo{person}{Andreas Geiger}, {and} \bibinfo{person}{Carsten Rother}.} \bibinfo{year}{2018}\natexlab{}.
\newblock \showarticletitle{Augmented reality meets computer vision: Efficient data generation for urban driving scenes}.
\newblock \bibinfo{journal}{\emph{International Journal of Computer Vision}} \bibinfo{volume}{126}, \bibinfo{number}{9} (\bibinfo{year}{2018}), \bibinfo{pages}{961--972}.
\newblock


\bibitem[AI(2022)]%
        {tesla_Stitching}
\bibfield{author}{\bibinfo{person}{TOWARDS AI}.} \bibinfo{year}{2022}\natexlab{}.
\newblock \bibinfo{title}{Tesla’s Self Driving Algorithm Explained}.
\newblock \bibinfo{howpublished}{\url{https://towardsai.net/p/l/teslas-self-driving-algorithm-explained}}.
\newblock
\newblock
\shownote{Accessed: 2022-12-20}.


\bibitem[Alireza~Golestaneh and Karam(2017)]%
        {alireza2017spatially}
\bibfield{author}{\bibinfo{person}{S Alireza~Golestaneh} {and} \bibinfo{person}{Lina~J Karam}.} \bibinfo{year}{2017}\natexlab{}.
\newblock \showarticletitle{Spatially-varying blur detection based on multiscale fused and sorted transform coefficients of gradient magnitudes}. In \bibinfo{booktitle}{\emph{Proceedings of the IEEE conference on computer vision and pattern recognition}}. \bibinfo{pages}{5800--5809}.
\newblock


\bibitem[Brennan(2022)]%
        {WillBrennan}
\bibfield{author}{\bibinfo{person}{Will Brennan}.} \bibinfo{year}{2022}\natexlab{}.
\newblock \bibinfo{title}{BlurDetection2}.
\newblock \bibinfo{howpublished}{\url{https://github.com/WillBrennan/BlurDetection2}}.
\newblock
\newblock
\shownote{Accessed: 2022-12-20}.


\bibitem[CARLA(2022)]%
        {carlasensor}
\bibfield{author}{\bibinfo{person}{CARLA}.} \bibinfo{year}{2022}\natexlab{}.
\newblock \bibinfo{title}{Sensors reference}.
\newblock \bibinfo{howpublished}{\url{https://carla.readthedocs.io/en/0.9.11/ref_sensors}}.
\newblock
\newblock
\shownote{Accessed: 2022-12-20}.


\bibitem[Casser et~al\mbox{.}(2019)]%
        {casser2019depth}
\bibfield{author}{\bibinfo{person}{Vincent Casser}, \bibinfo{person}{Soeren Pirk}, \bibinfo{person}{Reza Mahjourian}, {and} \bibinfo{person}{Anelia Angelova}.} \bibinfo{year}{2019}\natexlab{}.
\newblock \showarticletitle{Depth prediction without the sensors: Leveraging structure for unsupervised learning from monocular videos}. In \bibinfo{booktitle}{\emph{Proceedings of the AAAI conference on artificial intelligence}}, Vol.~\bibinfo{volume}{33}. \bibinfo{pages}{8001--8008}.
\newblock


\bibitem[Chang and Chen(2018)]%
        {chang2018pyramid}
\bibfield{author}{\bibinfo{person}{Jia-Ren Chang} {and} \bibinfo{person}{Yong-Sheng Chen}.} \bibinfo{year}{2018}\natexlab{}.
\newblock \showarticletitle{Pyramid stereo matching network}. In \bibinfo{booktitle}{\emph{Proceedings of the IEEE conference on computer vision and pattern recognition}}. \bibinfo{pages}{5410--5418}.
\newblock


\bibitem[Chen et~al\mbox{.}(2016)]%
        {chen2016single}
\bibfield{author}{\bibinfo{person}{Weifeng Chen}, \bibinfo{person}{Zhao Fu}, \bibinfo{person}{Dawei Yang}, {and} \bibinfo{person}{Jia Deng}.} \bibinfo{year}{2016}\natexlab{}.
\newblock \showarticletitle{Single-image depth perception in the wild}.
\newblock \bibinfo{journal}{\emph{Advances in neural information processing systems}}  \bibinfo{volume}{29} (\bibinfo{year}{2016}).
\newblock


\bibitem[Cheng et~al\mbox{.}(2022)]%
        {cheng2022physical}
\bibfield{author}{\bibinfo{person}{Zhiyuan Cheng}, \bibinfo{person}{James Liang}, \bibinfo{person}{Hongjun Choi}, \bibinfo{person}{Guanhong Tao}, \bibinfo{person}{Zhiwen Cao}, \bibinfo{person}{Dongfang Liu}, {and} \bibinfo{person}{Xiangyu Zhang}.} \bibinfo{year}{2022}\natexlab{}.
\newblock \showarticletitle{Physical attack on monocular depth estimation with optimal adversarial patches}. In \bibinfo{booktitle}{\emph{European Conference on Computer Vision}}. Springer, \bibinfo{pages}{514--532}.
\newblock


\bibitem[Crystal(2022)]%
        {AmlongCrystal}
\bibfield{author}{\bibinfo{person}{Amlong Crystal}.} \bibinfo{year}{2022}\natexlab{}.
\newblock \bibinfo{title}{{Amlong Crystal Premium Optical Glass Double Convex and Concave Lens Set, 50mm Diameter, 3 Double Convex (20, 30, 50cm FL) and 3 Double Concave (20, 30, 50cm FL), 6 Piece Set}}.
\newblock \bibinfo{howpublished}{\url{https://www.amazon.com/gp/product/B07Z3CVFMB/ref=ppx_yo_dt_b_search_asin_title?ie=UTF8&th=1}}.
\newblock
\newblock
\shownote{Accessed: 2022-12-20}.


\bibitem[Daimo et~al\mbox{.}(2021)]%
        {daimo2021black}
\bibfield{author}{\bibinfo{person}{Renya Daimo}, \bibinfo{person}{Takahiro Suzuki}, {and} \bibinfo{person}{Satoshi Ono}.} \bibinfo{year}{2021}\natexlab{}.
\newblock \showarticletitle{Black-box Adversarial Attacks on Monocular Depth Estimation Using Evolutionary Multi-objective Optimization}. In \bibinfo{booktitle}{\emph{2021 IEEE International Conference on Systems, Man, and Cybernetics (SMC)}}. IEEE, \bibinfo{pages}{3466--3471}.
\newblock


\bibitem[Datar et~al\mbox{.}(2023b)]%
        {datar2023toward}
\bibfield{author}{\bibinfo{person}{Aniket Datar}, \bibinfo{person}{Chenhui Pan}, \bibinfo{person}{Mohammad Nazeri}, {and} \bibinfo{person}{Xuesu Xiao}.} \bibinfo{year}{2023}\natexlab{b}.
\newblock \showarticletitle{Toward Wheeled Mobility on Vertically Challenging Terrain: Platforms, Datasets, and Algorithms}.
\newblock \bibinfo{journal}{\emph{arXiv preprint arXiv:2303.00998}} (\bibinfo{year}{2023}).
\newblock


\bibitem[Datar et~al\mbox{.}(2023a)]%
        {datar2023learning}
\bibfield{author}{\bibinfo{person}{Aniket Datar}, \bibinfo{person}{Chenhui Pan}, {and} \bibinfo{person}{Xuesu Xiao}.} \bibinfo{year}{2023}\natexlab{a}.
\newblock \showarticletitle{Learning to Model and Plan for Wheeled Mobility on Vertically Challenging Terrain}.
\newblock \bibinfo{journal}{\emph{arXiv preprint arXiv:2306.11611}} (\bibinfo{year}{2023}).
\newblock


\bibitem[Dijk and Croon(2019)]%
        {dijk2019neural}
\bibfield{author}{\bibinfo{person}{Tom~van Dijk} {and} \bibinfo{person}{Guido~de Croon}.} \bibinfo{year}{2019}\natexlab{}.
\newblock \showarticletitle{How do neural networks see depth in single images?}. In \bibinfo{booktitle}{\emph{Proceedings of the IEEE/CVF International Conference on Computer Vision}}. \bibinfo{pages}{2183--2191}.
\newblock


\bibitem[Dosovitskiy et~al\mbox{.}(2017)]%
        {Dosovitskiy17}
\bibfield{author}{\bibinfo{person}{Alexey Dosovitskiy}, \bibinfo{person}{German Ros}, \bibinfo{person}{Felipe Codevilla}, \bibinfo{person}{Antonio Lopez}, {and} \bibinfo{person}{Vladlen Koltun}.} \bibinfo{year}{2017}\natexlab{}.
\newblock \showarticletitle{{CARLA}: {An} Open Urban Driving Simulator}. In \bibinfo{booktitle}{\emph{Proceedings of the 1st Annual Conference on Robot Learning}}. \bibinfo{pages}{1--16}.
\newblock


\bibitem[Dxomark(2020)]%
        {dxomark}
\bibfield{author}{\bibinfo{person}{Dxomark}.} \bibinfo{year}{2020}\natexlab{}.
\newblock \bibinfo{title}{{Apple iPhone 12 Pro Max Camera test: Big and beautiful}}.
\newblock \bibinfo{howpublished}{\url{https://shorturl.at/eoP08}}.
\newblock
\newblock
\shownote{Accessed: 2022-12-20}.


\bibitem[Eigen and Fergus(2015)]%
        {eigen2015predicting}
\bibfield{author}{\bibinfo{person}{David Eigen} {and} \bibinfo{person}{Rob Fergus}.} \bibinfo{year}{2015}\natexlab{}.
\newblock \showarticletitle{Predicting depth, surface normals and semantic labels with a common multi-scale convolutional architecture}. In \bibinfo{booktitle}{\emph{Proceedings of the IEEE international conference on computer vision}}. \bibinfo{pages}{2650--2658}.
\newblock


\bibitem[Eykholt et~al\mbox{.}(2018)]%
        {eykholt2018robust}
\bibfield{author}{\bibinfo{person}{Kevin Eykholt}, \bibinfo{person}{Ivan Evtimov}, \bibinfo{person}{Earlence Fernandes}, \bibinfo{person}{Bo Li}, \bibinfo{person}{Amir Rahmati}, \bibinfo{person}{Chaowei Xiao}, \bibinfo{person}{Atul Prakash}, \bibinfo{person}{Tadayoshi Kohno}, {and} \bibinfo{person}{Dawn Song}.} \bibinfo{year}{2018}\natexlab{}.
\newblock \showarticletitle{Robust physical-world attacks on deep learning visual classification}. In \bibinfo{booktitle}{\emph{Proceedings of the IEEE conference on computer vision and pattern recognition}}. \bibinfo{pages}{1625--1634}.
\newblock


\bibitem[Garg et~al\mbox{.}(2016)]%
        {garg2016unsupervised}
\bibfield{author}{\bibinfo{person}{Ravi Garg}, \bibinfo{person}{Vijay~Kumar Bg}, \bibinfo{person}{Gustavo Carneiro}, {and} \bibinfo{person}{Ian Reid}.} \bibinfo{year}{2016}\natexlab{}.
\newblock \showarticletitle{Unsupervised cnn for single view depth estimation: Geometry to the rescue}. In \bibinfo{booktitle}{\emph{European conference on computer vision}}. Springer, \bibinfo{pages}{740--756}.
\newblock


\bibitem[Geiger et~al\mbox{.}(2013)]%
        {geiger2013vision}
\bibfield{author}{\bibinfo{person}{Andreas Geiger}, \bibinfo{person}{Philip Lenz}, \bibinfo{person}{Christoph Stiller}, {and} \bibinfo{person}{Raquel Urtasun}.} \bibinfo{year}{2013}\natexlab{}.
\newblock \showarticletitle{Vision meets robotics: The kitti dataset}.
\newblock \bibinfo{journal}{\emph{The International Journal of Robotics Research}} \bibinfo{volume}{32}, \bibinfo{number}{11} (\bibinfo{year}{2013}), \bibinfo{pages}{1231--1237}.
\newblock


\bibitem[Godard et~al\mbox{.}(2017)]%
        {godard2017unsupervised}
\bibfield{author}{\bibinfo{person}{Cl{\'e}ment Godard}, \bibinfo{person}{Oisin Mac~Aodha}, {and} \bibinfo{person}{Gabriel~J Brostow}.} \bibinfo{year}{2017}\natexlab{}.
\newblock \showarticletitle{Unsupervised monocular depth estimation with left-right consistency}. In \bibinfo{booktitle}{\emph{Proceedings of the IEEE conference on computer vision and pattern recognition}}. \bibinfo{pages}{270--279}.
\newblock


\bibitem[Godard et~al\mbox{.}(2019)]%
        {monodepth2}
\bibfield{author}{\bibinfo{person}{Cl{\'{e}}ment Godard}, \bibinfo{person}{Oisin {Mac Aodha}}, \bibinfo{person}{Michael Firman}, {and} \bibinfo{person}{Gabriel~J. Brostow}.} \bibinfo{year}{2019}\natexlab{}.
\newblock \showarticletitle{Digging into Self-Supervised Monocular Depth Prediction}.
\newblock  (\bibinfo{date}{October} \bibinfo{year}{2019}).
\newblock


\bibitem[Guizilini et~al\mbox{.}(2020)]%
        {guizilini20203d}
\bibfield{author}{\bibinfo{person}{Vitor Guizilini}, \bibinfo{person}{Rares Ambrus}, \bibinfo{person}{Sudeep Pillai}, \bibinfo{person}{Allan Raventos}, {and} \bibinfo{person}{Adrien Gaidon}.} \bibinfo{year}{2020}\natexlab{}.
\newblock \showarticletitle{3d packing for self-supervised monocular depth estimation}. In \bibinfo{booktitle}{\emph{Proceedings of the IEEE/CVF Conference on Computer Vision and Pattern Recognition}}. \bibinfo{pages}{2485--2494}.
\newblock


\bibitem[in~Colour(2022)]%
        {cambridgeincolour}
\bibfield{author}{\bibinfo{person}{Cambridge in Colour}.} \bibinfo{year}{2022}\natexlab{}.
\newblock \bibinfo{title}{{CAMERAS VS. THE HUMAN EYE}}.
\newblock \bibinfo{howpublished}{\url{https://www.cambridgeincolour.com/tutorials/cameras-vs-human-eye.htm}}.
\newblock
\newblock
\shownote{Accessed: 2022-12-20}.


\bibitem[Jing et~al\mbox{.}(2021)]%
        {jing2021too}
\bibfield{author}{\bibinfo{person}{Pengfei Jing}, \bibinfo{person}{Qiyi Tang}, \bibinfo{person}{Yuefeng Du}, \bibinfo{person}{Lei Xue}, \bibinfo{person}{Xiapu Luo}, \bibinfo{person}{Ting Wang}, \bibinfo{person}{Sen Nie}, {and} \bibinfo{person}{Shi Wu}.} \bibinfo{year}{2021}\natexlab{}.
\newblock \showarticletitle{Too good to be safe: Tricking lane detection in autonomous driving with crafted perturbations}. In \bibinfo{booktitle}{\emph{30th USENIX Security Symposium (USENIX Security 21)}}. \bibinfo{pages}{3237--3254}.
\newblock


\bibitem[Jocher et~al\mbox{.}(2023)]%
        {Jocher_YOLO_by_Ultralytics_2023}
\bibfield{author}{\bibinfo{person}{Glenn Jocher}, \bibinfo{person}{Ayush Chaurasia}, {and} \bibinfo{person}{Jing Qiu}.} \bibinfo{year}{2023}\natexlab{}.
\newblock \bibinfo{booktitle}{\emph{{YOLO by Ultralytics}}}.
\newblock
\urldef\tempurl%
\url{https://github.com/ultralytics/ultralytics}
\showURL{%
\tempurl}


\bibitem[Li and Ibanez-Guzman(2020)]%
        {li2020lidar}
\bibfield{author}{\bibinfo{person}{You Li} {and} \bibinfo{person}{Javier Ibanez-Guzman}.} \bibinfo{year}{2020}\natexlab{}.
\newblock \showarticletitle{Lidar for autonomous driving: The principles, challenges, and trends for automotive lidar and perception systems}.
\newblock \bibinfo{journal}{\emph{IEEE Signal Processing Magazine}} \bibinfo{volume}{37}, \bibinfo{number}{4} (\bibinfo{year}{2020}), \bibinfo{pages}{50--61}.
\newblock


\bibitem[Liu et~al\mbox{.}(2015)]%
        {liu2015deep}
\bibfield{author}{\bibinfo{person}{Fayao Liu}, \bibinfo{person}{Chunhua Shen}, {and} \bibinfo{person}{Guosheng Lin}.} \bibinfo{year}{2015}\natexlab{}.
\newblock \showarticletitle{Deep convolutional neural fields for depth estimation from a single image}. In \bibinfo{booktitle}{\emph{Proceedings of the IEEE conference on computer vision and pattern recognition}}. \bibinfo{pages}{5162--5170}.
\newblock


\bibitem[Loxonis(2022)]%
        {fixed_focus}
\bibfield{author}{\bibinfo{person}{Loxonis}.} \bibinfo{year}{2022}\natexlab{}.
\newblock \bibinfo{title}{Auto-Focus vs Fixed-Focus}.
\newblock \bibinfo{howpublished}{\url{https://shorturl.at/cfuyK}}.
\newblock
\newblock
\shownote{Accessed: 2022-12-20}.


\bibitem[Man et~al\mbox{.}(2020)]%
        {man2020ghostimage}
\bibfield{author}{\bibinfo{person}{Yanmao Man}, \bibinfo{person}{Ming Li}, {and} \bibinfo{person}{Ryan Gerdes}.} \bibinfo{year}{2020}\natexlab{}.
\newblock \showarticletitle{Ghostimage: Perception domain attacks against vision-based object classification systems}.
\newblock \bibinfo{journal}{\emph{arXiv preprint arXiv:2001.07792}} (\bibinfo{year}{2020}).
\newblock


\bibitem[Mayer et~al\mbox{.}(2016)]%
        {mayer2016large}
\bibfield{author}{\bibinfo{person}{Nikolaus Mayer}, \bibinfo{person}{Eddy Ilg}, \bibinfo{person}{Philip Hausser}, \bibinfo{person}{Philipp Fischer}, \bibinfo{person}{Daniel Cremers}, \bibinfo{person}{Alexey Dosovitskiy}, {and} \bibinfo{person}{Thomas Brox}.} \bibinfo{year}{2016}\natexlab{}.
\newblock \showarticletitle{A large dataset to train convolutional networks for disparity, optical flow, and scene flow estimation}. In \bibinfo{booktitle}{\emph{Proceedings of the IEEE conference on computer vision and pattern recognition}}. \bibinfo{pages}{4040--4048}.
\newblock


\bibitem[Nassi et~al\mbox{.}(2020)]%
        {nassi2020phantom}
\bibfield{author}{\bibinfo{person}{Ben Nassi}, \bibinfo{person}{Yisroel Mirsky}, \bibinfo{person}{Dudi Nassi}, \bibinfo{person}{Raz Ben-Netanel}, \bibinfo{person}{Oleg Drokin}, {and} \bibinfo{person}{Yuval Elovici}.} \bibinfo{year}{2020}\natexlab{}.
\newblock \showarticletitle{Phantom of the ADAS: Securing advanced driver-assistance systems from split-second phantom attacks}. In \bibinfo{booktitle}{\emph{Proceedings of the 2020 ACM SIGSAC conference on computer and communications security}}. \bibinfo{pages}{293--308}.
\newblock


\bibitem[of~Encyclopaedia~Britannica(2020)]%
        {Britannica}
\bibfield{author}{\bibinfo{person}{The~Editors of Encyclopaedia~Britannica}.} \bibinfo{year}{2020}\natexlab{}.
\newblock \bibinfo{title}{{lens}}.
\newblock \bibinfo{howpublished}{\url{https://www.britannica.com/technology/lens-optics}}.
\newblock
\newblock
\shownote{Accessed: 2022-12-20}.


\bibitem[Pillai et~al\mbox{.}(2019)]%
        {pillai2019superdepth}
\bibfield{author}{\bibinfo{person}{Sudeep Pillai}, \bibinfo{person}{Rare{\c{s}} Ambru{\c{s}}}, {and} \bibinfo{person}{Adrien Gaidon}.} \bibinfo{year}{2019}\natexlab{}.
\newblock \showarticletitle{Superdepth: Self-supervised, super-resolved monocular depth estimation}. In \bibinfo{booktitle}{\emph{2019 International Conference on Robotics and Automation (ICRA)}}. IEEE, \bibinfo{pages}{9250--9256}.
\newblock


\bibitem[Piotrowsky et~al\mbox{.}(2019)]%
        {piotrowsky2019enabling}
\bibfield{author}{\bibinfo{person}{Lukas Piotrowsky}, \bibinfo{person}{Timo Jaeschke}, \bibinfo{person}{Simon Kueppers}, \bibinfo{person}{Jan Siska}, {and} \bibinfo{person}{Nils Pohl}.} \bibinfo{year}{2019}\natexlab{}.
\newblock \showarticletitle{Enabling high accuracy distance measurements with FMCW radar sensors}.
\newblock \bibinfo{journal}{\emph{IEEE Transactions on Microwave Theory and Techniques}} \bibinfo{volume}{67}, \bibinfo{number}{12} (\bibinfo{year}{2019}), \bibinfo{pages}{5360--5371}.
\newblock


\bibitem[Rosebrock(2022)]%
        {Adrian_Rosebrock}
\bibfield{author}{\bibinfo{person}{Adrian Rosebrock}.} \bibinfo{year}{2022}\natexlab{}.
\newblock \bibinfo{title}{Blur detection with OpenCV}.
\newblock \bibinfo{howpublished}{\url{https://pyimagesearch.com/2015/09/07/blur-detection-with-opencv/}}.
\newblock
\newblock
\shownote{Accessed: 2022-12-20}.


\bibitem[Sato et~al\mbox{.}(2021)]%
        {sato2021dirty}
\bibfield{author}{\bibinfo{person}{Takami Sato}, \bibinfo{person}{Junjie Shen}, \bibinfo{person}{Ningfei Wang}, \bibinfo{person}{Yunhan Jia}, \bibinfo{person}{Xue Lin}, {and} \bibinfo{person}{Qi~Alfred Chen}.} \bibinfo{year}{2021}\natexlab{}.
\newblock \showarticletitle{Dirty road can attack: Security of deep learning based automated lane centering under Physical-World attack}. In \bibinfo{booktitle}{\emph{30th USENIX Security Symposium (USENIX Security 21)}}. \bibinfo{pages}{3309--3326}.
\newblock


\bibitem[Shen et~al\mbox{.}(2022)]%
        {shen2022sok}
\bibfield{author}{\bibinfo{person}{Junjie Shen}, \bibinfo{person}{Ningfei Wang}, \bibinfo{person}{Ziwen Wan}, \bibinfo{person}{Yunpeng Luo}, \bibinfo{person}{Takami Sato}, \bibinfo{person}{Zhisheng Hu}, \bibinfo{person}{Xinyang Zhang}, \bibinfo{person}{Shengjian Guo}, \bibinfo{person}{Zhenyu Zhong}, \bibinfo{person}{Kang Li}, {et~al\mbox{.}}} \bibinfo{year}{2022}\natexlab{}.
\newblock \showarticletitle{Sok: On the semantic ai security in autonomous driving}.
\newblock \bibinfo{journal}{\emph{arXiv preprint arXiv:2203.05314}} (\bibinfo{year}{2022}).
\newblock


\bibitem[Tay et~al\mbox{.}(2019)]%
        {tay2019aanet}
\bibfield{author}{\bibinfo{person}{Chiat-Pin Tay}, \bibinfo{person}{Sharmili Roy}, {and} \bibinfo{person}{Kim-Hui Yap}.} \bibinfo{year}{2019}\natexlab{}.
\newblock \showarticletitle{Aanet: Attribute attention network for person re-identifications}. In \bibinfo{booktitle}{\emph{Proceedings of the IEEE/CVF conference on computer vision and pattern recognition}}. \bibinfo{pages}{7134--7143}.
\newblock


\bibitem[Tesla(2022a)]%
        {Autopilot}
\bibfield{author}{\bibinfo{person}{Tesla}.} \bibinfo{year}{2022}\natexlab{a}.
\newblock \bibinfo{title}{Autopilot}.
\newblock \bibinfo{howpublished}{\url{https://www.tesla.com/autopilot}}.
\newblock
\newblock
\shownote{Accessed: 2022-12-20}.


\bibitem[Tesla(2022b)]%
        {Autopilot_and_full_self}
\bibfield{author}{\bibinfo{person}{Tesla}.} \bibinfo{year}{2022}\natexlab{b}.
\newblock \bibinfo{title}{Autopilot and Full Self-Driving Capability}.
\newblock \bibinfo{howpublished}{\url{https://www.tesla.com/support/autopilot}}.
\newblock
\newblock
\shownote{Accessed: 2022-12-20}.


\bibitem[Tesla(2022c)]%
        {tesla_exterior}
\bibfield{author}{\bibinfo{person}{Tesla}.} \bibinfo{year}{2022}\natexlab{c}.
\newblock \bibinfo{title}{Exterior}.
\newblock \bibinfo{howpublished}{\url{https://www.tesla.com/ownersmanual/model3/en_us/GUID-6C6C3944-9674-4E81-A0E8-94D60B6D87B9.html}}.
\newblock
\newblock
\shownote{Accessed: 2022-12-20}.


\bibitem[Tesla(2022d)]%
        {tesla_camera}
\bibfield{author}{\bibinfo{person}{Tesla}.} \bibinfo{year}{2022}\natexlab{d}.
\newblock \bibinfo{title}{Otopilot}.
\newblock \bibinfo{howpublished}{\url{https://shorturl.at/iloIK}}.
\newblock
\newblock
\shownote{Accessed: 2022-12-20}.


\bibitem[Tesla(2022e)]%
        {tesla_vision}
\bibfield{author}{\bibinfo{person}{Tesla}.} \bibinfo{year}{2022}\natexlab{e}.
\newblock \bibinfo{title}{Tesla Vision Update: Replacing Ultrasonic Sensors with Tesla Vision}.
\newblock \bibinfo{howpublished}{\url{https://www.tesla.com/support/transitioning-tesla-vision}}.
\newblock
\newblock
\shownote{Accessed: 2022-12-20}.


\bibitem[Uber(2022)]%
        {Uber}
\bibfield{author}{\bibinfo{person}{Uber}.} \bibinfo{year}{2022}\natexlab{}.
\newblock \bibinfo{title}{Self-driving car technology by uber}.
\newblock \bibinfo{howpublished}{\url{https://www.uber.com/us/en/atg/technology/}}.
\newblock
\newblock
\shownote{Accessed: 2022-12-20}.


\bibitem[Wang et~al\mbox{.}(2018)]%
        {wang2018learning}
\bibfield{author}{\bibinfo{person}{Chaoyang Wang}, \bibinfo{person}{Jos{\'e}~Miguel Buenaposada}, \bibinfo{person}{Rui Zhu}, {and} \bibinfo{person}{Simon Lucey}.} \bibinfo{year}{2018}\natexlab{}.
\newblock \showarticletitle{Learning depth from monocular videos using direct methods}. In \bibinfo{booktitle}{\emph{Proceedings of the IEEE conference on computer vision and pattern recognition}}. \bibinfo{pages}{2022--2030}.
\newblock


\bibitem[Wang et~al\mbox{.}(2017)]%
        {wang2017sonic}
\bibfield{author}{\bibinfo{person}{Zhengbo Wang}, \bibinfo{person}{Kang Wang}, \bibinfo{person}{Bo Yang}, \bibinfo{person}{Shangyuan Li}, {and} \bibinfo{person}{Aimin Pan}.} \bibinfo{year}{2017}\natexlab{}.
\newblock \showarticletitle{Sonic gun to smart devices: Your devices lose control under ultrasound/sound}.
\newblock \bibinfo{journal}{\emph{Black Hat USA}} (\bibinfo{year}{2017}), \bibinfo{pages}{1--50}.
\newblock


\bibitem[Watson et~al\mbox{.}(2019)]%
        {watson2019self}
\bibfield{author}{\bibinfo{person}{Jamie Watson}, \bibinfo{person}{Michael Firman}, \bibinfo{person}{Gabriel~J Brostow}, {and} \bibinfo{person}{Daniyar Turmukhambetov}.} \bibinfo{year}{2019}\natexlab{}.
\newblock \showarticletitle{Self-supervised monocular depth hints}. In \bibinfo{booktitle}{\emph{Proceedings of the IEEE/CVF International Conference on Computer Vision}}. \bibinfo{pages}{2162--2171}.
\newblock


\bibitem[Waymo(2022)]%
        {Waymo}
\bibfield{author}{\bibinfo{person}{Waymo}.} \bibinfo{year}{2022}\natexlab{}.
\newblock \bibinfo{title}{Waymo Technology}.
\newblock \bibinfo{howpublished}{\url{https://www.waymo.com/tech/}}.
\newblock
\newblock
\shownote{Accessed: 2022-12-20}.


\bibitem[Wikipedia(2022a)]%
        {Depthoffield}
\bibfield{author}{\bibinfo{person}{Wikipedia}.} \bibinfo{year}{2022}\natexlab{a}.
\newblock \bibinfo{title}{{Depth of field}}.
\newblock \bibinfo{howpublished}{\url{https://en.wikipedia.org/wiki/Depth_of_field}}.
\newblock
\newblock
\shownote{Accessed: 2022-12-20}.


\bibitem[Wikipedia(2022b)]%
        {fixed_focus_wiki}
\bibfield{author}{\bibinfo{person}{Wikipedia}.} \bibinfo{year}{2022}\natexlab{b}.
\newblock \bibinfo{title}{Fixed-focus lens}.
\newblock \bibinfo{howpublished}{\url{https://en.wikipedia.org/wiki/Fixed-focus_lens}}.
\newblock
\newblock
\shownote{Accessed: 2022-12-20}.


\bibitem[Wikipedia(2022c)]%
        {Focus}
\bibfield{author}{\bibinfo{person}{Wikipedia}.} \bibinfo{year}{2022}\natexlab{c}.
\newblock \bibinfo{title}{{Focus (optics)}}.
\newblock \bibinfo{howpublished}{\url{https://en.wikipedia.org/wiki/Focus_(optics)}}.
\newblock
\newblock
\shownote{Accessed: 2022-12-20}.


\bibitem[Wikipedia(2022d)]%
        {Pinholecameramodel}
\bibfield{author}{\bibinfo{person}{Wikipedia}.} \bibinfo{year}{2022}\natexlab{d}.
\newblock \bibinfo{title}{Pinhole camera model}.
\newblock \bibinfo{howpublished}{\url{https://en.wikipedia.org/wiki/Pinhole_camera_model}}.
\newblock
\newblock
\shownote{Accessed: 2022-12-20}.


\bibitem[Wittpahl et~al\mbox{.}(2018)]%
        {wittpahl2018realistic}
\bibfield{author}{\bibinfo{person}{Christian Wittpahl}, \bibinfo{person}{Hatem~Ben Zakour}, \bibinfo{person}{Matthias Lehmann}, {and} \bibinfo{person}{Alexander Braun}.} \bibinfo{year}{2018}\natexlab{}.
\newblock \showarticletitle{Realistic image degradation with measured PSF}.
\newblock \bibinfo{journal}{\emph{arXiv preprint arXiv:1801.02197}} (\bibinfo{year}{2018}).
\newblock


\bibitem[Wong et~al\mbox{.}(2020)]%
        {wong2020targeted}
\bibfield{author}{\bibinfo{person}{Alex Wong}, \bibinfo{person}{Safa Cicek}, {and} \bibinfo{person}{Stefano Soatto}.} \bibinfo{year}{2020}\natexlab{}.
\newblock \showarticletitle{Targeted adversarial perturbations for monocular depth prediction}.
\newblock \bibinfo{journal}{\emph{Advances in neural information processing systems}}  \bibinfo{volume}{33} (\bibinfo{year}{2020}), \bibinfo{pages}{8486--8497}.
\newblock


\bibitem[Yamanaka et~al\mbox{.}(2020)]%
        {yamanaka2020adversarial}
\bibfield{author}{\bibinfo{person}{Koichiro Yamanaka}, \bibinfo{person}{Ryutaroh Matsumoto}, \bibinfo{person}{Keita Takahashi}, {and} \bibinfo{person}{Toshiaki Fujii}.} \bibinfo{year}{2020}\natexlab{}.
\newblock \showarticletitle{Adversarial patch attacks on monocular depth estimation networks}.
\newblock \bibinfo{journal}{\emph{IEEE Access}}  \bibinfo{volume}{8} (\bibinfo{year}{2020}), \bibinfo{pages}{179094--179104}.
\newblock


\bibitem[Yan et~al\mbox{.}(2022)]%
        {yan2022rolling}
\bibfield{author}{\bibinfo{person}{Chen Yan}, \bibinfo{person}{Zhijian Xu}, \bibinfo{person}{Zhanyuan Yin}, \bibinfo{person}{Xiaoyu Ji}, {and} \bibinfo{person}{Wenyuan Xu}.} \bibinfo{year}{2022}\natexlab{}.
\newblock \showarticletitle{Rolling Colors: Adversarial Laser Exploits against Traffic Light Recognition}.
\newblock \bibinfo{journal}{\emph{arXiv preprint arXiv:2204.02675}} (\bibinfo{year}{2022}).
\newblock


\bibitem[Yi and Eramian(2016)]%
        {yi2016lbp}
\bibfield{author}{\bibinfo{person}{Xin Yi} {and} \bibinfo{person}{Mark Eramian}.} \bibinfo{year}{2016}\natexlab{}.
\newblock \showarticletitle{LBP-based segmentation of defocus blur}.
\newblock \bibinfo{journal}{\emph{IEEE transactions on image processing}} \bibinfo{volume}{25}, \bibinfo{number}{4} (\bibinfo{year}{2016}), \bibinfo{pages}{1626--1638}.
\newblock


\bibitem[Yin et~al\mbox{.}(2019)]%
        {yin2019enforcing}
\bibfield{author}{\bibinfo{person}{Wei Yin}, \bibinfo{person}{Yifan Liu}, \bibinfo{person}{Chunhua Shen}, {and} \bibinfo{person}{Youliang Yan}.} \bibinfo{year}{2019}\natexlab{}.
\newblock \showarticletitle{Enforcing geometric constraints of virtual normal for depth prediction}. In \bibinfo{booktitle}{\emph{Proceedings of the IEEE/CVF International Conference on Computer Vision}}. \bibinfo{pages}{5684--5693}.
\newblock


\bibitem[Yin and Shi(2018)]%
        {yin2018geonet}
\bibfield{author}{\bibinfo{person}{Zhichao Yin} {and} \bibinfo{person}{Jianping Shi}.} \bibinfo{year}{2018}\natexlab{}.
\newblock \showarticletitle{Geonet: Unsupervised learning of dense depth, optical flow and camera pose}. In \bibinfo{booktitle}{\emph{Proceedings of the IEEE conference on computer vision and pattern recognition}}. \bibinfo{pages}{1983--1992}.
\newblock


\bibitem[Zhang et~al\mbox{.}(2023)]%
        {zhang2023lite}
\bibfield{author}{\bibinfo{person}{Ning Zhang}, \bibinfo{person}{Francesco Nex}, \bibinfo{person}{George Vosselman}, {and} \bibinfo{person}{Norman Kerle}.} \bibinfo{year}{2023}\natexlab{}.
\newblock \showarticletitle{Lite-mono: A lightweight cnn and transformer architecture for self-supervised monocular depth estimation}. In \bibinfo{booktitle}{\emph{Proceedings of the IEEE/CVF Conference on Computer Vision and Pattern Recognition}}. \bibinfo{pages}{18537--18546}.
\newblock


\bibitem[Zhang et~al\mbox{.}(2020)]%
        {zhang2020adversarial}
\bibfield{author}{\bibinfo{person}{Ziqi Zhang}, \bibinfo{person}{Xinge Zhu}, \bibinfo{person}{Yingwei Li}, \bibinfo{person}{Xiangqun Chen}, {and} \bibinfo{person}{Yao Guo}.} \bibinfo{year}{2020}\natexlab{}.
\newblock \showarticletitle{Adversarial attacks on monocular depth estimation}.
\newblock \bibinfo{journal}{\emph{arXiv preprint arXiv:2003.10315}} (\bibinfo{year}{2020}).
\newblock


\bibitem[Zhou et~al\mbox{.}(2023)]%
        {zhou2023comprehensive}
\bibfield{author}{\bibinfo{person}{Ce Zhou}, \bibinfo{person}{Qian Li}, \bibinfo{person}{Chen Li}, \bibinfo{person}{Jun Yu}, \bibinfo{person}{Yixin Liu}, \bibinfo{person}{Guangjing Wang}, \bibinfo{person}{Kai Zhang}, \bibinfo{person}{Cheng Ji}, \bibinfo{person}{Qiben Yan}, \bibinfo{person}{Lifang He}, {et~al\mbox{.}}} \bibinfo{year}{2023}\natexlab{}.
\newblock \showarticletitle{A comprehensive survey on pretrained foundation models: A history from bert to chatgpt}.
\newblock \bibinfo{journal}{\emph{arXiv preprint arXiv:2302.09419}} (\bibinfo{year}{2023}).
\newblock


\bibitem[Zhou et~al\mbox{.}(2024)]%
        {zhou2024lensattack}
\bibfield{author}{\bibinfo{person}{Ce Zhou}, \bibinfo{person}{Qiben Yan}, \bibinfo{person}{Daniel Kent}, {and} \bibinfo{person}{Guangjing Wang}.} \bibinfo{year}{2024}\natexlab{}.
\newblock \showarticletitle{Optical Lens Attack on Deep Learning Based Monocular Depth Estimation}. In \bibinfo{booktitle}{\emph{20th EAI International Conference on Security and Privacy in Communication Networks}}.
\newblock


\bibitem[Zhou et~al\mbox{.}(2022)]%
        {zhou2022doublestar}
\bibfield{author}{\bibinfo{person}{Ce Zhou}, \bibinfo{person}{Qiben Yan}, \bibinfo{person}{Yan Shi}, {and} \bibinfo{person}{Lichao Sun}.} \bibinfo{year}{2022}\natexlab{}.
\newblock \showarticletitle{DoubleStar: Long-Range Attack Towards Depth Estimation based Obstacle Avoidance in Autonomous Systems}. In \bibinfo{booktitle}{\emph{31st USENIX Security Symposium (USENIX Security 22)}}. \bibinfo{pages}{1885--1902}.
\newblock


\bibitem[Zhou et~al\mbox{.}(2017)]%
        {zhou2017unsupervised}
\bibfield{author}{\bibinfo{person}{Tinghui Zhou}, \bibinfo{person}{Matthew Brown}, \bibinfo{person}{Noah Snavely}, {and} \bibinfo{person}{David~G Lowe}.} \bibinfo{year}{2017}\natexlab{}.
\newblock \showarticletitle{Unsupervised learning of depth and ego-motion from video}. In \bibinfo{booktitle}{\emph{Proceedings of the IEEE conference on computer vision and pattern recognition}}. \bibinfo{pages}{1851--1858}.
\newblock


\end{thebibliography}

\appendix
\section{Appendix}
\subsection{Pinhole Camera Model}\label{sec:Pinhole Camera Model}
\begin{figure}[htbp]
\centering
	\includegraphics[width=0.4\textwidth]{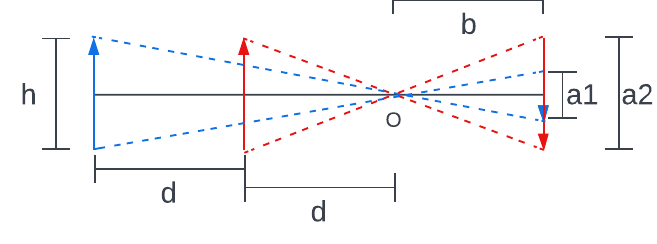}
	\caption{The perspective projection of the pinhole camera.}
	\label{fig:pinhole}
\end{figure}

The perspective projection of the pinhole camera is shown in Fig.~\ref{fig:pinhole}. $O$ denotes the pinhole position. $h$ represents the size of the two objects. $b$ is the distance between the pinhole to the image sensor, and $d$ and $2d$ stand for the distance between the two same-size objects and the pinhole, respectively. Based on the triangulation, we have: 
\begin{equation}\label{pinholeequ1}
    \frac{h}{d}=\frac{a_{1}}{b},
\end{equation}
\begin{equation}\label{pinholeequ2}
    \frac{h}{2d}=\frac{a_{2}}{b}.
\end{equation}    
Now, Using Eqs.~(\ref{pinholeequ1}) and (\ref{pinholeequ2}), we have: 
\begin{equation}
    \frac{2d}{d}=\frac{a_{1}}{a_{2}}=2.
\end{equation}   
Thus, we can conclude that the depth of the same object is inversely proportional to its object size in the image. 

\subsection{Attack Device on Manufactured Vehicle}\label{app:manufactured}

\begin{figure}
    \centering    
    \subfigure[Full view of the attack]{\includegraphics[width=0.45\textwidth]{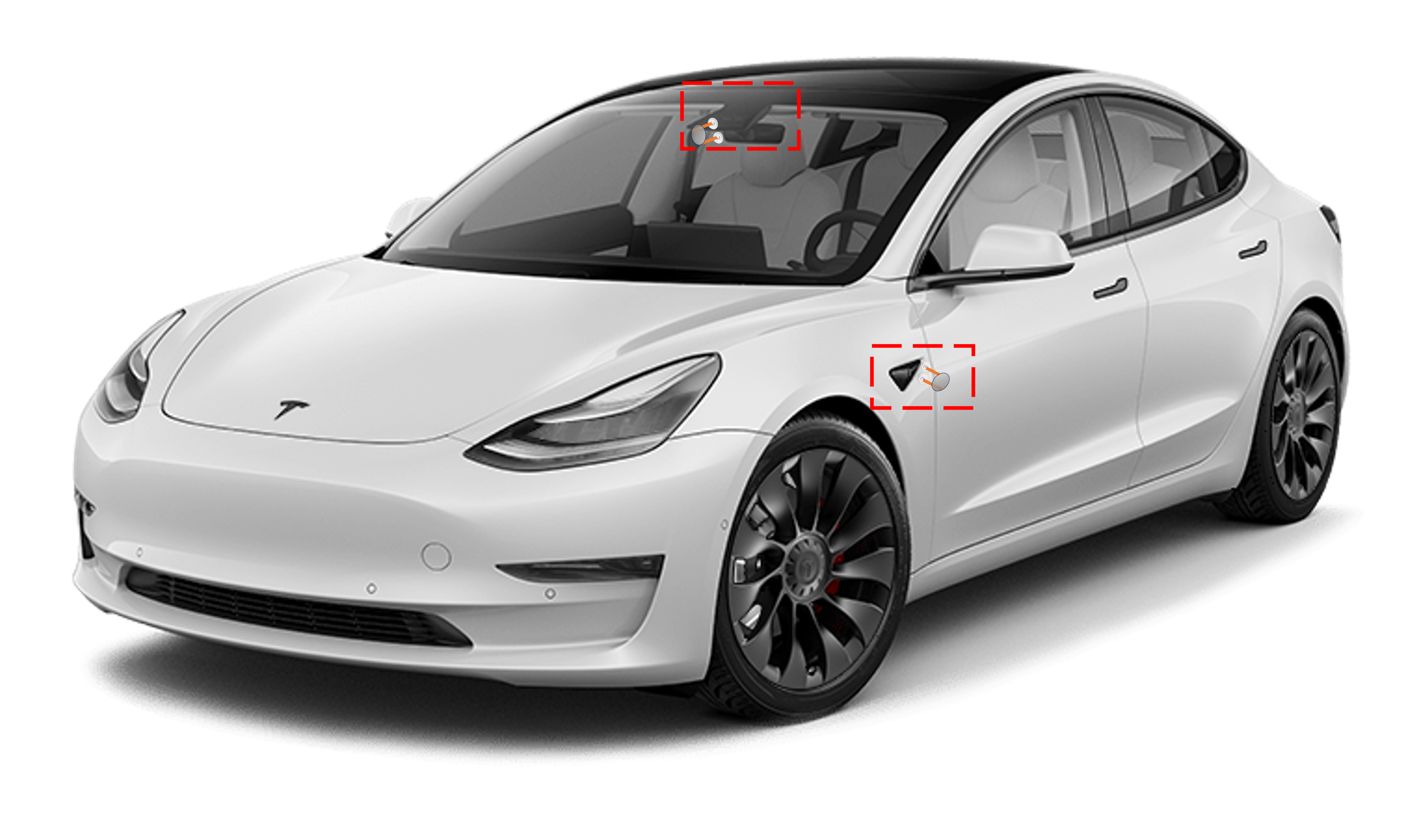}}
    \subfigure[Attack device on front camera]{\includegraphics[width=0.26\textwidth]{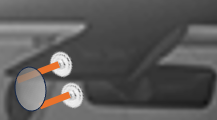}} 
    \subfigure[Attack device on side camera]{\includegraphics[width=0.26\textwidth]{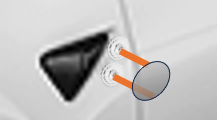}}      
    \caption{(a) In deploying the attack device on the Tesla Model 3, the designated location for the autopilot cameras is indicated by the red dotted line. The attack device consists of an attack lens (greyish transparent ellipse), a lens holder (the two yellowish sticks), two fixed suction cups (the bluish dots), and a remote control module (not shown in the figure). (b) and (c) details the attack device implemented on the front and side cameras. }
    \label{fig:Manufactured}
\end{figure}

We simulate the attack device on a manufactured vehicle (i.e., Tesla Model 3) in Fig.~\ref{fig:Manufactured}. The attack device comprises an attack lens, a lens holder, two fixed suction cups, and a remote control module (ensuring the proposed intermittent attack). Using two fixed suction cups, the attack device can be placed in different locations. Compromising the front autopilot camera may result in a collision with the car in front or behind. Conversely, an attack on the side autopilot camera can lead to a collision with another car in an adjacent lane when the AV is changing lanes.

Concerning stealthiness, from Fig.~\ref{fig:experimental_setup}(a) and Fig.~\ref{fig:Manufactured}(a), it is evident that the attack device appears relatively small in the overall view of the attack, making it less likely to be noticed by the human driver if they are not actively paying attention. However, when the attack targets the front camera, it becomes more noticeable to the human driver. In contrast, the attack is more stealthy when directed at the side camera.

\subsection{Attack Types in Real AD Scenarios}\label{sec:Attack Types in Real AD Scenarios}

In real AD scenarios, the value of $d_b$ typically remains relatively small, for instance, around $5cm$. Additionally, the object is frequently situated at a considerable distance from the camera, often reaching approximately $10m$. In a concave lens attack, no matter what the value of $d_{o1}$ is, the formed image is always virtual and upright, meaning that it always works.

Regarding the convex lens attack, when $0<d_{o1}<f$ (first attack scenario), the formed images are real and upright in the combination lenses as long as $d_b \geq 0$. However, the object will be very close to the camera (e.g., $50cm$), which is not normal in AD scenarios. For the second attack scenario, when $d_{o1}>f$, the formed images will either become real and inverted when $d_b-|d_{i1}|>f_c$, or virtual and upright when $0<d_b-|d_{i1}|<f_c$. Since $f_c$ is a very small value, we need to ensure $d_b>|d_{i1}|$, meaning that the distance between the attack lens and camera is larger than the focal length of the attack lens $f$, which contradicts the physical attack scenarios as discussed in the Section~\ref{sec:Real-World Physical Attack}. For the third attack scenario, $d_{o1}>f$ and $d_b<|d_{i1}|$, the formed images are real and upright.
Here, we highlight that only the concave lens attack and the third attack scenario of the convex lens attack are feasible in practical AD scenarios.





\end{document}